\documentclass{article}

\PassOptionsToPackage{numbers, compress, authoryear}{natbib}

 \usepackage[preprint]{neurips_2025}

\setcitestyle{authoryear,open={(},close={)}} %
\renewcommand\cite{\citep}

\usepackage{tikz, tikz-qtree}
\usepackage{microtype}
\usepackage{amsmath,amsthm,amssymb}
\usepackage{xy}
\usepackage{graphicx}
\usepackage{caption}
\usepackage{subcaption}
\usepackage[dvipsnames]{xcolor}
\usepackage[most]{tcolorbox}

\usepackage{tabularx}     %
\usepackage{array}  %
\usepackage{tabularray}
\usepackage{multirow}
\usepackage{longtable}
\usepackage{pifont}   %

\usepackage{enumitem}  %
\usepackage{graphicx}
\usepackage{array}
\usepackage{ragged2e}  %
\usepackage{float}
\usepackage{booktabs}
\usepackage{makecell}  %
\usepackage{url}

\newcommand{\PL}
{\mbox{\texttt{Passive}$\rightarrow$\texttt{Logical}}}
\newcommand{\PA}{\mbox{\texttt{Passive}$\rightarrow$\texttt{Active}}}
\newcommand{\swap}{\mbox{\texttt{Swap}}}

\newcommand{\RASPApp}{Appendix~\ref{sec:RASPApp}}
\newcommand{\qkvlgrammar}{Appendix~\ref{sec:qkvlgrammar}}
\newcommand{\Apppsltoqkvl}{Appendix~\ref{sec:Apppsltoqkvl}}
\newcommand{\TPR}{Appendix~\ref{sec:TPR}}
\newcommand{\trainscratch}{Appendix~\ref{sec:trainscratch}}

\newcommand{\llmtesting}{App.~\ref{sec:llmtesting}}
\newcommand{\pslgrammar}{App.~\ref{sec:pslgrammar}}
\newcommand{\datop}{App.~\ref{sec:datop}}
\newcommand{\RelWkapp}{App.~\ref{sec:appRelW}}

\usepackage{gb4e}   %
\usepackage[final]{showlabels}   %

\usepackage[breaklinks]{hyperref}   %

\renewcommand{\rm}[1]{\textrm{#1}}
\newcommand{\ignore}[1]{}

\newcommand{\Q}{$\mathcal{Q}$}
\newcommand{\A}{$\mathcal{A}$}
\newcommand{\F}{$\mathcal{F}$}
\newcommand{\q}{\textsf{q}}
\renewcommand{\k}{\textsf{k}}
\renewcommand{\v}{\textsf{v}}
\newcommand{\g}[1]{\color{orange}#1\color{black}}
\newcommand{\gsf}[1]{\g{\textsf{#1}}}
\newcommand{\ssf}[1]{\textsf{#1}}
\newcommand{\NF}{\textsc{NextField}}
\newcommand{\CF}{\textsc{ContField}}
\newcommand{\PG}{\textsc{ParGen}}

\newcommand{\bp}{$^{\backprime}$}
\newcommand{\pn}{\mbox{position[N]@pos\_decrement}}
\renewcommand{\c}[2]{{\color{#1}#2\color{black}}}

\renewcommand{\sf}[1]{{\color{blue}\textsf{#1}\color{black}}} %
\newcommand{\squeeze}[2]{\scalebox{#1}[1.0]{#2}}
\renewcommand{\vec}[1]{{\overrightarrow{#1}}}

\newcounter{boxNum}
\newcommand{\newBox}[2]{\refstepcounter{boxNum}\label{#1}\textbf{Box~\theboxNum.\ #2}}

\usepackage{listings}

\usepackage{xcolor}

\definecolor{codegreen}{rgb}{0,0.6,0}
\definecolor{codegray}{rgb}{0.5,0.5,0.5}
\definecolor{codepurple}{rgb}{0.58,0,0.82}
\definecolor{backcolour}{rgb}{0.95,0.95,0.92}

\lstdefinestyle{mystyle}{
  backgroundcolor=\color{backcolour}, commentstyle=\color{codegreen},
  keywordstyle=\color{magenta},
  numberstyle=\tiny\color{codegray},
  stringstyle=\color{codepurple},
  basicstyle=\ttfamily\footnotesize,
  breakatwhitespace=false,         
  breaklines=true,                 
  captionpos=b,                    
  keepspaces=true,                 
  numbers=left,                    
  numbersep=5pt,                  
  showspaces=false,                
  showstringspaces=false,
  showtabs=false,                  
  tabsize=2
}

\title{Mechanisms of Symbol Processing for In-Context Learning in Transformer Networks}

\author{%
  Paul Smolensky \\
  Deep Learning Group\\
  Microsoft Research\\
  Redmond, WA 98052, USA \\
  \texttt{psmo@microsoft.com} \\
   \And
  Roland Fernandez \\
  Deep Learning Group \\
  Microsoft Research\\
  Redmond, WA 98052, USA \\
  \texttt{rfernand@microsoft.com} \\
   \And
  Zhenghao Herbert Zhou  \\
  Linguistics Department \\
  Yale University \\
  New Haven CT 06511, USA \\
  \texttt{herbert.zhou@yale.edu}  \\
   \And
  Mattia Opper  \\
   ILCC \\
   University of Edinburgh \\
  Edinburgh EH8 9AB, UK  \\
  \texttt{M.Opper@ed.ac.uk} \\
   \And
  Adam Davies \\
  Siebel School of Computing \\
  UIUC \\
  Urbana IL 61801, USA \\
  \texttt{adavies4@illinois.edu } \\
   \And
  Jianfeng Gao \\
  Deep Learning Group \\
  Microsoft Research\\
  Redmond, WA 98052, USA \\
  \texttt{jfgao@microsoft.com} \\
}

\begin{document}

\maketitle

\begingroup
\renewcommand\thefootnote{*}
\footnotetext{Corresponding Authors: P. Smolensky and R. Fernandez.~
}
\endgroup

\begin{abstract}
Large Language Models (LLMs) have demonstrated impressive abilities in symbol processing through in-context learning (ICL).
This success flies in the face of decades of critiques asserting that artificial neural networks cannot master abstract symbol manipulation.
We seek to understand the mechanisms that can enable robust symbol processing in transformer networks, illuminating both the unanticipated success, and the significant limitations, of transformers in symbol processing.
Borrowing insights from symbolic AI and cognitive science on the power of Production System architectures, we develop a high-level Production System Language, PSL, that allows us to write symbolic programs to do complex, abstract symbol processing, and create compilers that precisely implement PSL programs in transformer networks which are, by construction, 100\% mechanistically interpretable.
The work is driven by study of a purely abstract (semantics-free) symbolic task that we develop, Templatic Generation (TGT).
Although developed through study of TGT, PSL is, we demonstrate, highly general: it is Turing Universal.
The new type of transformer architecture that we compile from PSL programs suggests a number of paths for enhancing transformers' capabilities at symbol processing.
We note, however, that the work we report addresses computability,
and not learnability, by transformer networks. 
\newline

\noindent{\textbf{Note:}} The first section provides an extended synopsis of the entire paper.
\end{abstract}

\section{Paper Synopsis: How can Transformers Perform Complex Symbol Processing?}
  \label{sec:intro}
The unprecedented performance of generative language models (LMs) such as those of the GPT family \cite[et seq.]{radford2018improving} creates a dilemma for the science of intelligence. 
Fundamental to virtually all classic theories of natural and artificial intelligence are structured symbolic representations and symbolic processing of such representations \cite[see Sec.~\ref{sec:motiv}]{hinzen2012oxford}. 
These provide the basis for explaining the pervasive compositionality of intelligent cognition (see Box \ref{box:comp}) \mbox{\cite{prince1988subsymbols}}.
Neural networks appear to be singularly unsuited for such computation, and should fail catastrophically, for example, at compositional language processing, including the ultimate challenge for abstract structure processing, generating complex, syntactically valid natural language \cite{pinker1988language}. 
Yet neural language models with transformer architectures \cite{vaswani2017attention} dramatically out-perform all symbolic-computation-based models of language processing, generating rich, syntactically complex English, virtually flawlessly \cite{chang2024language}.

\begin{tcolorbox}[breakable, enhanced]
\newBox{box:comp}
{Compositionality and Systematicity in Cognition}
\hfill \break

Human cognition copes effectively with a huge range of phenomena by representing complex entities as structured assemblies of simpler entities \cite{russin2024frege}.
Knowledge derived from previous experience with simpler entities can give rise to knowledge about a novel complex entity by composing existing knowledge of the simpler entities of which the complex entity is composed.
This is \textit{compositional generalization}.
\hfill \medskip

This is all so fundamental and natural that we take it for granted.
Yet even current state-of-the-art neural networks do not display the extremely robust compositional generalization that is characteristic of human cognition.
\hfill \medskip

Formalizing compositional structures as discrete symbol structures such as parse trees or knowledge graphs gave classic symbolic AI excellent compositional generalization --- to the extent that natural phenomena could be successfully decomposed into discrete parts.
However, adequately decomposing natural complex entities into recombinable discrete constituent parts typically proved beyond the capabilities of discrete symbolic methods.
If English syntax could be fully formalized with a discrete rewrite-rule grammar, then a symbolic NLP system in possession of that grammar would exhibit perfect compositional generalization across the entire language.
Despite decades of attempts, however, reducing natural language to such compositional rules has failed to adequately cover the actual richness of language use.
\hfill \medskip

Yet compositional analysis still provides the deepest understanding of the portion of natural language, and cognition generally, that it is able to cover.
And while contemporary neural AI systems display extraordinary coverage, their often dramatic failures of compositional generalization suggest that however these systems `understand' the world, they do so in a fundamentally different way than we do. 
\hfill \medskip

This cursory discussion of compositionality in cognition is necessarily greatly oversimplified, but faithfully captures the core ideas.
Relatedly, \textit{systematic generalization}, often seen as an aspect of compositional generalization, consists of generalizing, to new concepts, the capabilities already learned for other concepts.

\end{tcolorbox}

As predicted by influential critiques from the 1980s, transformer networks do often struggle with compositionality \cite{dziri2023faith}.
Nonetheless, they sometimes perform rather well on certain tests of systematicity and compositionality \cite{sinha2024survey} --- in particular, when tested via `in-context learning' (ICL), the subject of the work presented here \cite{brown2020language}.
As illustrated in Box~\ref{box:ICL1}, in the type of ICL we study, a transformer is given an input symbol string which is a `prompt' that includes an example in `question-answer' format; this exemplifies a symbolic template, specific to that particular prompt, for mapping the `question' into the `answer'.
The input next provides new symbolic material in the `question' format, which must be inserted into that template to generate a `completion', the corresponding symbol string in the `answer' format.
(A formal presentation is given in Sec.~\ref{sec:Fdef}.)

Most language tasks are heavily laden with semantics.
While traditional theories of semantics \cite{heim1998semantics}, and classical AI systems \cite{barr1981handbook}, attempted to capture semantic knowledge with discrete symbolic representations and rules (e.g., expert systems, Cowan, 2001\nocite{cowan2001expert}), the recent revolutionary progress in AI has shown the greater power of statistical inference on numerical vector representations induced from massive quantities of natural-language data.
Neural networks, especially transformers, have shown extraordinary capacity for such statistical inference \cite{goodfellow2016deep}.

\begin{tcolorbox}[breakable, enhanced]
\newBox{box:ICL1}
{Symbol Manipulation with In-Context Learning: Templatic Generation}
\hfill \break

Exemplifying `in-context learning', Large Language Models can take the prompt (\ref{ex:pr1}) and generate a continuation (\ref{ex:cont1}) (Sec.~\ref{sec:expts}), consistent with (i) recognizing the initial \Q/\A\ in the prompt as the template (\ref{ex:temp1}) --- the English-passive-voice-to-predicate-logical-form transformation --- with $x$ = `the program' $V$ = `translated', $y$ = `a compiler', and (ii) using this template to generate a continuation by inserting $x$ = `my big dog', $V$ = `chased', $y$ = `a small black cat'.  
\begin{exe} 
  \ex  \label{ex:pr1} \textit{Prompt:} \Q\ the program was translated by a compiler \A\  translated ( a compiler , the program ) \hfill\break \Q\ my big dog was chased by a small black cat \A\
  \ex \label{ex:cont1} \textit{Continuation:} chased ( a small black cat , my big dog )
  \ex \label{ex:temp1} \textit{Template:} \Q\ $x$ was $V$ by $y$ \A\ $V\ (\ y\ , x\ )$ 
\end{exe}
Another type of template is a simple inference rule in propositional logic.
\begin{exe}
    \ex \label{ex:pr2} \textit{Prompt:} \Q\ x =$>$ y \A\ y or not x \Q\ ( u and v ) =$>$ z \A
    \ex \label{ex:cont2} \textit{Continuation:} z or not ( u and v )
    \ex \label{ex:temp2} \textit{Template:} \Q\ $p$ =$>$ $q$ \A\ $q$ or not $p$ 
\end{exe}
Or a basic algebraic equality:
\begin{exe}
    \ex \label{ex:pr3} \textit{Prompt:} \Q\ log \{ x * y \} \A\ log ( x ) + log [ y ] \Q\ log \{ 3a * b\^{}2 \} \A
    \ex \label{ex:cont3} \textit{Continuation:} log ( 3a ) + log [ b\^{}2 ]
    \ex \label{ex:temp3} \textit{Template:} \Q\ log \{ $u$ * $v$ \} \A\ log ( $u$ ) + log [ $v$ ] 
\end{exe}
Or even a nonsensical pattern:
\begin{exe}
    \ex \label{ex:pr4} \textit{Prompt:}
\Q\ $\sim$ es zd ey db ak ) fx \$ \{ tr dz , + vj kj zo \% jq hu rd ag
\A\ \_ vj kj zo \$ es zd ey db ak / jq hu rd ag * fx .
\Q\  $\sim$ dv he ) vv bo td \$ \{ xh dp qc my mz , + qk \% hw oc cw uh 
\A
    \ex \label{ex:cont4} \textit{Continuation:}
\_ qk \$ dv he / hw oc cw uh * vv bo td .
    \ex \label{ex:temp4} \textit{Template:}
\Q\  $\sim$ $x$ ) $y$ \$ \{ $z$ , + $u$ \% $v$
\A\  \_ $u$ \$ $x$ / $v$ * $y$ .
\end{exe}

\end{tcolorbox}

\subsection{Semantics-Free Symbol Processing} \label{sec:sem-free}
Semantically-laden tasks have of course been the subject of extensive research in the AI community.
But language processing also demands \textit{semantics-free} inference, most notably, in syntax. 
Natural language syntax has been best characterized in terms of purely formal, abstract relationships between words and phrases, captured by grammars using symbolic computations that are, by design, blind to the meanings of elements, sensitive only to the formal patterns of combination among these elements.
Chomsky famously made this point vividly with his meaningless but syntactically valid sentence, \textit{colorless green ideas sleep furiously} \cite{chomsky1957syntactic}.

Influential critiques of neural network models of cognition have, since the rise to prominence of these models in the 1980s, argued that neural computation is ill-suited to just this kind of abstract, meaning-free computation, so purely captured by symbolic computation (Fodor and Pylyshyn \citeyear{fodor1988connectionism}, with many thousands of citations; Marcus \citeyear{marcus2001algebraic}). 
Yet the text generated by neural language models is essentially perfect syntactically.
Perhaps classical theory has been mistaken to believe that mastering syntax requires symbolic computation.
But perhaps neural LMs can actually implement a good approximation to symbolic computation where needed, and this is what enables their syntactic mastery.
This is the hypothesis we pursue in this work.

While LMs' syntax is virtually error-free, they and other neural models do not seem to have an equal mastery of compositionality and systematicity \cite{dziri2023faith}.
If these networks have implemented symbolic computational capabilities, these abilities are clearly limited.

\textit{\textbf{Our goal} in this work is to understand the mechanisms within transformers that provide them the symbol processing abilities they do possess, and to see how these mechanisms can be strengthened to overcome the limitations in symbol processing that transformer models still have.}
We will end up developing a type of attention head that enables symbolic computation.
In the future work we propose in Sec.~\ref{sec:unif}, these heads are to be combined with traditional attention heads in a new type of transformer which retains the current power for semantically-laden inference but fuses this with enhanced power for abstract meaning-free symbol processing. 

Our new attention heads result from studying what can enable a transformer to perfectly perform a task that is purely semantics-free: the \textit{Templatic Generation Task, TGT}.
The examples of TGT given in Box~\ref{box:ICL1} start off being patterns with some meaning for us, in terms of natural language syntax or logical or mathematical inference. 
The task we study encompasses such interesting cases, but we are not actually concerned with knowledge of that sort; the final example in the Box (\ref{ex:pr4} -- \ref{ex:temp4}) 
achieves the meaning-free symbol manipulation capability we are targeting: the symbol strings through which the templates are exemplified in the prompt are randomly-generated sequences of randomly-generated symbols.

The distinction between semantics-laden and semantics-free inference can be clarified by considering (\ref{ex:pr2}).
Generating the continuation in (\ref{ex:cont2}) via identifying the template in (\ref{ex:temp2}) requires only copying and rearranging the symbols in the prompt; no knowledge is required of the semantics of the symbol `$=>$' or of the notion of material implication in logic.
Genuinely testing for reasoning ability calls for a model to recognize from the meaning of the phrases in a passage that certain concepts stand in an implicational relation and knowing that this relation licenses drawing new conclusions about those concepts.
Testing ICL with (\ref{ex:pr2}), as we do, is testing not for reasoning abilities, but rather, abstract-symbol-processing abilities.%
\footnote{Thus the model we design here is not to be tested on standard reasoning tasks, but rather, on the purely semantics-free symbol-processing task we design, TGT.}

In this context, consider previous work on ICL that has examined prompts such as `\Q\ France \A\ Paris \Q\ Spain \A' \cite{hendel2023context}, or prompts that provide numerical-vector pairs $(x,y)$ where $y$ is a hidden affine function of $x$, which the model must infer from the given pairs and apply to a novel value of $x$ 
\cite[see Secs.~\ref{sec:iclfunclearning}--\ref{sec:icllearnprocess}]{garg2022can,akyurek2022whatlearningalgorithm}.
Now following natural-language-semantic associations, e.g., between countries and their capitals, and inferring affine numerical functions, are exactly the kinds of abilities neural networks have long been known to possess.
In contrast, semantics-free ICL tasks such as the Templatic Generation Task test networks for just the type of abstract processing that classic critiques have claimed is beyond neural computation. 
As a minimal illustration motivating the use of our new task, TGT, note that if we present the prompt `\Q\ twice x \A\ x x \Q\ twice a \A' and we get the desired continuation `a a', we want to be sure it is the result of correctly filling in the given template, and not simple application of the model's prior knowledge (e.g., from massive English pretraining data) of the semantics of `twice'; in fact, we'd want the same continuation from the prompt `\Q\ thrice x \A\ x x \Q\ thrice a \A', in violation of the English semantics of `thrice', and indeed the same continuation from `\Q\ GBq3 x \A\ x x \Q\ GBq3 a \A'.
(For a rare study of meaning-free-ness in a LM, see Lasri et al., 2022\nocite{lasri2022subject}.)

The semantics-free TGT is instantiated in the paper's primary case-study, which is inspired by cases such as (\ref{ex:pr1} -- \ref{ex:temp1}): the \swap\ task, presented in Sec.~\ref{sec:case}. 
This task allows us to focus entirely on how abstract, meaning-free pattern recognition over abstract, meaning-free tokens is possible within neural computation.
The TGT not only covers many interesting cases like those illustrated in Box~\ref{box:ICL1}, it also demands nearly all the abstract symbol-processing capabilities, provided for free by symbolic computation, which have long been claimed to be both necessary for human-level cognition and beyond the capabilities of neural computation; we present these in (\ref{ex:props}). 

The main contribution of this paper is an explicit demonstration of how transformers can actually provide an implementational platform for fundamental aspects of symbol processing: a platform which, of course, supports extremely powerful learning. 
To the extent that future work reveals that the neural mechanisms we bring to light here are at work in trained language models, this would constitute a vindication of symbolic theory, which, when embedded in neural computation, would have been proved capable of arising through data-driven learning --- a result that has yet to be achieved with purely symbolic computation.

\subsection{Research Questions   \label{sec:Qs}}

Our goal is to understand the mechanisms that provide transformers the capabilities that they do possess for abstract, semantics-free symbol processing.
For guidance in this challenging pursuit, we turn to Richard Feynman's famous dictum, ``what I cannot create, I do not understand'' \cite{gleick1993genius}: we design and hand-program a type of transformer network that demonstrably performs ICL of the type illustrated in Box~\ref{box:ICL1}.
In this network, the contribution that is made by every neuron and every connection  to producing this capability is perfectly well understood --- because we designed the network ourselves.

Our research questions are formulated in (\ref{ex:ourQs}).

\begin{exe}
    \ex Our questions
    \label{ex:ourQs}
    \begin{xlist}
        \ex How \textit{could} any neural network employing fairly standard, general-purpose mechanisms 
        do ICL?
        \label{ex:ourQ1}
        \ex Can general-purpose operations used by \textit{the transformer architecture} help design such a network?
        \label{ex:ourQ2}
        \ex Can insights from classic, \textit{symbolic AI} help design such a network?
        \label{ex:ourQ3}
   \end{xlist}
\end{exe}
A preview of the answers we offer is given in (\ref{ex:ourAs}).

\subsection{The Transformer Production Framework (TPF)   \label{sec:TPF}}
The primary contribution of the present work is the \textbf{\textit{Transformer Production Framework}} (\textbf{TPF}); we use it to study in-context learning to fill in meaningless, randomly-generated symbolic templates, but it can be applied much more widely. 
In TPF, a computational system is described at multiple levels: we follow the proposal
that contemporary machine-learning work take advantage of the three levels of description of a computational system famously proposed by the legendary cognitive scientist David Marr (\citeyear{marr1982vision}), both in training \cite{hamrick2020levels} and interpreting \cite{davies2024cognitive} foundation models.
These levels are: the \textit{functional} level (what Marr somewhat confusingly called the `computational' level), which abstractly specifies the function a computational system computes; the \textit{algorithmic} level, specifying the system's algorithm for computing this function; and the \textit{implementational} level, specifying how that algorithm is implemented in a physical system.

For our algorithmic level, we develop two symbolic languages for transformer computation.
The higher-level programming language, the \textit{Production-System Language} (\textit{PSL}), borrows the notion of production system architecture from classic symbolic computation theory, briefly summarized in (\ref{ex:PS}) \cite{jones2003production}.
Production systems include Emil Post's rewrite-rule systems \cite{post1943formal} and the string-rewriting systems at the foundation of the theory of formal grammars \cite{book1993string}, all key to the classic theory of (symbolic) computation \cite{hopcroft2000introduction}.
Early, special-purpose neural models --- very different from the transformer --- were developed in the late 1980s for implementing production systems \cite{touretzky1988distributed,dolan1989tensor}.

\begin{exe}
    \ex Production Systems
    \label{ex:PS}
    \begin{xlist}
        \ex A \textit{production} is a \textit{Condition-Action} pair: when the symbols in a common workspace meet the purely formal (meaning-free) requirements of the Condition, the Action may be taken, which adds or deletes symbols from the workspace.
        \ex In the simplest case, the productions are linearly ordered and apply in sequence, the sequence being executed repeatedly until some termination condition is met.
        \ex That human intelligence is best modeled by a production system is the founding principle of multiple leading theories of the computational architecture of human cognition \cite{anderson2005human,laird2019soar,ritter2019act}.%
        \footnote{See Ryu \& Lewis (\citeyear{ryu2021accounting}) for explication of a transformer's sentence-processing behavior in terms of a theory of human sentence processing \cite{lewis2005activation} couched in a general production-system-based cognitive architecture \cite{anderson2005human}.
        }
        \ex Cognitive Architecture production systems often have built-in complex capabilities of symbolic-pattern-matching, enabling Conditions to be quite complex; but in our production system, conditions can only require that specified variables have specified values, and actions can only write values into variables. 
        We want to understand how complex symbol processing can emerge through the interactions of productions that individually possess much less powerful built-in symbolic capabilities: specifically, productions simple enough to be precisely implementable in neural networks with a type of transformer architecture.
    \end{xlist}
\end{exe}

An outline of the paper is given in (\ref{ex:outline}); the core consists of the 3-level presentation of TPF in (\ref{ex:outfn}) -- (\ref{ex:outimpl}) [Secs.~\ref{sec:fun} -- \ref{sec:impl}].

\begin{exe}
\ex Paper outline
    \label{ex:outline}
    \begin{xlist}
        \ex We first discuss related research in which this work is situated [Sec.~\ref{sec:rel}].
        \label{ex:outrel}
        \ex We next motivate the case study forming the core of the paper: \swap\ [Sec.~\ref{sec:motiv}].
        \label{ex:outmotiv}
        \ex Adopting a functional level of description, we present a class of symbolic
        template-filling functions which we seek to compute, motivated by the discussion (\ref{ex:outmotiv}); we also show that these functions can be computed, with varying success, by pre-trained LMs and by transformers trained from scratch on the task [Sec.~\ref{sec:fun}].
        \label{ex:outfn}
        \ex Moving to the algorithmic level, we present an algorithm, \PG, for computing such functions using two abstract symbolic machines that we design: 
        \begin{xlist}
              \ex first, the algorithm is stated in a Production System Language (PSL) for a Production System Machine (PSM) [Sec.~\ref{sec:algI}]; 
              \ex then the algorithm is compiled into a program for the QKV Machine, a type of symbolic transformer architecture [Sec.~\ref{sec:algII}].
        \label{ex:outalg}        
        \end{xlist}
           \ex An implementation-level description is produced by a compiler for exactly implementing QKV programs in a version of the transformer neural network architecture that deploys the residual stream/hidden state discretely, and uses a type of discrete attention: DAT, a Discrete-Attention-only Transformer.
           Experiments verify the correctness of the algorithms and their implementation.
        \label{ex:outimpl} [Sec.~\ref{sec:impl}].
        \ex The generality of the framework is established by theorems asserting the Turing-Universality of the language PSL [Sec.~\ref{sec:Turing}] (see Sec.~\ref{sec:univ} below).
        \ex A general discussion of the work is offered, including how it can promote mechanical interpretability of LMs and trained transformers.
        Future work is suggested, including a discussion of extending compositionality further [Sec.~\ref{sec:fut}].
        \ex Appendices within the article provide
        \begin{enumerate}[label=\Alph*.] 
            \item  a walk-through of the symbolic computation implicit in the Swap task 
        
            \item  details of our ICL algorithm, \PG   

            \item further related work
            
            \item a formal grammar for Templatic Generation Tasks 
            
            \item a formal grammar for the Production-System Language PSL 
            
            \item details on the operation of the DAT 

            \item the system-prompt prefix used to test trained transformers on the Templatic Generation Task

            \item exploratory testing of the GPT-4 LLM model on the TGT dataset

            \item a detailed discussion of Weiss et al.'s (\citeyear{weiss2021rasp}) RASP language for programming transformers, and relations between RASP and our TPF

            \item a formal description of QKVL programs             
            
            \item a formal compiler for translating a PSL program to an equivalent QKVL program 
            
            \item a generalization of the analysis which exploits Tensor Product Representations 
            
            \item exploratory from-scratch training and testing of various sequence-to-sequence architectures on the TGT dataset
        \end{enumerate}

    \end{xlist}
\end{exe}

As a brief preview, our results yield the following general answers to our questions (\ref{ex:ourQs}).

\pagebreak

\begin{exe}
    \ex Our questions: preview of answers
    \label{ex:ourAs}
    \begin{xlist}
        \ex How \textit{could} any neural network employing fairly standard, general-purpose mechanisms do ICL?
        \label{ex:ourA1}
        \begin{itemize}
            \item[] By using hidden-state --- or `residual-stream' \cite{elhage2021mathematical} --- encodings of state variables describing each input symbol, including structural variables that encode the role of the symbol in an abstract parse tree. 
        \end{itemize}
        \ex Can general-purpose operations used by \textit{the transformer architecture} help design such a network?
        \label{ex:ourA2}
        \begin{itemize}
            \item[] The network can target relevant previous symbols for information access by `query-key matching' of variable values (crucially including structural variables), and can set new values for variables through `value vectors' returned by attention. 
        \end{itemize}
        \ex Can insights from classic, \textit{symbolic AI} help design such a network?
        \label{ex:ourA3}
        \begin{itemize}
            \item[] Just as Production System architectures have enabled powerful symbolic systems for AI and for modeling human higher-level cognition, a Production System programming language can be designed which (i) can be used to write a general, fully interpretable symbolic  program for templatic text generation through ICL, and (ii) can be implemented in a transformer by using key-query matching to satisfy production Conditions, and value vectors to perform production Actions. 
        \end{itemize}
    \end{xlist}
\end{exe}

In more detail, our work develops the correspondences spelled out in (\ref{ex:correspondences}), some of which are familiar from previous work. 
(We return to these in Sec. \ref{sec:lessons}.)

\begin{exe}
    \ex How can ICL in a transformer perform symbolic templatic text generation? 
    \label{ex:correspondences}
    
    Via the following [transformer element] $\sim$ [symbolic element] correspondences: 
    \begin{xlist}
        \ex a cell's residual stream $\sim$ a variable-value structure%
        \footnote{
        As when defining an environment for evaluating a function or creating a function closure.
        }
        \begin{xlist}
            \ex a subspace of the hidden space $\sim$ a state variable
            \ex a vector component within a variable's subspace $\sim$ a value of that variable
        \end{xlist}
        \ex a layer's internal connections $\sim$ a production\footnote{Simple production systems are like NNs in lacking global control structure beyond mere sequencing of productions/layers.}
        \begin{xlist}
            \ex query-key matching in attention $\sim$ evaluating the condition of the layer's production
            \ex value vectors $\sim$ the production's action        
            \ex query-key matching on a subspace corresponding to a goal $\sim$ conditional branching for goal-directed action 
        \end{xlist}
        \ex a nested set of structural variables $\sim$ hierarchical data structure 
       \begin{xlist}
            \ex sharing the value of a level-$l$ structural variable $\sim$ in the same (type of) level-$l$ constituent
            --- adapted from Hinton (\citeyear{hinton2023represent})
       \end{xlist}
       \ex  a sequence of structural-variable values (at the `field' level) $\sim$ the sequence of abstract roles defining a template
    \end{xlist}
\end{exe}

It remains for future work to determine whether TPF can shed light on the ICL performed by trained language models. 
To this end, we suggest concrete hypotheses, and propose methods for testing them, in Sec.~\ref{sec:LMs}. 
How language-model training gives rise to such computation is beyond the scope of the current paper: the work reported here addresses \textit{computability}, and not \textit{learnability}, by transformer networks --- despite the potentially confusing `L' in `ICL'.

\subsection{Turing-Universality of the Results     \label{sec:univ}}
Although the presentation here focuses heavily on the ICL of templatic text generation, our results on what transformers can compute is actually much more general.
We show here how programs in the Production System Language PSL can be naturally compiled into transformer networks whose behavior is completely explainable through the PSL programs they implement.
But how general is the class of computations that can be expressed as PSL programs?
In fact, \textit{every computable function} can be computed by a PSL program: PSL is Turing complete.
This is shown in Sec.~\ref{sec:Turing}.
Note that this is Turing-completeness of a language for programming transformers, a higher-level result than the Turing-completeness of transformer networks themselves --- the Turing-completeness of DAT transformers follows from that of PSL; the Turing-completeness of hard-attention transformers shown by P\'erez et al.~(\citeyear{perez2021attention}) is a closely-related result. 
Thus the present work speaks not only to how transformers can carry out the powerful symbolic computation involved in ICL templatic text generation, but also how transformers can in fact function as universal computers.

\section{Related Work \label{sec:rel}}
This work complements other ongoing work on understanding neural computation, and specifically, in-context learning in transformers.  Unlike most mechanistic interpretability work, we are not analyzing trained models, probing for trees or other data structures.  Our question is one of \textit{mechanistic \textbf{computability}}: what can the transformer architecture compute in ICL, and exactly how? We focus on using ICL for the general but tightly constrained task of templatic text generation, as defined in \ref{sec:Fdef}. Our approach is to design an algorithm for performing templatic generation, in a symbolic form, which can be compiled into weights for a modified transformer architecture. The resulting network is, by construction, fully mechanistically interpretable.

The work pursued here is complementary to much existing work in another respect: we focus on NL-semantics-free, number-free ICL, as emphasized in Sec.~\ref{sec:sem-free}.

\subsection{Mechanistic Interpretation of ICL\label{sec:icldiscovery}}
The concept of in-context learning (ICL) was introduced by Brown et al.~(\citeyear{brown2020language}), who showed that their GPT-3 transformer could solve new tasks from a few examples in the prompt, without any gradient updating or fine-tuning. Most tasks depended on text meaning, although one task required repairing syntactic errors. The tasks did not include any cases of the pure symbolic manipulation of Templatic Generation, where the generated text is required to be analogous to an example contained in the prompt in that both were generable from a common template, inserting different text strings to the template's slots. 

In their quest for mechanistic interpretability of ICL, 
Elhage et al.~(\citeyear{elhage2021mathematical}) report the discovery of ``induction heads'' in small Transformer models with simplified architectures, with subsequent work finding similar mechanisms across a variety of Transformer architectures and parameter scales \cite{olsson2022context}.
These logical heads are actually specialized pairs of attention heads (in separate layers) that use past sequence pairs in the context to predict the successor to the current column: they predict that the symbol that will follow the most recently-generated symbol \sf{S} is the same as the symbol that followed the previous instance of \sf{S}.
The authors propose that induction heads might constitute the mechanism for the majority of all ICL in large transformer models; they present detailed (but indirect) evidence to support this hypothesis.
Our algorithm for Templatic Generation uses a procedure \CF\ that is identical to induction heads.
The abstract symbol-processing abilities of this algorithm, however, derive from a procedure \NF\ that is an abstraction of the induction head: it predicts that the \textit{abstract syntactic category} of the to-be-generated symbols is the same as the category that followed the previous instance of the category of the most recently-generated text.
Yang et al.~(\citeyear{yang2025emergent}) find ``symbolic induction heads'' that perform a similar function, predicting whether the role of the symbol that will follow \sf{S} is the same as the role of the symbol following the last occurrence of \sf{S}.

Other work in interpreting ICL has argued that LLMs' capacity to perform ICL is more attributable to ``function vector'' heads \cite{hendel2023context,todd2024function}, which transport a compact representation of ICL task mappings from example outputs to generated outputs. Such ``function vectors'' (activations of corresponding attention heads in the presence of ICL exemplars) can be used to induce the same task behaviors by inserting them into the forward pass of models in a variety of contexts, including zero-shot prompts containing only task inputs. 
Most recently, Yin \& Steinhardt~(\citeyear{yin2025which}) empirically demonstrate that function vector heads play a much larger role in ICL performance than induction heads, also showing that many induction heads from early stages of LLM pre-training eventually learn to perform as function vector heads instead.
In our algorithm, the ``function'' to be performed for a given prompt is conveyed by its template, which our PARSE algorithm identifies and represents in a vector encoding of an abstract parse structure; the heads that perform this are analogous to ``function vector'' heads.

Another relevant family of work in (mechanistic) interpretability concerns how pre-trained models learn to structure internal representations of discrete variables.
For instance, the \emph{linear representation hypothesis} (LRH) argues that models encode distinct variables in separate low-dimensional linear subspaces of the residual stream \cite{bolukbasi2016debiasing,alain2017understanding,vargas2020linear,geiger2024finding,makelov2024is}.\footnote{
    Note that there is considerable debate regarding the definition of the LRH --- e.g., it is often understood specifically in terms of 1-dimensional subspaces \cite{elhage2022superposition,park2024linear,engels2025not,lee2025evaluating}, or defined negatively as the belief that nonlinear methods are unnecessary for interpreting model representations \cite{pimentel2020information,pimentel2022attentional,davies2023competence,canby2024reliable,sutter2025non}.
    Here, we state the version of the hypothesis that is most relevant to TPF.
}
In DAT, we represent state structures as a concatenation of 1-hot vectors encoding the value taken by a given variable (\ref{ex:FLSSSemb}). As such, DAT fully satisfies the LRH, where the subspace representing each variable is simply the span of standard basis vectors from its first to final index in the concatenated vector (i.e., its register).

\subsection{The RASP Language for Programming Transformers \label{sec:progtransformers}}

Since our approach to understanding symbol-processing in transformers is to propose a higher-level language for programming them to do such processing, we focus the remainder of this discussion of related work on a particularly pertinent line of previous research, that based in the Restricted Access Sequence Processing language, RASP%
\footnote{Not to be confused with the random-access stored-program machine, also known as a RASP machine \cite{hartmanis1971computational}: a random-access-memory version of the Universal Turing Machine (Sec.~\ref{sec:DATUTM}).
}%
, proposed by Weiss et al.~(\citeyear{weiss2021rasp}).
For a more detailed discussion of RASP-based work, including an introduction to the language, see \RASPApp; for a wider discussion of related work, see \RelWkapp.

RASP-based work and TPF share the approach of deriving a high-level symbolic programming language which is compiled into a transformer network, with the goal of advancing mechanistic explanation of transformer computation.
Intuitively, however, on several conceptual dimensions the approaches differ. 

\begin{exe}
    \ex TPF vs.~RASP-based work
    \begin{xlist}
        \ex TPF development has focused on text generation, rather than the types of sequence-analysis functions centrally studied with RASP.%
        \footnote{
        Among the tasks that Weiss et al.~(\citeyear{weiss2021rasp}) provide RASP programs to solve is computing, for each input token \textsf{t}, the number of distinct token types in the string that occur with the same frequency in the string as does \textsf{t}, and Dyck-$k$, detecting balanced brackets in a string with \textit{k} distinct types of brackets.}
        Rather like the sequence-processing tasks studied with RASP, the PARSE program in TPF (Sec.~\ref{sec:parsalg}) analyzes the prompt by assigning abstract structural properties to symbols in the prompt, but these properties are not produced as output values per se; rather they are embedded as keys and values to drive attention during text generation by the GEN program (Sec.~\ref{sec:GenAlg}).
        \ex As in the implementation of RASP programs, in TPF the residual stream is decomposed into subspaces encoding the values of variables, but whereas the variables in the RASP work are associated to the nodes of a computation graph \cite[p.~5, point 5]{lindner2023tracr}, the variables in TPF are associated only to individual symbol positions in the prompt and have a declarative, rather than a procedural, character --- static properties of the symbols which are needed to support generation of new text.
        \ex The TGT requires identifying symbol sequences in the prompt as values of variables in a template, and moving them to new positions without alteration. Unlike for RASP, there is no use of MLPs in the current version of TPF because symbols are only copied, not modified --- not used to generate different symbols.
        \ex In RASP work, the matrix of attention values can be defined by invoking an arbitrary predicate relating the query- and key-positions. In TPF, attention is determined only by exactly matching values for variables specified in the query and key. Multiple variables are specified in an individual query or key, so attention can be conditioned on complex combinations of variable values; in RASP, attention is conditioned solely on pairwise relations between variables. A single PSL production can also write values into multiple variables, whereas each operation in RASP writes to only a single variable.
        \ex TPF provides an exact neural implementation of all of PSL, whereas a complete transformer implementation of RASP has not proved possible \cite{lindner2023tracr}.
        \ex Unlike for RASP, for TPF, publicly released software provides a comprehensive visualization tool for simultaneously viewing the program and network implementing it.
    \end{xlist}
\end{exe}

\section{Motivating a Case-Study of In-Context Learning  \label{sec:motiv}}
We are attempting to understand how neural networks can perform symbolic computation, but just what is `symbolic computation' in this context? To address this, this section follows the outline in  (\ref{ex:swapsecout}).

\begin{exe}
    \ex Section outline     \label{ex:swapsecout}
    \begin{xlist}
        \ex identify fundamental properties of symbolic computation [Sec.~\ref{sec:propSym}]
        \ex present an illustrative in-context learning task --- \swap\ --- that calls on the functionality expressed in these properties [Sec.~\ref{sec:case}]
        \ex preview the remainder of the paper, which presents: a general function class that \swap\ exemplifies; the symbolic languages PSL and QKVL for expressing algorithms to compute these functions at two different levels of description; and a compiler that translates these programs into a novel type of transformer neural network. [Sec.~\ref{sec:walk}]
    \end{xlist}
\end{exe}

\subsection{Fundamental Properties of Symbolic Computation  \label{sec:propSym} }
As in Box~\ref{box:ICL1} (\ref{ex:pr1} -- \ref{ex:temp1}), consider the following mapping, a simple instance of semantic parsing in which an English sentence in passive voice is mapped to a predicate-calculus-style logical form:

\begin{exe} 
  \ex  the program was translated by a compiler $\mapsto$ translated(a compiler, the program)   \label{ex:PLengl}
\end{exe}
This exemplifies a general schema or template (with variables in italics and constants in roman font):

\begin{exe}
    \ex \textit{x} was \textit{V} by \textit{y} $\mapsto$ \textit{V}(\textit{y},  \textit{x})   \label{ex:PLtemplate1}
\end{exe}

\noindent The variables take values that are \textit{strings} of symbols: this binding of values to variables is of particular interest because it has been argued to be a capability crucial for intelligence but beyond the capabilities of the simple neural architectures typical of cognitive and AI models prior to the advent of the transformer  (e.g., Marcus, \citeyear{marcus2001algebraic}; cf.~Smolensky \citeyear{smolensky1987analysis, smolensky1990tensor}).

Henceforth we drop the punctuation marks used above to aid human readability, so the template becomes 

\begin{exe}
    \ex \textit{x} was \textit{V} by \textit{y} $\mapsto$ \textit{V} \textit{y}  \textit{x}   \label{ex:PLtemplate}
\end{exe}
This can be cast in binary tree-to-tree form as an instance of a function we call \PL: 
\newpage

\begin{exe} 
     \ex \PL  \label{ex:PLtree}
\end{exe}
\hspace{6mm}
\begin{tabular}{c}
     \Tree [.S  \textit{x}  [.VP was [.VP \textit{V} [.PP by \textit{y} ]]]] \ \ \ $\mapsto$ \ \ \ \Tree [.LF \textit{V} [.args \textit{y} \textit{x} ]]
\end{tabular}    

\medskip
The symbols constituting the values of \textit{x, y} are themselves trees, moved as wholes from their positions in the input (syntactic) tree to their positions in the output (LF) tree.

In addition to variable binding, \PL\ displays a constellation (\ref{ex:props}) of crucial capabilities of symbol processing that are beyond the abilities of simple neural models
(Fodor and Pylyshyn \citeyear{fodor1988connectionism}, Marcus \citeyear{marcus2001algebraic}; 
cf.~Smolensky \citeyear{smolensky2012symbolic}).
We will see explicitly the mechanisms by which these abilities can in fact be naturally achieved in a class of transformer networks.
These properties collectively characterize much of the systematicity and compositionality that gives symbolic computation such great power for explaining --- and generating --- intelligent behavior (Box~\ref{box:comp}).

\begin{exe} 
    \ex Fundamental properties of symbolic computation \cite{newell1980physical} 
    \label{ex:props}
     \begin{xlist}
      \ex  Representations have\textit{ part/whole structure}: they are composed of \textit{constituents} which
      \begin{xlist}
         \ex  function as wholes in themselves for processing (e.g., they can, as wholes, be moved, copied, deleted, compared for equality); \label{ex:wholes}
         \ex  preserve their identities across different positions;  \label{ex:ident}
         \ex have \textit{types}: each constituent is a member --- a \textit{token} --- of a particular category of constituents. \label{ex:types}
      \end{xlist}
      \ex Representations typically have hierarchical structure: as a whole in its own right, a constituent may be composed of subconstituents. \label{ex:hier}
      \ex Representations have abstract roles that constituents fill: \label{ex:roles}
      \begin{xlist}

          \ex  the type of a constituent is characterized by a sequence of distinct \textit{roles} that its subconstituents \textit{fill} (e.g., the binary tree type can be specified with two roles: left-child, right-child); 
          \label{ex:roleset}
          \ex  a role is a \textit{variable}, and in a particular structural instance, it is \textit{bound} to a \textit{value}, its \textit{filler}: the constituent that fills it;
          \label{ex:varval}
          \ex  copying a constituent as a whole (\ref{ex:wholes}) entails copying its sequence of roles, each bound to the particular structure that fills it in the constituent token being copied.
          \label{ex:roleseq}
      \end{xlist}
      \ex Representations contain symbols: a constituent that has no subconstituent is an atomic element --- a \textit{symbol}, a token of its atomic-constituent type (e.g., in many grammar formalisms, a word's part-of-speech) \label{ex:symbols}
      \ex Representations may be \textit{recursive}: a constituent may have the same type as one of its subconstituents, or subsubconstituents, etc. (e.g., VP in (\ref{ex:PLtree})). \label{ex:recur}
      \ex Processing of representations: 
      \begin{xlist}
        \ex can include conditional process \textit{branching}, conditioned on a representation's structure, as well as its content;
        \label{ex:condBranch}
        \ex can include building a structure by \textit{binding} its roles (variables) to particular fillers (values); 
        \label{ex:binding}
        \ex can include extraction from a structure by \textit{unbinding} one of its roles, yielding that role's particular filler as output; 
        \label{ex:unbinding}
        \ex is \textit{compositional}: a constituent is processed by processing its subconstituents and combining the results into a new structure following a composition procedure determined by the type of the constituent.
        \label{ex:compos}
      \end{xlist}
    \end{xlist}
\end{exe}

Experimental evidence that transformers can implement these properties (\ref{ex:props}) is provided in Sec.~\ref{sec:tgtrel}, which shows that mappings like (\ref{ex:PLtemplate}) can be performed by language models such as GPT-4 with in-context learning: based only on its pretraining, given a prompt in question (Q)/answer (A) form like (\ref{ex:PLprompt}),

\begin{exe}
    \ex \PL-like GPT prompt \\
        \sf{\Q\ J was V by K \A\ V K J \Q\ B was V by C \A}
        \label{ex:PLprompt}
\end{exe}  

\noindent GPT-4 (gpt-4-0613, 2023) can correctly continue (\ref{ex:PLprompt}) with (\ref{ex:PLcont}):

\begin{exe}
    \ex \PL-like GPT continuation \\
     \color{orange}\textsf{V} \color{orange}\textsf{C} \color{orange}\textsf{B}
    \label{ex:PLcont}
\end{exe}  
\noindent (Throughout the paper we follow the convention of setting the prompt string in blue text and the continuation in orange text. This color contrast, and the special font used for \Q\ and \A\ here, are simply to aid the human reader; they are not part of the formalism of TPF that we are developing.)

Although examples of symbolic mappings are easier for \textit{us} to process when the symbols are words, and the strings are phrases, and punctuation characters are used (as in Box~\ref{box:ICL1}), we will focus on prompts in which symbols are typically simply individual characters, as in (\ref{ex:SWprompt}) below.
This is because in this work, as emphasized in Sec.~\ref{sec:sem-free},
we seek to understand the capabilities of transformer networks to perform pure symbol-manipulation tasks in which symbol meanings are irrelevant or non-existent.
Pretraining gives language models great facility with English, which can be exploited to complete prompts in English without necessarily behaving strictly on the basis of meaning-free patterns of symbols.
As noted above, completing \sf{\Q\ twice x \A\ x x \Q\ twice a b \A} with \squeeze{1.0}{\gsf{a b a b}}\ could reflect either knowledge of the semantics of English `twice' acquired in LM pretraining, or an ability to perform abstract, semantics-free templatic generation --- both are of considerable interest, but the work here is focused on the latter, as most other work on ICL assesses semantically-laden knowledge, and the present research is intended to be complementary to that large body of other work.

Recognizing the pattern in the prompt (\ref{ex:PLprompt}) as the instantiation of a template (\ref{ex:PLtemplate}) with variables $x$ and $y$ assigned values \sf{J} and \sf{K}, and then generating a continuation by reassigning the variables the new values \sf{B} and \sf{C}, is already a non-trivial instance of symbol processing.
A considerably more challenging class of such tasks allows the variables to take on values that are not just single symbols, but symbol strings: these must be parsed out of the input to identify the template structure; this was already the case in (\ref{ex:PLengl}), where each of the variables $x$ and $y$ in (\ref{ex:PLtemplate}) had two-symbol values: ``the program'' and ``a compiler'', respectively.
We will study prompts in which constituents take on values with variable numbers of symbols; this will be referred to as the `length' of constituents.
In addition, prompts will vary in the number of constituents they employ (number of variable `slots' in the templates).

\subsection{A Case Study: \swap \label{sec:case}}
An instance of this more challenging type of task is \swap, which will provide a primary case study for the remainder of the paper.
An illustrative prompt for \swap\ is given in (\ref{ex:SWprompt}).
The location of \sf{V} in the answer substring has now been shifted relative to \PL\ so that the pattern now more closely resembles the task \PA, where the passive form ``the program was translated by a compiler'' is mapped to the active form
``a compiler translated the program'': the linear positions of the subject and object have been swapped.

\begin{exe}
    \ex Instance of \swap \label{ex:SWinst}
        \begin{xlist} 
            \ex Prompt: \sf{ \Q\ B C V D E \A\ D E V B C \Q\ F G V J K L \A}
            \label{ex:SWprompt}
            \ex Continuation: 
             \color{orange}\textsf{J} \color{orange}\textsf{K} \color{orange}\textsf{L}
             \color{orange}\textsf{V} 
             \color{orange}\textsf{F} \color{orange}\textsf{G}
             \label{ex:SWcont}
        \end{xlist}
\end{exe}

The template for \swap\ is simply
\begin{exe}
    \ex \swap\ template:
        \sf{\Q} \textit{x} \sf{V} \textit{y} \sf{\A} \textit{y} \sf{V} \textit{x}
    \label{ex:SWtempl}
\end{exe}
\noindent The symbols \sf{\Q}, \sf{V}, and \sf{\A} function as fixed \textit{delimiters}, delimiting the strings providing the values of the variable \textit{constituents} (or `slots', or `arguments') $x$ and $y$. We will be studying templates with varying numbers of constituents: \swap\ has 2 (\ref{ex:SWtempl}).

To make examples more transparent to readers, we will typically follow the convention used in (\ref{ex:SWprompt}) according to which the value of a template slot (\textit{x} or \textit{y} here) is a string of individual characters in alphabetic sequence. 
The \swap\ task we study does not require this: aside from the reserved delimiter symbols \sf{\Q} and \sf{\A} which respectively initiate Question- and Answer-regions of the prompt, the identities of the individual symbols in the examples are arbitrary, of no relevance to the task; in particular, these symbols need not be single characters and could be words or non-alphanumeric symbols, as in (\ref{ex:pr1}), or nonsensical strings as in (\ref{ex:pr4}).

The prompts we study are a concatenation of two strings.
First, an initial question-answer \textit{example} string, which we label X --- starting at a prompt-initial token of the reserved symbol \sf{\Q} and terminating before a second token of \sf{\Q}.
Next is a \textit{continuation-cue} string, which we label C: this consists of a Question-region string followed by the reserved symbol \sf{\A}, a prompt for completing an Answer-region; the continuation-cue string starts at the prompt's second \sf{\Q} and continues through the prompt-final \sf{\A}.
Note that we will reserve the terms \textit{example} and (\textit{continuation-})\textit{cue} for this usage.
When training or testing models on ICL, we will refer to the entire input as a `prompt', reserving `example' for the portion of the initial portion of the prompt that instantiates the template driving the prompt's continuation.
(Thus training set size is measured by the number of `prompt/completion' pairs it contains, rather than the number of `examples'.)

For (\ref{ex:SWprompt}), this structure is shown in (\ref{ex:excont}) (and in more complete tree-form below in (\ref{ex:SWfullText})).

\begin{exe}
    \ex Example-Cue structure of (\ref{ex:SWinst})   \label{ex:excont}
      \begin{xlist} 
            \ex example X = \sf{ \Q\ B C V D E \A\ D E V B C}
            \label{ex:SWex}
            \ex cue C = \sf{\Q\ F G V J K L \A} 
            \label{ex:SWcontext}
      \end{xlist}
\end{exe}

The (prompt, continuation) pairs we study are the (input, output) pairs of some function in a class \F\ we define below in (\ref{ex:Fspec}).
In the following informal discussion, when we mention some structural property of the prompt or continuation, that property is imposed by the definition of \F. 

\subsection{Preview of the Symbolic Computation Implicit in the \swap\ Task   \label{sec:walk} }

\noindent How to rationalize the continuation in (\ref{ex:SWinst})? Intuitively, in the example that initiates the prompt, we recognize the Q-region (QR) substring \sf{\Q\ B C V D E} as instantiating the template \sf{\Q} \textit{x} \sf{V} \textit{y} of (\ref{ex:SWtempl}), and from this, recognize the A-region (AR) substring \sf{\A\ D E V B C} as instantiating \sf{\A} \textit{y} \sf{V} \textit{x}.
Then in the continuation cue, using the template given in the Q-region of the example, we recognize a new instance of the template in which \textit{x} now has value \sf{F G} while \textit{y} now has value \sf{J K L}. 
Inserting these values into the template for the example's A-region, \sf{\A} \textit{y} \sf{V} \textit{x}, determines that the continuation from the final \sf{\A} should be \color{orange}\textsf{ J K L V F G}\color{black}.

\begin{exe}
    \ex \swap\ instance (\ref{ex:SWinst}) full (\color{blue} prompt \color{black} + \color{orange}continuation \color{black}) structure: Target
    \label{ex:SWfullText} \\
        \footnotesize 
    \resizebox{\linewidth}{!}{%
    \Tree 
    [.S 
        [.X 
            [.QR 
                [.FQ 
                    [.\underline{v}_{FQ} 
                        [.\sf{Q} ]]]
                [.F1 
                    [.v_{F1} 
                        \sf{B} \sf{C} ] ] 
                [.FV  
                    [.\underline{v}_{FV} 
                        [.\sf{V} ]]] 
                [.F2  
                    [.v_{F2} 
                        \sf{D} \sf{E} ]]] 
            [.AR 
                [.FA 
                    [.\underline{v}_{FA} 
                        [.\sf{A} ]]]
                [.F2 
                    [.v_{F2} 
                        \sf{D} \sf{E} ] ] 
                [.FV  
                    [.\underline{v}_{FV} 
                        [.\sf{V} ]]] 
                [.F1  
                    [.v_{F1} 
                        \sf{B} \sf{C} ]]]] 
        [.C 
            [.QR 
                [.FQ 
                    [.\underline{u}_{FQ} 
                        [.\sf{Q} ]]]
                [.F1 
                    [.u_{F1} 
                        \sf{F} \sf{G} ] ] 
                [.FV  
                    [.\underline{u}_{FV} 
                        [.\sf{V} ]]] 
                [.F2  
                    [.u_{F2} 
                        \sf{J} \sf{K} \sf{L} ]]] 
            [.AR 
                [.FA 
                    [.\underline{u}_{FA} 
                        [.\sf{A} ]]]
                [.\color{orange}F2 
                    [.\color{orange}u_{F2} 
                        \color{orange}\textsf{J} \color{orange}\textsf{K} \color{orange}\textsf{L} ] ] 
                [.\color{orange}FV  
                    [.\color{orange}\underline{u}_{FV} 
                        [.\color{orange}\textsf{V} ]]] 
                [.\color{orange}F1  
                    [.\color{orange}u_{F1} 
                        \color{orange}\textsf{F} \color{orange}\textsf{G} ]]]]]
    }
\end{exe} 

Supporting this intuitive analysis is the structure in (\ref{ex:SWfullText}).
A walk-through of our algorithm for \swap, showing how this parse structure is produced and then utilized for generation, is provided in Appendix~\ref{sec:appWalk}; the full algorithm is written out in Appendix~\ref{sec:appPSLalg}.
The walk-through shows how this task is a strong test of symbolic computational capabilities as it calls on nearly all the properties of such computation listed in (\ref{ex:props}).

Our algorithm for the generation process calls for executing two procedures.
\NF\ determines which abstract position type in the template like (\ref{ex:SWtempl}) needs to be generated next: perhaps a constant delimiter symbol such as \sf{V}, or a variable slot such as \textit{x}. 
The continuation cue determines what string provides the value for a variable slot, and \NF\ looks this up and generates the first symbol in that string.
Then, if the string contains more than one symbol, \CF\ fills in the remaining symbols in the cued string.
Note that \CF\ is the functionality provided by an \textit{induction head} \cite{olsson2022context}, and \NF\ is a generalization of this functionality to abstract categories. 

\medskip

\noindent The algorithms referred to  here and given as informal walk-throughs in Appendix~\ref{sec:appWalk} will be presented formally below when we discuss the algorithmic level of TPF in Secs.~\ref{sec:algI} and \ref{sec:algII}.
But first we must formally present TPF at the highest level, the functional level.

\section{TPF, Functional Level. A Class of In-Context Learning Tasks: Templatic Generation \label{sec:fun}}
Appendix~\ref{sec:appWalk} shows that the \swap\ task implicitly incorporates 11 of the 13 properties of symbolic computation given in (\ref{ex:props}).%
\footnote{
Higher-level compositionality (\ref{ex:compos}) across multiple templates given through multiple examples in the prompt, and recursion (\ref{ex:recur}), are important properties to be incorporated in future work: see the preliminary discussion in Sec.~\ref{sec:compRec}.
}
This motivates the study of a class \F\ of ICL functions including \swap\ which we formalize in this section.
\F\ provides the functional-level description of the Transformer Production Framework TPF we now develop.
We call \F\ \textit{templatic generation} tasks: \textit{TGT}.
(Recall the examples given in Box \ref{box:ICL1}.)

\subsection{Templatic Generation Defined   \label{sec:Fdef} }
Each input-output mapping in \F\ defining an instance of our ICL templatic generation task is generated from an instance of the structural template in (\ref{ex:PRtree}).
\\
\begin{exe}
    \ex  The \color{blue} prompt\color{black}-\color{orange}continuation \color{black}\ (\color{blue} input\color{black}-\color{orange}output\color{black}) structure of templatic generation  \F
    \label{ex:PRtree} \\
    \scriptsize 
    \resizebox{\linewidth}{!}{%
    \Tree 
    [.S 
        [.X 
            [.QR 
                [.FQ 
                    [.\c{blue}{\underline{v}_{FQ}} ]]
                [.C_1 
                    [.\c{blue}{v_{C_1}} ] ] 
                [.D_1  
                    [.\c{blue}{\underline{v}_{D_1}} ]] 
                [.C_2  
                    [.\c{blue}{v_{C_2}} ]] 
                [....  
                    [.\c{blue}{v_{\cdots}} ]]] 
            [.AR 
                [.FA 
                    [.\c{blue}{\underline{v}_{FA}} ]]
                [.C^\prime_1 
                    [.\c{blue}{v_{C^\prime_1}} ] ] 
                [.D^\prime_1  
                    [.\c{blue}{\underline{v}_{D^\prime_1}} ]] 
                [.C^\prime_2  
                    [.\c{blue}{v_{C^\prime_2}} ]] 
                [....  
                    [.\c{blue}{v_{\cdots}} ]]]]
        [.C 
            [.QR 
                [.FQ 
                    [.\c{blue}{\underline{u}_{FQ}} ]]
                [.C_1 
                    [.\c{blue}{u_{C_1}} ] ] 
                [.D_1  
                    [.\c{blue}{\underline{u}_{D_1}} ]] 
                [.C_2  
                    [.\c{blue}{u_{C_2}} ]] 
                [....  
                    [.\c{blue}{u_{\cdots}} ]]] 
            [.AR 
                [.FA 
                    [.\c{blue}{\underline{u}_{FA}} ]]
                [.\color{orange}C^\prime_1 
                    [.\color{orange}u_{C^\prime_1} ] ] 
                [.\color{orange}D^\prime_1  
                    [.\color{orange}\underline{u}_{D^\prime_1} ]] 
                [.\color{orange}C^\prime_2  
                    [.\color{orange}u_{C^\prime_2} ]]
                [.\color{orange}...  
                    [.\color{orange}u_{\cdots} ]]]]]
    }
\end{exe}

The yield (sequence of terminal symbols) of the non-orange portion of (\ref{ex:PRtree}) is the \textit{prompt} $P$: the concatenation of the symbol sequences $\c{blue}{\underline{\rm{v}}_{FQ}\ \rm{v}_{C_1}\ \cdots\ \underline{\rm{u}}_{FA}}$.
This is the input sequence to the function $f \in$ \F\ being computed by our TPF system.
The output \g{$f(P)$}\ is the \textit{continuation} --- the yield of the orange portion: $\g{\rm{u}_{C^\prime_1}\ \underline{\rm{u}}_{D^\prime_1}\ \cdots }$.

(\ref{ex:Fspec}) specifies the tree template depicted in (\ref{ex:PRtree}). [In square brackets are comments concerning the motivation for some of the specifications.]
A formal grammar for the TGT is provided in Appendix~\ref{sec:tgtgrammar}.

\pagebreak

\begin{exe}
    \ex TPF at the functional level: the \c{blue}{prompt}-\g{continuation}\ (\c{blue}{input}-\g{output}) structure S 
    \label{ex:Fspec}
    \begin{xlist}
        \ex Region structure
        \begin{xlist}
            \ex S is a sequence of two subconstituents: the \textit{example} (X) followed by the \textit{continuation-cue} (C). [A characteristic of typical `1-shot' ICL tasks.]
            \label{ex:XC}
            \ex Each of X and C is a sequence of two subconstituents called \textit{regions}: the \textit{Q(uestion)-region} (QR) followed by the \textit{A(nswer)-region} (AR). 
        \end{xlist}
        \ex Field structure
        \begin{xlist}
            \ex Each region is a sequence of subconstituents called \textit{fields}.
            A field may not appear more than once in a region.
            The sequence of fields comprising a QR constituent is the same for the QR within X and for the QR within C; the same is true for the AR. 
            [This is the sense in which the continuation follows the template established by the example; the to-be-generated completion of the AR within C consists of the same sequence of fields as that of the AR within the example given in the prompt.]
            \ex Fields fall into two classes: \textit{delimiters} (D) and \textit{constituents} (C).    
            \ex The sequence of fields constituting a region alternate between D and C fields. \label{ex:alt}
            \ex The first field of a QR (resp. AR) is the delimiter field FQ (resp. FA).
\end{xlist}
        \ex Field values
        \begin{xlist}            
            \ex A field can be viewed as a variable which takes a symbol string as its \textit{value}. 
            Within any region, no symbol may appear more than once. 
            [A provisional assumption simplifying the matching and copying of symbol-string values between regions.]
            \ex Within X, we denote the value of any field F by v_F; within C, by u_F.
            If F is a delimiter field, we underline the value name (\underline{v} or \underline{u}).
            \ex Within S, a given delimiter field has a fixed value.
            \label{ex:delfix}
            [Together with (\ref{ex:alt}), this assures that the delimiters can be used to parse out the strings that are the varying constituent-field values.]
            The fixed value of FQ is always \sf{\Q}; of FA, \sf{\A}.
            \ex Within X, a given field type has a unique value (a symbol string) [i.e., the same in QR and AR].
        \end{xlist}
        \ex Field constraints
        \begin{xlist}
            \ex Within X, the constituent (non-fixed-value) fields comprising AR --- call them C^\prime_k --- are a subset of those comprising QR --- C_j. 
            Thus each C^\prime_k field in AR is identical to a corresponding field C_j in QR: C^\prime_k = C_j for some function $\varphi: k \mapsto j$.
            \ex It follows that v_{C^\prime_k} = v_{C_j} where $j = \varphi(k)$: the symbol string which is the value of C^\prime_k in AR is a repetition of the value in QR of  the corresponding C_j [it has been `copied' from position $j$ in QR to position $k$ in AR].
            \ex Note that $\varphi$ need not be onto: a C_j in QR may be absent in AR. [The symbol string v_{C_j} in the question QR has then been `deleted' from the answer AR.]
        \ex As within X, within C, a given field type has a unique value.
        The values of a given \textit{constituent}-field type in X and in C must be different (but from (\ref{ex:delfix}), the value of a given \textit{delimiter} field type is the same in X and C --- i.e., it is fixed throughout S).
        [It follows that the to-be-generated value of field \g{C^\prime_k}\ in AR of the completion C --- \g{u_{C^\prime_k}}\ --- is the given value of the field \c{blue}{C_j} within QR of C --- \c{blue}{u_{C_j}} --- for $j = \varphi(k)$.]
        \label{ex:Cuniqval}
        \end{xlist}
    \end{xlist}
\end{exe}

Here, and throughout the theory development in the text, we are considering `1-shot' ICL in which the prompt provides a single example X of the target input-output template. 
The `$k$-shot' case is the simple generalization in which the structure S has not just one, but a sequence of $k$ examples X preceding the cue constituent C. 
Unlike many tasks explored with ICL, in our task (which is semantics-free, purely symbolic: Sec.~\ref{sec:sem-free}), a single example uniquely determines the correct continuation: so in developing the theory, we put aside the additional complexities of multiple examples.
While logically unnecessary for a model that has mastery of the task, additional `shots' may be helpful for trained models: we explore this in \llmtesting.

\subsection{Relevance of the Templatic Generation Task for Symbolic Computation in Transformers\label{sec:tgtrel}}

We are studying how mechanisms within the transformer architecture enable advanced symbol processing, and we have seen through the case study of \swap\ how the Templatic Generation Task calls on most of the general capabilities involved in symbol processing.
But is this task, as formalized above, a task that transformer models can actually perform? 
In this section we examine this question with respect to both pre-trained language-model transformers and transformers trained from scratch to perform the task.

\subsubsection{The TGT Dataset\label{sec:tgtds}}
In order to assess how baseline models train and perform on templatic generation tasks, we created a synthetic dataset, generating prompts according to the TGT grammar and constraints outlined in Appendix~\ref{sec:tgtgrammar}. 
Each line in a dataset split file contains a prompt and the associated correct completion.
Note that, except where indicated otherwise, in this dataset, within a constituent, each symbol is a ``random-letter `word''' (`rlw') --- a random 2-letter sequence of lower-case letters; within delimiters, each symbol is an individual special character --- see (\ref{ex:pr4}) and illustrations below.
(Recall our semantics-free focus, Sec.\ref{sec:sem-free}.) This dataset is publicly available.\footnote{\url{https://huggingface.co/datasets/microsoft/templatic_generation_tasks}}

The dataset consists of the tasks and the splits shown in Table~\ref{tab:splits}.
Each line of a file containing a split contains a prompt/continuation pair generated by an associated prompt-template.  
The prompt-template defines the number of constituents and delimiters, and their order of appearance, in the \Q\ and \A\ of the examples given in the prompt. 
Within a split file, each prompt/continuation line has 2-3 parts, separated by a \verb|<tab>| character:

\begin{tabular}{@{}cl@{}}
     \textbf{x} & a prompt \\
     \textbf{y} & the correct completion \\
     info & optional text identifying the example type (for possible filtering during training)
\end{tabular}

`Echo' prompts are used to introduce out-of-distribution vocabulary symbols to the model (in the train split).  

\paragraph*{Echo prompt.}
\begin{verbatim}
    Q ZW A ZW . Q VI A  <tab>  VI . <tab>  {"type": "echo"}

    breakdown:
        1 example Q-region:   Q ZW
        1 example A-region:   A ZW .
        continuation-cue:     Q VI A
        target continuation:  VI .
        prompt info:          {"type": "echo"}
\end{verbatim}

\paragraph*{One-shot example.}
Here is a prompt/continuation line from the train split of the \texttt{1-shot\_rlw} task:
\begin{verbatim}
    Q oy xf kq be ` ? jp A jp = . Q jf ty zu np ` ? cx A  <tab>  cx = .  
             <tab>  {"cons_count": "Q2A1", "cons_len": "Q41.Q41"}

    breakdown:
        1 example Q-region:    Q oy xf kq be ` ? jp
        1 example A-region:    A jp = .
        continuation-cue:      Q jf ty zu np ` ? cx A
        target continuation:   cx = .
        prompt info:           {"cons_count": "Q2A1", "cons_len": "Q41.Q41"}
\end{verbatim}

\begin{table}[h]
    \centering
    \caption{TGT tasks and dataset splits} \label{tab:splits}
    \begin{tabular}{|l|l|}
    \hline
    \textbf{Task} & \textbf{Description} \\ \hline
    1\_shot\_rlw & each prompt is 1 example (\Q/\A, input/output, pair) + continuation-cue \\ \hline
    2\_shot\_rlw & each prompt is 2 examples + cue \\ \hline
    3\_shot\_rlw & each prompt is 3 examples + cue \\ \hline
    5\_shot\_rlw & each prompt is 5 examples + cue \\ \hline
    10\_shot\_rlw & each prompt is 10 examples + cue \\ \hline
    1\_shot\_eng & symbols in constituents are English words (vs.~2-letter random sequences) \\ \hline
    1\_shot\_rlw\_10x & same as 1\_shot\_rlw but with 10x as many training prompts \\ \hline
    \end{tabular}
\end{table}

\begin{table}[h]
    \centering
    \begin{tabular}{|l|l|}
        \hline
        \textbf{Split} & \textbf{Description} \\
        \hline
        train & contains 1, 2, or 4 constituents; each with 1, 2, or 4 symbols \\
        \hline
        dev & contains 1, 2, or 4 constituents; each with 1, 2, or 4 symbols \\
        \hline
        test & contains 1, 2, or 4 constituents; each with 1, 2, or 4 symbols \\
        \hline
        ood\_lexical & the constituent symbol vocabulary is absent from training examples \\
        & (UPPER CASE 2-letter random sequences) \\
        \hline
        ood\_cons\_len\_3 & all template constituent values contain 3 symbols \\
        \hline
        ood\_cons\_len\_5 & all template constituent values contain 5 symbols \\
        \hline
        ood\_cons\_len\_7 & all template constituent values contain 7 symbols \\
        \hline
        ood\_cons\_len\_10 & all template constituent values contain 10 symbols \\
        \hline
        ood\_cons\_count\_3 & all templates have 3 constituents \\
        \hline
        ood\_cons\_count\_5 & all templates have 5 constituents \\
        \hline
        ood\_cons\_count\_7 & all templates have 7 constituents \\
        \hline
        ood\_cons\_count\_10 & all templates have 10 constituents \\
        \hline
    \end{tabular}
\end{table}

In the prompt  string \textbf{x}, each example (`shot') begins with a ``\Q'', includes an ``\A'',  and ends with a period (``.'').  At the end of \textbf{x} is a continuation-cue  (beginning with a ``\Q'' and ending with an ``\A''), to be completed by the model.  The ground-truth completion is contained in the \textbf{y} string.

The above example was randomly generated using the following prompt template generated from the TGT grammar in Appendix~\ref{sec:tgtgrammar}:

\Q\  $\langle$\textit{constituent 1}$\rangle$ ‘ ? $\langle$\textit{constituent 2}$\rangle$ 
\newline
\indent \A\ $\langle$\textit{constituent 2}$\rangle$  = .

The default size of a training split in tasks is approximately 280,000 prompts.
The \texttt{1\_shot\_rlw\_10x} training set has about 2.8 million examples.

\subsubsection{Performance of Pre-Trained Language Models on Templatic Generation\label{sec:expts}}
For our LLM Transformer testing, we choose several popular models that we had access to through various API services.  We started by testing all of these models on the \texttt{1\_shot\_rlw} task of the TGT dataset, using the test split.

The LLMs were fed a prompt from our TGT dataset, prepended with a system prompt containing a basic instruction to complete the abstract pattern (see  Appendix~\ref{sec:tgtprompt}).
Each model was tested using the specified number of prompts shown in Table~\ref{tab:LLM}.

\begin{table}[h!]
    \centering    
    \caption{LLM Testing Summary} \label{tab:LLM}
    \begin{tabular}{|l|l|l|c|c|}
        \hline
        \textbf{Model} & \textbf{API Service} & \textbf{Test Date} & \textbf{Prompts} & \textbf{Accuracy} \\ \hline
        gpt-4 & OpenAI & Sep-23-2024 & 100 & 0.75 \\ \hline
        llama-3.1-405b & Together & Sep-26-2024 & 100 & 0.60 \\ \hline
        gpt-4o & OpenAI & Sep-23-2024 & 100 & 0.57 \\ \hline
        claude-3-opus & Anthropic & Sep-26-2024 & 100 & 0.48 \\ \hline
        gemini-1.5-pro & Google & Sep-26-2024 & 100 & 0.48 \\ \hline
        01-mini & OpenAI & Sep-23-2024 & 100 & 0.45 \\ \hline
        claude 3.5 sonnet & Anthropic & Sep-26-2024 & 100 & 0.38 \\ \hline
        gemini-1.5-flash & Google & Oct-06-2024 & 100 & 0.23 \\ \hline
        o1-preview & OpenAI & Oct-06-2024 & 100 & 0.18 \\ \hline
        gpt-4o-mini & OpenAI & Sep-23-2024 & 100 & 0.12 \\ \hline
        llama-3.1-70b & Together & Sep-26-2024 & 100 & 0.12 \\ \hline
        llama-3.1-8b & Together & Sep-26-2024 & 100 & 0.05 \\ \hline
     \end{tabular}
\end{table}

These results show that pretrained transformer LMs can perform the Templatic Generation Task to varying degrees, but not even the best model exceeded 75\% accuracy.  This speaks to the relevance of TGT for understanding the symbol processing capabilities of pretrained transformer LLMs.
The capability to perform TGT is present in LMs in some form, but particularly as the templates become larger and the number of symbols in the constituents grow, they struggle: there is certainly room for strengthening this capability in future work drawing on the insights from TPF: see Sec.~\ref{sec:enhancingTf}.
From the initial tests shown in Table~\ref{tab:LLM}, we chose the best performing model, GPT-4, and then did a series of exploratory tests to see how the performance varied under different task variants.  See \llmtesting\ for details.

\subsubsection{Training Models on Templatic Generation\label{sec:tgttraining}}
Given our LLM testing, we  know that pre-trained LLMs have the ability to solve the templatic generation tasks to varying degrees, depending on the model.  Can the ability to solve these tasks be learned from scratch, even by smaller models?  To find out, we trained from scratch on the TGT 1-shot task 6 basic sequence-to-sequence models listed in \trainscratch.  
The results are presented in Table~\ref{tab:1shot}.
The case of `OOD Lexical' was already illustrated: 2-letter random uppercase symbols (RLW) were used as the values of constituents, having been seen in the training set only in single-symbol `echo' prompts.
The `OOD ConLen 7' test used prompts containing 1, 2 or 4 constituents, each comprising a length-7 (rlw) symbol string: the models saw only constituent lengths of 1, 2, or 4 in training.
The `OOD ConCnt7' test used prompts containing 7 constituents (each comprised of 1, 2 or 4 symbols), while the training set only contained prompts with 1, 2 or 4 constituents.

\begin{table}[h!]
\centering
\caption{Results for \texttt{1\_shot\_rlw} task} \label{tab:1shot}
\begin{tabular}{lccccc}
\toprule
\textbf{Model} & \textbf{Train} & \textbf{Dev} & \textbf{OOD} & \textbf{OOD} & \textbf{OOD} \\
 & \textbf{Acc} & \textbf{Acc} & \textbf{Lexical} & \textbf{ConLen 7} & \textbf{ConCnt 7} \\
\midrule
transformer        & 0.9838 & 0.8568 & 0.0052 & 0.6344 & 0.1828 \\
nano\_gpt          & 0.9997 & 0.9997 & 0.0074 & 0.6908 & 0.2247 \\
nano\_gpt\_attn\_only & 0.9992 & 0.9989 & 0.0070 & 0.7118 & 0.3346 \\
cnn               & 0.6284 & 0.4766 & 0.0000 & 0.0062 & 0.0025 \\
lstm\_attn        & 0.6995 & 0.6432 & 0.0000 & 0.0000 & 0.0069 \\
mamba             & 0.9980 & 0.9136 & 0.0137 & 0.0000 & 0.0739 \\
\bottomrule
\end{tabular}
\end{table}

These results show that transformers can learn to perform the in-distribution TGT tasks directly, without LM training; other neural architectures, CNNs and GRUs, struggle, although Mamba succeeds.  

All models exhibit poor OOD lexical generalization, indicating that the knowledge acquired during training is not of an abstract-pattern-based nature but instead tied rather strongly to particular symbols seen. 

On OOD generalization in the number of constituents (`ConCnt 7'), only transformers achieve modest success. 
The transformers' OOD generalization is stronger in the length of constituents (`ConLen 7'), which is intuitively easier than increasing the number of constituents in the template that must be extracted from the prompt and suitably arranged.

The superiority of transformers at OOD generalization is consistent with the hypothesis that within the transformer architecture there are internal mechanisms that facilitate templatic generation; the work here provides detailed and comprehensive hypotheses about what exactly those mechanisms may be (Sec.~\ref{sec:LMs}) as well as suggestions for how the standard transformer architecture can be enhanced to improve OOD generalization (Sec.~\ref{sec:enhancingTf}).
The modified-transformer model we design below achieves 100\% on all OOD tests (Sec.~\ref{sec:exptsDAT}).

To further understand how altering the TGT task would affect the performance of these models, we also trained them on the other tasks in the TGT dataset.  See \trainscratch\ for details.

\section{TPF, Higher Algorithmic Level: The Production System Machine\label{sec:algI}}
Having documented the partial success achieved by transformers on the task defined by the functional level description of TPF systems presented in Sec.~\ref{sec:fun} --- the class \F\ of templatic generation functions instantiated in the TGT dataset --- we now descend to the algorithmic level, which actually contains two sub-levels that are formalized as two symbolic abstract machines that compute the functions in \F.
Closest to the functional level is the Production System Machine PSM, which we present in this section.
Beneath that, closer to the implementation level, is the QKV Machine, presented in Sec.~\ref{sec:algII}: like PSM, this is a symbolic abstract machine, but it can be directly implemented in a type of discrete-attention transformer network --- DAT --- defined in Sec.~\ref{sec:impl}.
In Sec.~\ref{sec:algII} we describe a compiler that takes a program for the PSM, written in the Production System Language PSL introduced next, and translates it to an equivalent program in QKVL, a language for expressing programs for the QKVM.
In Sec.~\ref{sec:impl} we then describe a second compiler that translates QKVL programs into an equivalent neural network with isomorphic states and state dynamics.
It is only at this lowest level that numeric neural computation appears. 

\subsection{The PSM Architecture}
The symbolic processing in both PSM and QKVM consists of two phases: prompt processing --- i.e., parsing --- followed by continuation generation. 
These phases are described in general terms for PSM in (\ref{ex:PSMparsing}) and (\ref{ex:PSMgen}).
\\
\begin{exe}
    \ex Production-System Machine state dynamics: Parallel processing of the prompt \label{ex:PSMparsing}
    \begin{xlist}
        \ex Machine states
        \begin{xlist}
            \ex The state of the Production-System Machine is a sequence of cell states.
            \ex A cell is identified by its value of the variable \textsf{position} (or $p$), a natural number.
            \ex A prompt containing $P$ symbols is encoded in the cells in positions $1$ through $P$, the \textit{prompt cells}; the completion will be encoded in the subsequent \textit{completion cells}. 
        \end{xlist}
        \ex Cell-state structure
        \begin{xlist}
            \ex Each cell has a \textit{state} which is characterized by the values of a set of \textit{state variables}, including \textsf{position} ($p$) and \textsf{symbol} ($s$).
            The possible values of each state variable form a discrete set.
            \ex $s[m]$ is the type of the symbol encoded in the cell with $p = m$.
            \ex Other state variables are introduced below. Some of them are used to encode the parse tree structure (\ref{ex:PRtree}), as visualized in (\ref{ex:PRmatrix}) --- these are the \textit{structural variables}: \textsf{region} ($r$), \textsf{field} ($f$) and \textsf{index} ($d$). 
            \ex The state of a cell is a \textit{state structure} encoding the values of the state variables for that cell; some variables may have the null value \textsf{nil}.
            The space of all possible state structures is the \textit{state-structure space} SSS.
            \ex \textit{Notation.} For any state structure $\textsf{S} \in$ SSS, let the value of state variable \textit{x} in $\textsf{S}$ be denoted $\textsf{S}.x$.
            When $\textsf{S}$ is understood, we abbreviate $\textsf{S}.x = a$ to $x:a$. 
            If $\textsf{S}$ has, say, two state variables with non-null values, $\textsf{S}.x = a$ and $\textsf{S}.y = b$, we abbreviate $\textsf{S}$ itself to $x:a, \; y:b$.
        \end{xlist}
        \ex Layer structure
            \begin{xlist}
            \ex The dynamics of the cells is unrolled in time, so that for each step of computation there is a \textit{layer} of cells.
            \ex At layer 1, each cell's state includes its \textsf{position} value and a value for \textsf{symbol} that is supplied by the prompt. \label{ex:promptCells}
            \ex (\ref{ex:PRmatrix}) depicts the state of the machine at a single time step, i.e., a single layer.
            Each state variable is represented as a row containing the values of the variable across the cells in that layer.
            The state structure for the first cell is shown in (\ref{ex:statestruc}).
        \end{xlist}
        \ex Production System Language, PSL
        \label{ex:PSL}
        \begin{xlist}
            \ex A particular PS Machine is specified by a program in the language PSL: this is a sequence of $L$ productions, with production $\ell$ specifying the updating process for layer $\ell$.  
            Each layer is updated according to a single production.
            \ex The prompt cells in a given layer are all updated in parallel (i.e., not autoregressively) by the production corresponding to that layer. 
            \ex Production $\ell$ is specified in PSL by: (i) a \textit{Condition}, which defines requirements on the values of state variables in layer $\ell$ cells in order to allow the production to execute; and (ii) an \textit{Action}, which assigns values to state variables of cells to be set in layer $\ell+1$.
            \ex Production Conditions and Actions deploy two meta-variables $n$ and $N$. 
            \label{ex:nN}
            \begin{itemize} 
                 \item[\ding{192}] When production $\ell$ executes, for each cell position $N$ it sets values for state variables in cell $N$ in layer $\ell+1$, using the values of variables in cell $n$ of layer $\ell$ (where possibly $n = N$).
                \item[\ding{193}] A production can only execute when $n$ and $N$ satisfy its Condition.
                There is no requirement of causal interaction, $n < N$ [but see (\ref{ex:causal})].
                \item[\ding{194}] When a production executes to update a cell $N$, if multiple positions $n$ satisfy the production's Condition, the least value of $n$ is used [but see (\ref{ex:rmost})].
                If no position $n$ satisfies the Condition, no Action is taken.
                \item[\ding{195}] If Production $\ell$ does not assign a value to a given state variable in cell $N$, that variable has the same value in layer $\ell+1$ as it did in layer $\ell$.
            \end{itemize}
            \ex A sub-sequence of productions can constitute a \textit{repeat block}, in which case the layers corresponding to the productions in that block are evaluated in sequence repeatedly until a \textit{termination condition} specified for that block is met (e.g., no further changes in state variables).
            
        \end{xlist}
    \end{xlist}
\end{exe}

\begin{exe}
    \ex Production-System Machine state dynamics: Autoregressive generation of the continuation   \label{ex:PSMgen}
    \begin{xlist}
        \ex Given a prompt of length $P$, the level-1 states of cells $1, ..., P$ are determined by the prompt (\ref{ex:promptCells}). 
        These are processed in parallel according to (\ref{ex:PSMparsing}).
        The level-1 state of cell $P+1$ is set equal to the level-$L$ state of cell $P$ (except that $p := P + 1$). [Note that, except for $p$, the entire state structure ---  not just the value of \textsf{symbol} --- is copied from the final state of cell $P$ to the initial state of cell $P + 1$.
        The responsibility of cell $P$ is to compute the values of \textit{all} state variables (except $p$) for the next cell.]
        \label{ex:PSMgen1}
        \ex The states of the continuation cells following the \textit{P} prompt cells are updated in sequence: first, the position-\textit{P}+1 cell is updated repeatedly, through all \textit{L} layers of the machine corresponding to the $L$ productions of the program; these layers operate just as they do for the prompt cells.
        \ex The level-1 state of cell $P+2$ is set equal to the level-\textit{L} state of cell $P+1$ (except that $p := P + 2$), and then cell $P+2$ is processed through all $L$ layers. This process iterates until a termination condition is met (e.g., the generation of a special termination symbol).
        \ex The sequence of level-1 values of the variable \textsf{symbol} for the continuation cells constitute the continuation string: the output of the function being computed.
    \end{xlist}
\end{exe}

\subsection{\swap\ in the PSM   \label{sec:SWPSM} }
We start with the encoding of the hierarchical parse structure (\ref{ex:SWfullText}) in the PSM.
The algorithm we develop does not use the highest-level constituents X and C per se, only their particular subconstituents, so we flatten the tree by naming the QR within X `XQ', the AR within X `XA', and similarly naming the subconstituents of C `CQ' and `CA'.
The prompt portion of (\ref{ex:SWfullText}) --- excluding the continuation and the value nodes $v_i, u_i$ which are no longer useful --- is now (\ref{ex:SWflat}).

\begin{exe}
    \ex \swap\ instance (\ref{ex:SWinst}) (\color{blue} prompt\color{black}) structure
    \label{ex:SWflat} \\
    \footnotesize 
    \Tree 
    [.\color{white}S\color{black}
             [.XQ 
                [.FQ 
                        [.\sf{Q} ]]
                [.F1 
                        \sf{B} \sf{C} ] 
                [.FV  
                        [.\sf{V} ]] 
                [.F2  
                        \sf{D} \sf{E} ]]
            [.XA 
                [.FA 
                        [.\sf{A} ]]
                [.F2 
                        \sf{D} \sf{E} ]  
                [.FV  
                        [.\sf{V} ]] 
                [.F1  
                        \sf{B} \sf{C} ]]
            [.CQ 
                [.FQ 
                        [.\sf{Q} ]]
                [.F1 
                        \sf{F} \sf{G} ]  
                [.FV  
                        [.\sf{V} ]] 
                [.F2  
                        \sf{J} \sf{K} \sf{L} ]] 
            [.CA 
                [.FA 
                        [.\sf{A} ]]]
     ]
\end{exe}

In the PSM, the hierarchical structure (\ref{ex:SWflat}) is encoded as in (\ref{ex:PRmatrix}), with each row showing the values of the state variable named at the left.
The state structure of the first cell is shown in (\ref{ex:statestruc}).

\pagebreak

\begin{exe}
    \ex PSM representation of the parse (\ref{ex:SWflat}) of the prompt (\ref{ex:SWprompt})
    \label{ex:PRmatrix}
\end{exe}
\begin{center}
\tiny
\renewcommand{\arraystretch}{1.5}
\addtolength{\tabcolsep}{-2pt}
\resizebox{\linewidth}{!}{%
    \begin{tabular}{l | *{26}{c|}}
    \cline{2-26}
    \textit{r}
    &XQ &XQ &XQ &XQ &XQ &XQ &XA &XA &XA &XA &XA &XA &CQ &CQ &CQ &CQ &CQ &CQ &CQ &CA &   &   &   &   &    \\
    \cline{2-26}
    \textit{f}
    &FQ &F1 &F1 &FV &F2 &F2 &FA &F2 &F2 &FV &F1 &F1 &FQ &F1 &F1 &FV &F2 &F2 &F2 &FA &   &   &   &   &    \\
    \cline{2-26} 
    \textit{s}
    &\sf{Q} &\sf{B} &\sf{C} &\sf{V} &\sf{D} &\sf{E} &\sf{A} &\sf{D} &\sf{E} &\sf{V} &\sf{B} &\sf{C} &\sf{Q} &\sf{F} &\sf{G} &\sf{V} &\sf{J} &\sf{K} &\sf{L} &\sf{A} &   &   &  &   &    \\
    \cline{2-26}
    \textit{p}
    &1  &2  &3  &4  &5  &6  &7  &8  &9  &10 &11 &12 &13 &14 &15 &16 &17 &18 &19 &20 &21 &22 &23 &24 &25  \\
    \cline{2-26}
    \textit{d}
    &0  &0  &1  &0  &0  &1  &0  &0  &1  &0  &0  &1  &0  &0  &1  &0 &0  &1  &2  &0  &   &   &   &   &    \\
    \cline{2-26}
    \end{tabular}
}
\end{center}

\begin{exe}
    \ex The state structure for the position-1 cell in (\ref{ex:PRmatrix}). ($\; \vdots$ stands in for other state variables not shown in (\ref{ex:PRmatrix}).)
    \label{ex:statestruc}
    
    \begin{tabular}{|r<{:} l|}
        \hline
        $r$ & XQ \\
        $f$ & FQ \\
        $s$ & \sf{Q} \\
        $p$ & 1 \\
        $d$ & 0 \\
        \ldots & \ $\vdots$   \\
        \hline
    \end{tabular}
\end{exe}

The hierarchical structure encoded \textit{explicitly} in the tree (\ref{ex:SWflat}) is now encoded \textit{implicitly} in the spans of the values of the \textit{structural variables}: that the substring \sf{B C} is the value of a field constituent (of type F1) is encoded by the \textit{field variable} \textsf{field} (or \textit{f}) having the same value (F1) for both \sf{B} and \sf{C} (and a different value for the preceding and following symbols).
Similarly, that the prompt prefix \sf{Q A B V C D} is the value of a region constituent is encoded by the \textit{region variable} \textsf{region} (or \textit{r}) having the same value, XQ, for all the symbols in that prefix.
When implemented below in a transformer, this is the method for encoding hierarchy proposed in Hinton's  (\citeyear{hinton2023represent}) GLOM model.
(Extending this scheme to enable recursive structure faces a number of challenges, but see Sec.~\ref{sec:compRec}.)

The final structural variable, \textsf{index} ($d$), denotes the position of each symbol within its string (\ref{ex:index}).

\begin{exe}
    \ex
    \label{ex:index}
    \begin{xlist}
        \ex The values of \textsf{index} --- the string-internal positions of field values --- are shown as $0, 1, 2, ...$ in (\ref{ex:PRmatrix}) (see \sf{J K L}).
        \ex However, our algorithms will only need the binary distinction between field-initial ($d = 0$) and non-field-initial, $d \neq 0$; therefore, henceforth we will simply index all non-field-initial symbols with $d = 1$ --- e.g., see \sf{J K L} in (\ref{ex:CFmatrix}).
    \end{xlist}
\end{exe}

\noindent Our algorithm \PG\ for solving TGT consists of (i) a parsing algorithm PARSE to generate the encoded hierarchical structure of the input prompt in (\ref{ex:PRmatrix}), and (ii)  a generation algorithm GEN to use this parse to produce the output continuation string: these are respectively presented in Secs.~\ref{sec:PARSE} and \ref{sec:GENPSL}.
Although logically prior, because of its complexity, we postpone discussion of the parsing algorithm until after we present the generation algorithm.
\subsection{Generation Algorithm GEN in the Production-System Language PSL   \label{sec:GENPSL}  }
In Sec.~\ref{sec:GenAlg} we illustrate through informal walk-throughs the two operations used during generation of the continuation string: \CF\ and \NF.
Which of these is executed at a particular step depends on whether the most-recently-generated symbol is final in its field (calling for \NF) or not (calling for \CF).

\subsubsection{\CF\label{sec:contfield} }
We discuss \CF\ first because it is simpler than \NF, although we need to jump ahead a step into the generation process to reach a point where \CF\ is the appropriate operation.
So suppose the first continuation symbol \gsf{J}\ has just been generated (along with its structural variable values $f=$ F2, $r=$ CA, $d = 0$).
\gsf{J}\ is not the final symbol in its field F2, so the next symbol must be generated by \CF, which performs the actions in (\ref{ex:CFex}); see the visualization in (\ref{ex:CFmatrix}).
This operation is justified by (\ref{ex:Cuniqval}), which requires that --- in the notation of (\ref{ex:SWfullText}) --- \g{$\rm{u}_{\rm{F2}}$}, the value of field F2 within CA, must equal u$_{\rm{F2}}$, the value of field F2 within CQ.

\pagebreak

\begin{exe}
    \ex Use of \CF\ to continue generating the current field's value string
    \label{ex:CFmatrix}
\end{exe}
\begin{center}
\tiny
\renewcommand{\arraystretch}{1.5}
\addtolength{\tabcolsep}{-2pt}
\resizebox{\linewidth}{!}{%
    \begin{tabular}{l | *{26}{c|}}
    \cline{2-26}
    \textit{r}
    &XQ &XQ &XQ &XQ &XQ &XQ &XA &XA &XA &XA &XA &XA &CQ &CQ &CQ &CQ &CQ &CQ &CQ &CA &CA   &\color{gray}CA  &   &   &    \\
    \cline{2-26}
    \textit{f}
    &FQ &F1 &F1 &FV &F2 &F2 &FA &F2 &F2 &FV &F1 &F1 &FQ &F1 &F1 &FV &F2 &F2 &F2 &FA &F2   &\color{gray}F2   &   &   &    \\
    \cline{2-26} 
    \textit{s}
    &\sf{Q} &\sf{B} &\sf{C} &\sf{V} &\sf{D} &\sf{E} &\sf{A} &\sf{D} &\sf{E} &\sf{V} &\sf{B} &\sf{C} &\sf{Q} &\sf{F} &\sf{G} &\sf{V} &\c{purple}{\textsf{J}} &\c{red}{\textsf{K}} &\sf{L} &\sf{A} &\gsf{J}  &\color{gray}\textsf{K}   &  &   &    \\
    \cline{2-26}
    \textit{p}
    &1  &2  &3  &4  &5  &6  &7  &8  &9  &10 &11 &12 &13 &14 &15 &16 &17 &18 &19 &20 &21 &22 &23 &24 &25  \\
    \cline{2-26}
    \textit{d}
    &0  &0  &1  &0  &0  &1  &0  &0  &1  &0  &0  &1  &0  &0  &1  &0 &0  &1  &1  &0  &0   &\c{gray}1   &   &   &    \\
    \cline{2-26}
    &   &   &   &   &   &   &   &   &   &   &   &   &   &   &   &   &\c{purple}{$n_0$}&\c{red}{$n$}&  &   &$N$ &   &   &   &    \\
    \end{tabular}
}
\end{center}

\begin{exe}
    \ex \CF\ in action
    \label{ex:CFex}
    \begin{xlist}
        \ex Current most-recently-generated symbol: \gsf{J}\ in position $N = 21$; $s[N] =$ \gsf{J}
        \ex Match to symbol type \gsf{J}\ in CQ: position \c{purple}{$n_0$} $= 17$; $s[n_0] = s[N]$ (where $n_0$ is not field-final)
        \ex Next position: \c{red}{$n$} $= n_0 + 1 = 18$
        \ex Symbol at position $n$: $s[n] =$ \gsf{K}
        \ex Update \textsf{symbol}[$N$] to \gsf{K}\ for subsequent propagation (\ref{ex:PSMgen1}) to the next position $N + 1 = 22$: set $s[N] = s[n] =$ \gsf{K}
        \ex Update structural-variable values at position $N$ for subsequent propagation to position $N+1$: set $f[N] = f[n] =$ F2; $r[N] =$ CA, $d = 1$
    \end{xlist}
\end{exe}

Since $n_0 = n - 1$, (\ref{ex:CFex}) can be compactly expressed as the condition-action production rule (\ref{ex:CFprod}).

\begin{exe}
    \ex \CF\ as a production operating on state variables 
    \label{ex:CFprod}
    \begin{xlist}
        \ex Condition: $n, N$ satisfy $s[n-1] == s[N]$ and $r[n-1] ==$ CQ \\ 
        (where $n-1$ is not field-final) 
        \label{ex:CFprodC}
        \ex Action: set $s[N] := s[n]$; $f[N] := f[n]$; $r[N] :=$ CA, $d[N] := 1$
        \label{ex:CFprodA}
   \end{xlist} 
\end{exe}
As pointed out above, this describes the effect of a transformer `induction head' \cite{olsson2022context}.

Note that the condition that $n-1$ not be a field-final position is equivalent to the condition that $n$ not be field-initial; this can be simply expressed as $d(n) = 1$: recall that within the symbol strings that are values of fields, the index 0 labels the initial symbol, with index 1 labelling the non-initial symbols (\ref{ex:CFmatrix}).
The production (\ref{ex:CFprod}) copies a symbol from within the same field-value string as the symbol in CQ that matches the current symbol $s[N]$ in CA: this copied symbol cannot be field-initial, so it necessarily has $d = 1$.

The production (\ref{ex:CFprod}) makes reference to $s[n - 1]$ in its condition, and $s[n]$ in its action; these symbols are used to update \textsf{symbol}[$N$] to then be used to generate $s[N+1]$.
Throughout the paper, $N$ will always denote the position being updated by a production, with $n$ denoting a position containing information needed to perform the update (with possibly $n = N$).
Note that a given production applies in parallel to update all positions $N$, and for each $N$, the relevant corresponding position $n$ is determined independently of the other positions being updated.

In the transformer implementation below, position $N$ will attend to position $n$ to retrieve the information needed to generate the next symbol.
Rather than having to attend additionally to position $n-1$ to implement the condition (\ref{ex:CFprodC}), it is convenient to localize all the necessary information in one position, $n$. 
To do this we introduce the auxiliary state variable $s^*$, setting $s^*[n] = s[n-1]$; the condition `$s[n-1] == s[N]$' now becomes `$s^*[n] == s[N]$' and now all needed information can be gathered at $n$.

More generally, for each state variable $x$ we will define the related variable \textsf{prev\_x}, $x^*$ for short, defined by $x^*[n] = x[n-1]$.
Further auxiliary variables will also be needed; corresponding to state variable $x$ will be \textit{x}\bp, assigned values by the action of productions.

When a production executes at step $\ell$ of the computation, the production reads values of state variables at step $\ell$ and the Action writes the values of variables at step $\ell + 1$.
Together with the use of the auxiliary state variables introduced above, this means that the production (\ref{ex:CFprod}) can be written in what we'll call its \textit{Transformed} notation.
The Transformed version of the production (\ref{ex:CFprod}) can be written solely in terms of $N$ (which, in the transformer implementation, will issue the appropriate query) and $n$ (which will carry the appropriate matching key) (\ref{ex:CFprod***}).

\begin{exe}
    \ex \CF\ as a Transformed production  
    \label{ex:CFprod***}
    \begin{xlist}
        \ex Condition: $n, N$ satisfy $s^{*(\ell)}[n] == s^{(\ell)}[N]$, $r^{*(\ell)}[n] ==$ CQ, $d^{(\ell)}[n] == 1$ 
        \ex \squeeze{.95}{Action: set $s^{(\ell + 1)}[N] := s^{(\ell)}[n]$, $f^{(\ell + 1)}[N] := f^{(\ell)}[n]$, $r^{(\ell + 1)}[N] :=$ CA, $d^{(\ell + 1)}[N] := 1$}
   \end{xlist} 
\end{exe}

Since, for step $\ell$, variable values are always read from the step-$\ell$ state structure and written to the step-($\ell + 1$) state structure, these step values can be left implicit; (\ref{ex:CFprod***}) can be written in the abbreviated form (\ref{ex:CFprod*}). 

\begin{exe}
    \ex \CF\ as a Transformed production, abbreviated: \textsf{ContField}  
    \label{ex:CFprod*}
    \begin{xlist}
        \ex Condition: $n, N$ satisfy $s^*[n] == s[N]$, $r^*[n] ==$ CQ, $d[n] = 1$ 
        \ex Action: set $s[N] := s[n]$, $f[N] := f[n]$, $r[N] :=$ CA, $d[N] := 1$
   \end{xlist} 
\end{exe}

\subsubsection{\NF}\label{sec:nextfield} 
Because the newly-generated symbol \sf{K} does not complete its field F2, the next step of generating the continuation also needs \CF; this generates the next symbol \sf{L} (along with its structural-variables' values).

\sf{L} completes the F2 field, so generating the following symbol, \sf{V}, requires \NF.
Its action is spelled out in (\ref{ex:NFex}), visually supported by (\ref{ex:NFmatrix}).

\begin{exe}
    \ex Use of \NF\ to start the generation of the next field's value-string
    \label{ex:NFmatrix}
\end{exe}
\begin{center}
\tiny
\renewcommand{\arraystretch}{1.5}
\addtolength{\tabcolsep}{-2pt}
\resizebox{\linewidth}{!}{%
\begin{tabular}{l | *{26}{c|}}
\cline{2-26}
\textit{r}
&XQ &XQ &XQ &XQ &XQ &XQ &XA &XA &XA &XA &XA &XA &CQ &CQ &CQ &CQ &CQ &CQ &CQ &CA &CA   &CA  &CA   &\color{gray}\textsf{CA}   &    \\
\cline{2-26}
\textit{f}
&FQ &F1 &F1 &FV &F2 &F2 &FA &F2 &\color{purple}F2 &\c{red}{FV} &F1 &F1 &FQ &F1 &F1 &\c{red}{FV} &F2 &F2 &F2 &FA &F2   &F2   &\color{purple}F2   &\color{gray}\textsf{FV}   &    \\
\cline{2-26} 
\textit{s}
&\sf{\Q} &\sf{B} &\sf{C} &\sf{V} &\sf{D} &\sf{E} &\sf{\A} &\sf{D} &\sf{E} &\sf{V} &\sf{B} &\sf{C} &\sf{\Q} &\sf{F} &\sf{G} &\c{red}{\textsf{V}} &\sf{J} &\sf{K} &\sf{L} &\sf{\A} &\gsf{J}  &\gsf{K}   & \gsf{L} & \color{gray}\textsf{V}  &    \\
\cline{2-26}
\textit{p}
&1  &2  &3  &4  &5  &6  &7  &8  &9  &10 &11 &12 &13 &14 &15 &16 &17 &18 &19 &20 &21 &22 &23 &24 &25  \\
\cline{2-26}
\textit{d}
&0  &0  &1  &0  &0  &1  &0  &0  &1  &0  &0  &1  &0  &0  &1  &0 &0  &1  &1  &0  &0   &1   &1   &\c{gray}0   &    \\
\cline{2-26}
&   &   &   &   &   &   &   &   &$\color{purple}n_1\color{black}$   &\c{magenta}{$n_2$}   &   &   &   &   &   &\color{red}$n$\color{black}   &  &  &  &   &   &   &$N$   &   &    \\
\end{tabular}
}
\end{center}

\begin{exe}
    \ex \NF\ in action
    \label{ex:NFex}
    \begin{xlist}
        \ex Current most-recently-generated symbol: \gsf{L}\ in position $N = 23$; $s[N] =$ \gsf{L}, with field $f[N] =$ \c{purple}{F2}
        \ex Match to field \c{purple}{F2} in XA, field-final position: position $\color{purple}n_1\color{black} = 9$; $f[n_1] = f[N]$ (where $n_1$ is field-final)
        \ex Next position: \c{magenta}{$n_2$} $= n_1 + 1 = 10$
        \ex Field at position $n_2$: $f[n_2] =$ \c{red}{FV}
        \ex Match to field \c{red}{FV} in CQ, field-initial position, \color{red}$n$\color{black}: $f[n] =$ \c{red}{FV}; $n =$ 16  (where $n$ is field-initial)   
        \ex Symbol at position \c{red}{$n$}: $s[n] =$ \sf{V}
        \ex To subsequently generate symbol \sf{V} for the continuation-string's next position $N + 1 = 24$, update $s[N] = s[n] =$ \sf{V}
        \ex To subsequently generate structural-variable values for position $N+1$, update $f[N] = f[n] =$ FV; $r[N] =$ CA; $d[N] = 0$
    \end{xlist}
\end{exe}

(\ref{ex:NFex}) can be compactly expressed with two productions (\ref{ex:NFprod1})--(\ref{ex:NFprod2}), communicating through a new variable $f$\bp\ that stores the next-field name (FV here) in the state structure at position $N$.
Note that the condition ``$n_1$ is field-final'' is equivalent to ``$n_1 + 1 = n_2$ is field-initial'', i.e., $d[n_2] = 0$; ``$n$ is field-initial'' is simply $d[n] = 0$.

\begin{exe}
    \ex \NF\ as a sequence of productions: first production
    \label{ex:NFprod1}
    \begin{xlist}
        \ex Condition: $n_2, N$ satisfy $f[n_2-1] == f[N]$, $r[n_2 - 1]$ == XA, $d[n_2] == 0$ 
        \ex Action: set $f$\bp$[N] := f[n_2]$
    \end{xlist} 
\end{exe}

\pagebreak

\begin{exe}
    \ex \NF\ as a sequence of productions: second production
    \label{ex:NFprod2}
    \begin{xlist}
        \ex Condition: $n, N$ satisfy $f[n] == f$\bp$[N]$, $r[n] ==$ CQ, $d[n] == 0$
        \ex Action: set $s[N] := s[n]$, $f[N] := f[n]$, $r[N] :=$ CA
    \end{xlist} 
\end{exe}
 
Note that the condition ``$n_1 = n_2 - 1$ is in region XA'' is equivalent to ``$r^*[n_2] =$ XA''. 
Because condition-matching of each production at each position $N$ operates independently within its own layer, the dummy variable $n_2$ in (\ref{ex:NFprod1}) can be replaced with $n$; at any position, when the first production executes, this $n$ will be bound to a position independently of the position $n$ appearing in (\ref{ex:NFprod2}) that will be bound when the second production executes.
Transforming the productions (\ref{ex:NFprod1})--(\ref{ex:NFprod2}) then yields (\ref{ex:NFprod1*})--(\ref{ex:NFprod2*}).

\begin{exe}
    \ex \NF\ as a sequence of Transformed productions: \textsf{NextField1}
    \label{ex:NFprod1*}
    \begin{xlist}
        \ex Condition: $n, N$ satisfy $f^*[n] == f[N]$, $r^*[n]$ == XA, $d[n] == 0$
        \ex Action: set $f$\bp$[N] := f[n]$
    \end{xlist} 
\end{exe}

\begin{exe}
    \ex \NF\ as a sequence of Transformed productions: \textsf{NextField2}
    \label{ex:NFprod2*}
    \begin{xlist}
        \ex Condition: $n, N$ satisfy $f[n] == f$\bp$[N]$, $r[n] ==$ CQ, $d[n] == 0$
        \ex Action: set $s[N] := s[n]$, $f[N] := f[n]$, $r[N] :=$ CA, $x[N] := 1$
    \end{xlist} 
\end{exe}

The \NF\ productions so far cover the case shown in our working example (\ref{ex:NFmatrix}).
There is another case however, which will explain the appearance of $x$ in (\ref{ex:NFprod2*}b):
when XA includes a delimiter that is not present in XQ and therefore not in CQ.
This occurs when the Answer inserts fixed material absent in the Question, as in the \textsf{Active} $\rightarrow$ \textsf{Passive} template:
$x\ V\ y \mapsto y\ \rm{was}\ V\ \rm{by}\ x$.
In this case, the last-generated continuation symbol's field must be found in XA rather than in CQ. 
Thus another conditional branch is required: if the \textsf{NextField2} production fails to execute (no matching field in CQ) then we need to execute \textsf{NextField3}, which is identical to \textsf{nextField2} except that CQ is replaced by XA:

\begin{exe}
    \ex \NF\ as a sequence of Transformed productions: \textsf{NextField3}
    \label{ex:NFprod3*}
    \begin{xlist}
        \ex Condition: $n, N$ satisfy $f[n] == f$\bp$[N]$, $r[n] ==$ XA, $d[n] == 0$, $x[N] == 0$
        \ex Action: set $s[N] := s[n]$, $f[N] := f[n]$, $r[N] :=$ CA
    \end{xlist} 
\end{exe}

\textsf{x\_temp} ($x$ for short) is a `branch variable', explained shortly; it is initialized to 0 but set to 1 if \textsf{NextField2} executes; this blocks \textsf{NextField3} because of its Condition $x[N] == 0$. 

\subsubsection{The generation algorithm GEN}\label{sec:GEN} 
The core of the generation algorithm GEN is provided by the four productions \textsf{ContField} (\ref{ex:CFprod}), \textsf{NextField1} (\ref{ex:NFprod1*}), 
\textsf{NextField2} (\ref{ex:NFprod2*}), and \textsf{NextField3} (\ref{ex:NFprod3*}).
However there is a conditional branch here: IF the most-recently-generated symbol is not field-final, THEN \textsf{ContField} should execute (its Condition will be satisfied in that case), but \textsf{NextField} should \textit{not} execute (even if its Condition as currently stated is satisfied); ELSE the \textsf{NextField} productions \textit{should} execute.

We implement this branching through a \textit{branch variable} --- an additional state variable \textsf{end} ($e$ for short), which is initialized in each cell to \textsf{end} := 0, and set to value 1 in cell $N$ by the execution of \textsf{ContField} in that cell; the Conditions of the \textsf{NextField} productions are supplemented to include \textsf{end}[$N$] == 0.
This ensures that the \textsf{NextField} productions will not execute if \textsf{ContField} has.

This gives the core of GEN in PSL as the sequence of productions (\ref{ex:CFprod**}) -- (\ref{ex:NFprod3**}).

\begin{exe}
    \ex \textsf{ContField}: final form  [production G1 of (\ref{tbl:gen})]
    \label{ex:CFprod**}
    \begin{xlist}
        \ex Condition: $n, N$ satisfy $s^*[n] == s[N]$, $r^*[n] ==$ CQ, $d[n] = 1$ 
        \ex Action: set $s[N] := s[n]$, $f[N] := f[n]$, $r[N] :=$ CA, $d[N] := 1, e[N] := 1$
   \end{xlist} 
\end{exe}

\begin{exe}
    \ex \textsf{NextField1}: final form [production G2 of (\ref{tbl:gen})]
    \label{ex:NFprod1**}
    \begin{xlist}
        \ex Condition: $n, N$ satisfy $f^*[n] == f[N]$, $r^*[n]$ == XA, $d[n] == 0, e[N] == 0$
        \ex Action: set $f$\bp$[N] := f[n]$
    \end{xlist} 
\end{exe}

\begin{exe}
    \ex \textsf{NextField2}: final form [production G3 of (\ref{tbl:gen})]
    \label{ex:NFprod2**}
    \begin{xlist}
        \ex \squeeze{0.92}{Condition: $n, N$ satisfy $f[n] == f$\bp$[N]$, $r[n] ==$ CQ, $d[n] == 0, e[N] == 0, x[N] == 0$}
        \ex Action: set $s[N] := s[n]$, $f[N] := f[n]$, $r[N] :=$ CA,
        $x[N] := 1$
    \end{xlist} 
\end{exe}

\begin{exe}
    \ex \textsf{NextField3}: final form [production G3$^\prime$ of (\ref{tbl:gen})]
    \label{ex:NFprod3**}
    \begin{xlist}
        \ex \squeeze{0.92}{Condition: $n, N$ satisfy $f[n] == f$\bp$[N]$, $r[n] ==$ XA, $d[n] == 0, e[N] == 0, x[N] == 0$}
        \ex Action: set $s[N] := s[n]$, $f[N] := f[n]$, $r[N] :=$ CA
    \end{xlist} 
\end{exe}

In addition to these four productions, additional productions are required for book-keeping purposes.
One (production G0) initializes \textsf{end} to 0, and the others (production Gpre-1, Gpre-2) update the \textsf{prev\_v} ($v$*) state variables to correctly provide the value of $v[N-1]$ to each cell's $v^\ast[N]$ state variable, after a production has altered the values of $v$.
All 7 productions constituting GEN are given a complete PSL specification in (\ref{tbl:gen}) in Appendix~\ref{sec:appPSLalg}.

We now take up the parsing algorithm, which is presented below in (\ref{ex:parsing1}); this also omits the book-keeping productions such as those for updating the \textsf{prev\_v} variables after $v$ values have changed. 
These are included in the full 24-production PSL program for PARSE presented in (\ref{tbl:parse}) in Appendix~\ref{sec:appPSLalg}.

\subsection{The Parsing Algorithm PARSE   \label{sec:PARSE}   }
The generation algorithm relies heavily on the structural variables \textsf{region, field, index} ($r, f, d$): these state variables encode the parse of the input produced by an algorithm PARSE that we now discuss.
PARSE is presented descriptively below in (\ref{ex:parsing1}).

Note that the field names F1, FV, F2 which we have used previously are just more readable versions of the names assigned by this algorithm.
For instance, since all field values are initialized to equal the position value (Production P0), the symbol \sf{B} in position 2 of the prompt in (\ref{ex:PRmatrix}) is initialized to $f = 2$, and never changed by the algorithm.
Thus F1, FV, F2 above correspond to the values $f = 2, 4, 5$ resulting from the algorithm (on this particular prompt).
The names given to region-delimiter fields above (FQ, FA) are, however, those used by the algorithm.

The `P\#' labels 0, 1a, 1b, \ldots, 11 in (\ref{ex:parsing1}) identify particular productions in the parsing algorithm: they are given in Appendix~\ref{sec:appPSLalg}. 
Here we illustrate with three of these 16 productions: one typical, and two with distinguishing features.

\begin{exe}
    \ex Parsing algorithm: operations on structural variables \\
    Showing for each production: its goal (with a pointer to its role in the walk-through in Sec.~\ref{sec:parsalg}) (\ref{ex:SWreg}) -- (\ref{ex:SWpromptfield}); the specification(s) in (\ref{ex:Fspec}) justifying it; its number P\#; and a verbal description 
    \label{ex:parsing1}
\end{exe}
\begin{center}
\tiny
\renewcommand{\arraystretch}{1.2}
\begin{tabular}{p{2.4cm} p{3cm} |p{9mm} @{}p{4mm} @{}p{7.5cm}}  %
\hline
Goal   &   &   \hspace{-1mm}(\ref{ex:Fspec})   &   P\#   &   Description \\
\hline
initialize variables & 
    mark region, type as `unset'; give & 
        &   0 &
                everywhere set region = R (r:R), type = T (t:T), index = 1 (d:1) \\
&   each cell a unique field value &
        &   &
                everywhere set field = position (f:p) \\
identify regions (\ref{ex:SWreg}) & 
    mark start of Q-regions & 
        a  &   1a &
                mark position 0 with r:XQ, f:FQ, t:D \\
&  &    b-i,iv  &   1b &
                mark start of CQ at repeat of symbol starting XQ with r:CQ, f:FQ, t:D \\
&   fill in Q-regions & 
        a   &   2a &
                spread r:XQ rightward to 1st delimiter (t:D), where field CQ starts [now all Ds are region-Ds; region-internal Ds are inserted later; `XQ' includes XA for now] \\
&   &   ''   &   2b &
                spread CQ rightward until end of prompt [`CQ' includes CA for now]\\
& mark start of A-regions & 
        a-ii,b-iv &    3a &
                mark start of XA region at \sf{A} in current `XQ' region with r:XA, f:FA, t:D \\
&   &   ''  &   3b &
                mark start of CA at \sf{A} in current `CQ' region with r:CA, f:FA \\
&   fill in A-regions & 
        a-ii   &   4 &
                spread r:XA right until first t:D, where field CQ starts\\
identify region-internal delimiters, constituents& 
    identify region-internal delimiters &
        c-i,iii  &   5a &
                mark a symbol in XQ that is repeated in CQ as a (region-internal) delimiter: t:D [\textit{non-causal}] \\

(\ref{ex:SWpromptfield}) &   &   ''  &   5b &
                mark a symbol in CQ that repeats a symbol in XQ as a delimiter, same field: t:D\\
(\ref{ex:SWXfield}) &   &   ''  &   5c &
                mark a symbol in XA that repeats a symbol in CQ as a delimiter, same field: t:D \\
assign fields 1 (\ref{ex:SWpromptfield}) & &  
        c-i,iv  &   6 & 
            identical untyped symbols in X have the same C field \\
set remaining types &   
    identify remaining delimiters &   
        b-ii   &   7 &
                mark all unset types in XA as delimiters: t:D \\
&   
    identify constituents  &   
           &   7' &
                mark all remaining unset types as constituents: t:C \\
assign fields 2 (\ref{ex:SWpromptfield}) &   &   b-i   &   8   &
            XQ (QR in X) and CQ (QR in C) have identical field-sequences: \\
&   &       &       &
            a constituent field following a particular delimiter field in CQ is the same as the one following the same delimiter field in XQ \\
&   &   b-iii   &   9   &
            constituent fields change only at delimiters \\ 
assign indices (\ref{ex:index})  
&   &       &   10  &
            at a change in field: d:0  \\          
mark end of parsing  
&   &       &   11  &
            at end of prompt, set parse=0: a:0  \\
            \hline
\end{tabular}
\end{center}

\medskip

Production P6 is a typical, simple production, virtually a literal translation from the English description to the formal expressions.

\begin{exe}
    \ex \textit{Production P6.} Identical symbols are contained in the values of identical fields.
    \begin{xlist}
        \ex Condition: $n, N$ satisfy $s[n] == s[N]$ 
        \ex Action: set $f[N] := f[n]$
    \end{xlist}
    \label{ex:prod6}
\end{exe}

The intended effect is that when two symbols at positions $j$ and $k$ match, where $j < k$, the value of $f$ at the earlier position $j$ is copied rightward to become the new value of $f$ at the later position $k$.

Note that when the same symbol appears in cells $j$ and $k$, with $j < k$, P6's condition is satisfied under multiple bindings of $n$ and $N$.
Suppose the fields for these positions are, for concreteness, $f[j] =$ F1 and $f[k] =$ F2.
Then the four $n, N$ pairs meeting P6's Condition are shown in (\ref{ex:P7nN}).

\begin{exe}
\ex Effect of $f[N] := f[n]$ when $j < k$, $f[j] =$ F1, $f[k] =$ F2
\label{ex:P7nN}

\begin{tabular}{*{5}{l}}
        &   $n$     & $N$    & effect if production executes & comment \\
\cline{2-5}
a.    &   $j$     & $k$    & set $f[k] :=$ F1 $= f[n] = f[j]$ & desired effect \\
  b.    &   $k$     & $k$   &   set $f[k] :=$ F2 $= f[n] = f[k]$    & no effect \\
  c.    &   $k$     & $j$   & set $f[j] :=$ F2 $= f[n] = f[k]$  & undesired effect \\ 
  d.    &   $j$     & $j$   &   set $f[j] :=$ F1  $= f[n] = f[j]$   & no effect
\end{tabular}
\end{exe}
According to \ding{194} of (\ref{ex:nN}), when a cell $N$ is updated, it can only read information from at most one cell $n$ (possibly itself); if there are multiple $n$ that meet the condition for updating $N$, the lowest-valued $n$ will be used.
When updating $N=k$, the lowest-valued $n$ meeting the condition will be $n = j < k$, so case a of (\ref{ex:P7nN}) will be the operative one, and the desired effect will occur.
When updating $N=j$, the lowest-valued $n$ meeting the condition will again be $n = j < k$, case d: this re-assigns F1 to $f[j]$, yielding no effect.
Crucially, case d blocks the undesired effect of leftward copying from $k$ to $j$ that would result if c rather then d were the operative bindings.
Thus the lowest-match condition \ding{194} of (\ref{ex:nN}) forces information to only travel from earlier to later positions when P6 executes. 

The third column of the table in (\ref{ex:parsing1}), headed `(\ref{ex:Fspec})', identifies for each production the clauses in the definition of our function class \F\ (\ref{ex:Fspec}) that license that production.
For this production P6, this is (\ref{ex:Fspec}c-i,iv), which state that `no symbol may appear in the value of more than one field type' and `within X, a given field type has a unique value'.\footnote{The qualifier `within X' here raises a subtlety in the action of this production. Two symbols can match, meeting P6's Condition, in two cases. If one of the symbols is contained in the value of a constituent field, then the two matching symbols must both occur in X, because the values of constituent fields in C must be different from those in X (\ref{ex:Cuniqval}). 
If a symbol is contained in the value of a delimiter field, it need not lie within X, but the matching symbol must also be within the value of the same delimiter field, as these fields have a constant value throughout the prompt (\ref{ex:delfix}).
P6 is actually used to handle both these cases.
}

An exceptional production is Production P5a.

\begin{exe}
    \ex \textit{Production P5a.} Mark a symbol in XQ that is repeated in CQ as a (region-internal) delimiter: t:D.
    \begin{xlist}
        \ex Condition: $n, N$ satisfy $r[n] ==$ CQ, $r[N] ==$ XQ, $s[n] == s[N]$ 
        \ex Action: set $t[N] :=$ D
    \end{xlist}
    \label{ex:prod5a}
\end{exe}

(\ref{ex:prod5a}) is the only production requiring $n > N$: the to-be-updated position $N$, in region XQ, precedes the position $n$ in CQ which matches the symbol $s[N]$ and hence determines that $N$ is a delimiter position.
In the implementation, this will require non-causal (forward-looking) attention.
All other productions involve only causal (backward-looking) attention.
Thus the DAT implementation of our PSL program for TGT shares with non-causal (prefix) language models \cite{wang2022languagemodelarchitecturepretraining} the property of using strictly causal attention for generation, but (extremely limited) non-causal attention for parsing the prompt prefix.

An atypically complex production is Production P8.

\begin{exe}
    \ex \textit{Production P8.} XQ and CQ have identical field-sequences: a constituent field following a particular delimiter field in CQ is the same as the one following the same delimiter field in XQ. \\ \hspace{-2mm}
        \begin{tabular}{*{4}{@{}l}}
           a. \hspace{2mm} & Condition: \hspace{1mm} & 
            $n, N$ satisfy & $r^*[n] ==$ XQ, $r[n] ==$ XQ, $t^*[n] ==$ D, $t[n] ==$ C, \\
             & & & $r[N] ==$ CQ, $t^*[N] ==$ D, $t[N] ==$ C, $f^*[n] == f^*[N]$ \\
          b. &  Action:  set \hspace{1mm} & $f[N] := f[n]$
        \end{tabular}
    \label{ex:prod8}
\end{exe}

(\ref{ex:prod8}) in fact has the most complex Condition of all the productions for parsing and generation; the complexity of the formal expressions follows that of the English description.
This is the core production enabling the identification of template structure in which  abstract (syntactic) categories (fields) appearing within the `question' are identified with those appearing in the `answer', e.g., $x$ in the template (\ref{ex:temp1}).

\subsection{Combining PARSE and GEN   \label{sec:combin}  }
The first goal of the \PG\ algorithm  is to parse the prompt; then the second goal is to generate a continuation.
As is standard in production systems \cite{jones2003production}, there is a state variable \textsf{parse} ($a$ for short) encoding the current goal, which the Condition of productions evaluates.
\textsf{parse} = 1 signals that the current goal is to execute the productions comprising PARSE; \textsf{parse} = 0 signals the goal to execute GEN.
So every production in the PARSE program includes in its Condition $a == 1$.
The final production of PARSE (P11) sets $a := 0$, which is required by the Conditions of all the productions of GEN.
This production P11 only applies to the final prompt cell (housing the last symbol of the prompt); $a == 0$ is then propagated to the first generation cell and from there to all subsequent generation cells via the autoregressive updating of the generation cells (\ref{ex:PSMgen}).
The end of the prompt is not marked by an end-of-prompt symbol; rather, within the final prompt cell (holding the last symbol of the prompt), a dedicated state variable $z$ is set to the symbol EOP in the input.

\subsection{The PSL Programming Language}   \label{sec:TSL}

Not just those exhibited above, but in fact nearly all the productions we use for parsing and generation share the form shown in (\ref{ex:prodForm}); for the other productions, see (\ref{ex:addexpr}). 

\begin{exe}
    \ex For some state variables $x, y, z, u, v, w$ and constant field values CONST_i, $1 \leq i \leq 3$:
    \begin{xlist}
        \ex Condition: $n, N$ satisfy $x[n] ==$ CONST_1, $x[n] == y[N]$, $z[N] ==$ CONST_2 \ldots
        \ex Action: $u[N] :=$ CONST_3, $v[N] := w[n]$ \ldots
    \end{xlist}
    \label{ex:prodForm}
\end{exe}

The ellipses `$\ldots$' indicate that any number of equality tests ($\ldots == \ldots$) of the forms shown for the Condition, and any number of assignment statements ($\ldots := \ldots$) of the forms shown for the Action, are permitted.

Note that the condition $x[N] == y[N]$ is not included among the possible Conditions in (\ref{ex:prodForm}), but that effect can be achieved by combining the two allowed Conditions $x[n] == y[N],\ p[n] == p[N]$, as in production P10 in (\ref{tbl:parse}).

Thus the core of a PSL program is a sequence of productions of the form (\ref{ex:prodForm}).
Implementing such a program in a transformer network requires only implementing each production as a transformer layer, the layers sequenced to match the sequence of productions.

Additional expressiveness needed for our ICL programs is also provided by PSL (\ref{ex:addexpr}).

\begin{exe}
    \ex Additional expressiveness in the PSL programming language
    \label{ex:addexpr}
    \begin{xlist}
        \ex \textit{Inequality tests}. Conditions can include inequality tests: $x[N]$ != CONST, as in production P1b in (\ref{tbl:parse}), or $x[N]$ != $y[N]$, as in production P10 in (\ref{tbl:parse}), or $x[n]$ != CONST, as in production G1 in (\ref{tbl:gen}).
        \label{ex:ineq}
        \ex \textit{Operators in Conditions.} Conditions can include operators such as $F$ in ``$x[n] == y[N]@F$'', i.e., $x[n]$ equals the image of $y[N]$ under the mapping $F$. These operators map variable values into other values, such as the function $F_{\rm{pos\_increment}}$ that increments by 1 any value of $p$, the position index, as in the production Ppre-1 in (\ref{tbl:parse}).%
        \footnote{
        For implementation, we will require that $F$ be implementable as a linear transformation; this is always possible when the vectors embedding the values of $y$ are linearly independent, as they are when they are one-hot.
        \label{fn:linearF}
        }%
        \label{ex:operators}
        \ex \textit{Repeat blocks.} A sequence of consecutive productions can be executed repeatedly until some condition (e.g., no state change, or a requirement on state variables with the form of a production's Condition) is achieved. This is illustrated in (\ref{tbl:parse}) by the block containing the two-production sequence Ppre-2a, P2a.
        \label{ex:iter}
        \ex \textit{Causal propagation.}
        Optionally, a production may be specified as requiring `causal' information flow: $n < N$.
       \label{ex:causal}
        \ex \textit{Rightmost selection.}
        Optionally, a production $\ell$ may specify $\texttt{right\_match}(\ell) = \texttt{T}$: then, among positions $n$ whose keys tie for a perfect match with a query, the rightmost (rather than the default case, leftmost: $\texttt{right\_match}(\ell) = \texttt{F}$) position is selected for value propagation. 
        \label{ex:rmost}
        \ex \textit{`In' construction.}
        Rather than specifying a single value for a variable in a production's condition, a construction `\textit{x} in [x_1, x_2, \ldots]' can be used to identify multiple matching values. 
        A `not in' construction is also available.
        \label{ex:in}
    \end{xlist}
\end{exe}

In the code base, a default production is specified as
\verb|where <condition>: <action>|.
For the exact syntax, and specification of the optional features in (\ref{ex:addexpr}), see the formal grammar for PSL provided in \pslgrammar.

\section{TPF, Lower Algorithmic Level. The QKV Machine: Symbolic Attention \label{sec:algII}}
\subsection{The QKVM Architecture     \label{sec:QKVMarch}}
As shown in the preceding section, based in symbolic-AI-style production system computation, the PSL language can be straightforwardly deployed to write fully interpretable programs for the PSM to perform both the parsing and generation phases of \PG.
To bridge to the level of implementation in a transformer neural network, however, we need another 
abstract machine closer in architecture to the transformer itself: this machine --- the QKV Machine --- uses query-key attention to update machine states.
Like the PSM, however, it is purely symbolic. 
In this section, we will present a compilation process for translating PSL programs into QKVM programs.
In the next section, another compiler will be presented that translates a particular QKVM into a numerical machine that uses neural computation: an attention-only transformer network using a form of discrete attention and discrete state normalization, the Discrete-Attention-only Transformer, DAT.

\begin{exe}
    \ex QKV Machine state dynamics: Parallel  processing of the prompt \label{ex:QKVMparsing}
    \begin{xlist}
        \ex Machine states
        \begin{xlist} 
            \ex Like the PSM, the QKV Machine state is a sequence of cell states.
            But now, each cell state is specified by the values of five \textit{cell attributes}: \textsf{query, key, value} ($q, k, v$), as well as \textsf{input} ($i$) and \textsf{output} ($o$).
            The value of an attribute is a structure in SSS: the same set of state variables as for the PSM, each assigned a particular value (possibly \textsf{nil}). 
            \ex \textit{Notation.} Structures in SSS are specified as they are for the PSL.
            For a given cell, the SSS-valued attribute \textsf{query}, for example, might have the value notated ``$r$:CA, $f$:FV'' --- the state structure \textsf{S} $\in$ SSS with two non-null state-variable values, $\textsf{S}.r =$ CA and $\textsf{S}.f =$ FV.
        \end{xlist}
        \ex Layer structure: as for the PSM
        \begin{xlist}
            \ex The cell dynamics is unrolled in time, so that for each step of computation there is a layer of cells.
            \ex In layer 1, the \textsf{input} for cells containing the prompt symbols --- the \textit{prompt cells} --- are assigned appropriate values for \textsf{position} and \textsf{symbol}, the latter taken directly from the prompt string. \label{ex:promptCells*}
            \ex The prompt cells are all updated in parallel; the structure in SSS assigned to the \textsf{input} attribute to the cell in position \textit{p} of layer 2 is the structure assigned to the \textsf{output} attribute for the cell in position $p$ of layer 1.
            This is repeated for all layers.
            \ex A sub-sequence of layers can constitute a repeat block, in which case the layers in that block are evaluated in sequence repeatedly until a termination condition specified for that block is met.
        \end{xlist}
        \ex Cell-state updating \\
        Rather than by productions, in the QKVM, cell updates are performed by a uniform \textit{discrete attention} process: 
        \begin{xlist}
            \ex The structure assigned to the \textsf{output} attribute of a cell $N$ is determined by (i) \textsf{input}[$N$] and (ii) the structure assigned to the \textsf{value} attribute of a cell $n$ with a \textsf{key} that matches the \textsf{query} of cell $N$:
            \\ \mbox{\hspace{1 cm}}
            \textsf{query}[$N$] matches \textsf{key}[$n$]  $\Rightarrow$ \textsf{output}[$N$] := \textsf{value}[$n$] $\triangleright\hspace{-8pt}+$ \textsf{input}[$N$].
            \ex (definition of $\ \triangleright\hspace{-8pt}+$) \textsf{output}[\textit{N}] is a state structure in SSS which is the same as \textsf{input}[\textit{N}] except that the values of state variables that have been assigned non-null values in the structure \textsf{value}[\textit{n}] overwrite whatever values these variables may have had in \textsf{input}[\textit{N}].
            \ex A \textsf{query} structure (exactly) \textit{matches} a \textsf{key} structure if, for every state variable $x$ with a non-null value $v$ in \textsf{query}, the value of $x$ in \textsf{key} is $v$.
            (Any additional variables with non-null values in \textsf{key} are ignored.)
            \ex When a cell $N$ is updated in layer $\ell$, if no position $n$ has a \textsf{key} that matches cell $N$'s \textsf{query} exactly, no Action is taken; if multiple positions $n$ have a matching \textsf{key}, the least such value of $n$ is used (unless $\texttt{right\_match}(\ell) = \texttt{T}$, in which case the largest such value is used).
            \ex Our TGT program does not exploit this additional option, but it is available: if a layer $\ell$ has $\texttt{causal\_atn}(\ell) = \texttt{T}$, then attention to a query at position $N$ can only match positions $n \leq N$.
            
        \end{xlist}
        \ex QKV Language, QKVL
        \label{ex:QKVLang}
        \begin{xlist}
            \ex A particular QKV Machine is specified by a program in the language QKVL: this specifies, for each layer $\ell = 1, ..., L$, a map $\texttt{W}_q^{(\ell)}$ from SSS to SSS which takes the \textsf{input} structure for a cell and maps it to the \textsf{query} structure for that cell.
            \label{ex:homogen}
            All cells in layer $\ell$ use the same map $\texttt{W}_q^{(\ell)}$.
            There are corresponding maps $\texttt{W}_k^{(\ell)}$ and $\texttt{W}_v^{(\ell)}$ mapping from the \textsf{input} to \textsf{key} and \textsf{value}.
            \ex $\texttt{W}_q^{(\ell)}$ (and likewise $\texttt{W}_k^{(\ell)}$ and $\texttt{W}_v^{(\ell)}$) is specified by identifying, for some set of state variables, what value those variables take in the SSS state structure $q^{(\ell)}[N]$. These values are specified in terms of the values of state variables in $i^{(\ell)}[N]$. 
            For example, if the space of possible values $V_x$, $V_y$ for the state variables $x$ and $y$ are the same, $V$, the value of $x$ in $q^{(\ell)}[N]$ might be specified as the value of $y$ in $i^{(\ell)}[N]$: $q^{(\ell)}[N].x = i^{(\ell)}[N].y$
            When it is understood that it is the level-$\ell$ state structure of $q$ that is being specified, we abbreviate this to $x : y$.
            \ex QKVL allows specifications of the form $q^{(\ell)}[N].x = \textsf{F} \big( i^{(\ell)}[N].y \big)$
            for some specified function $\textsf{F}$ on the value-space $V$. 
            (Also written $x : y @ F$.)
            (E.g., for $y = p$, $q^{(\ell)}[N].x = \textsf{F_{\textsf{pos\_increment}}} \: i^{(\ell)}[N].p$  where $\textsf{F_{\textsf{pos\_increment}}}$ increases by 1 any position value for $p$.)
            \ex QKVL also allows inequality specifications of the form $q^{(\ell)}[N].x$ != $i^{(\ell)}[N].y$; this entails that $q$ does not match any key $k$ for which $k.x = i^{(\ell)}[N].y$, but otherwise the value of $k.x$ is ignored in evaluating a match. Similarly, $q^{(\ell)}[N].x$ != $x_k$ is permitted, with $x_k$ a fixed possible value of $x$.
            \ex See (\ref{ex:b}) for further discussion of QKVL.
        \end{xlist}     
    \end{xlist}
\end{exe}

\begin{exe}
    \ex QKV Machine state dynamics: Autoregressive generation of the continuation [as for PSM (\ref{ex:PSMgen})].  \label{ex:QKVMgen}
    \begin{xlist}
        \ex Given a prompt of length $P$, the level-1 states of cells with $p = 1, ..., P$ are determined by the prompt (\ref{ex:promptCells*}). 
        These are processed in parallel according to (\ref{ex:QKVMparsing}).
        The continuation cells are updated autoregressively.
        The level-1 state of cell $P+1$ is the level-$L$ state of cell $P$ (except that $p := P + 1$).
        The values of all state variables (except $p$) are copied from $P$ to $P+1$, not just the \textsf{symbol} variable $s$.
        \ex Cell $P+1$ is processed through all $L$ layers, and then the level-1 state of cell $P+2$ is set equal to the level-\textit{L} state of cell $P+1$ (except that $p := P + 2$). This process iterates until a termination condition is met (e.g., the generation of a special termination symbol).
        \ex The generated continuation string is read from the level-1 continuation cells as the sequence of values of \textsf{input}.\textsf{symbol}.
    \end{xlist}
\end{exe}

\subsubsection{The QKV programming language, QKVL}\label{sec:QKVL} 

\begin{exe}
     \ex QKVL Programming Language 
        \label{ex:b}
        \begin{xlist}
            \ex For a given layer $\ell$, specify each of the $q/k/v$ structures in  $V_{\rm{SSS}}$ via instructions of the form \textit{target-variable : source-variable}, such as
            \begin{itemize}
                \item $x : y$
            \end{itemize}
            which means that for any cell, in the $q/k/v$ structure being specified, the variable \textit{x} is given the value that variable \textit{y} has in the \textsf{input} structure for that cell, or
            \begin{itemize}
                \item $x : \mathsf{x}_i$
            \end{itemize}
            which means that $x$ is assigned the fixed value $\mathsf{x}_i$.  
            This specifies the mappings $\texttt{W}_{q/k/v}^{(\ell)}$ of (\ref{ex:QKVLang}).
            \ex More general specifications are also possible:
            \begin{itemize}
                \item $x : y @ F$
            \end{itemize}
            which means that $x$ is assigned the value $F(\mathsf{input}.y)$ where $F$ is a function over the shared space of possible values for $x$ and $y$ that is defined in a library accessible to QKVL (\ref{ex:QKVLang}.iii). 
            And
            \begin{itemize}
                \item $x$ != $\mathsf{x}_i$
            \end{itemize}
            means that the specified state structure will match any value of $x$ except $\mathsf{x}_i$ (\ref{ex:QKVLang}.iv).
            \ex \textit{Note:} In PSL, expressions such as ``$x[n] : y[N]$'' make reference to the values of the variables $x, y$ at two locations $n, N$; but in QKVL code, the instruction ``$x : y$'' references the values of both variables at the some location. The $y$ value is taken from the \textsf{input} to a cell, and this is assigned to the value of $x$ in that cell's \textsf{query}, \textsf{key}, or \textsf{value} structure, whichever is being specified by the instruction.
        \end{xlist} 
\end{exe}

In the code base, a QKVL program is a description of QKVM layers, in the JSON data format.  
(See \qkvlgrammar.) 
At the highest level, it is an array of layers.  
Each layer is a Python dictionary with the fields ``layer\_comment'' (a text description of the layer's purpose), ``causal\_attn'' (a boolean specifying whether causal attention is enabled for the layer), ``right\_match'' (a boolean specifying whether right-most attention selection should be applied to the layer), and most crucially ``weights'' --- a dictionary describing the three mappings $\texttt{W}_{q,k,v}$ of (\ref{ex:homogen}) which will next be compiled to determine the numerical weights in a layer of a DAT transformer implementing the QKV program.
These mappings are specified by instructions of the form given in (\ref{ex:b}).

An example layer dictionary is:
\texttt{\{``layer\_comment'': ``// parse step pre 1. set prev\_position and prev\_symbol'', ``causal\_attn'': false, ``right\_match'': false, 
``weights'': \{``q'': \{``p'': ``p'',``a'': ``a''\}, ``k'':  \{``p'': ``p@pos\_decrement'', ``a'': ``1''\}, ``v'': \{``p*'': ``p'', ``s*'': ``s''\}\}\}} 

\subsection{Compiling PSL Code to QKVL Code   \label{sec:PSL2QKVL}}
(\ref{ex:PSL2QKVL}) shows how we compile PSL productions of the form (\ref{ex:prodForm}) into lower-level QKVL instructions.
(See also \Apppsltoqkvl.)

\begin{exe}
    \ex From PSL to QKVL \\
    \label{ex:PSL2QKVL}
    \addtolength{\tabcolsep}{-1pt}
    \begin{tabular}{l|cc|l}
        PSL & \multicolumn{2}{c|}{QKV instructions} \\
        \cline{2-3}
        Condition & $\q[N]$ & $\k[n]$ & Comment\\
        \hline
        $z[N] == C_z$   &   $z:z[N]$      &   $z:C_z$       &   $C_z$ is a constant value of $z$  \\   
        $x[n] == y[N]$  &   $x\bp:y[N]$   &   $x\bp:x[n]$   &   $y[N]$ may be replaced by a constant value of $x$ \\
        & & & or a transformed value $y[N]@F$ (\ref{ex:b}b) \\
        \hline
        Action           & \multicolumn{2}{c|}{$\v[n]$} & \\
        \cline{1-3}
        $u[N] := w[n]$   & \multicolumn{2}{c|}{$u:w[n]$}   & $w[n]$ may be replaced by a constant value of $u$
    \end{tabular}
\end{exe}

In the instructions,``$(n)$'' and ``$(N)$'' are redundant, since for the query, the state variables are always evaluated at the cell being updated ($N$), and the information used to do the updating ($\ssf{k}, \ssf{v}$) always comes from the cell $n$ meeting the Condition relating it to $N$.
We therefore typically omit ``$(n)$'' and ``$(N)$'' when giving QKVL instructions, as in (\ref{tbl:parse}) -- (\ref{tbl:gen}) in Appendix~\ref{sec:appPSLalg}.
Note that the specifications of \q, \k, \v\ here apply to every cell in the layer realizing the given production (\ref{ex:homogen}): the use of the labels $N, n$ is only to indicate which of all the cells with the specified \q, \k, \v\ values correspond to the cells labelled $N, n$ in the PSL production being compiled.

It is not hard to see why the compilation of a PSL production to QKVL instructions follows (\ref{ex:PSL2QKVL}), which should be read as follows.
The cell attributes $\ssf{q}, \ssf{k}, \ssf{v}$ are all state structures.
The structure given in (\ref{ex:PSL2QKVL}) for cell $N$'s \textsf{query} has two non-null state variable values: $z$ --- which has the value assigned to $z$ in $N$'s \textsf{input} attribute, \textsf{input}[$N$].$z$, here called $z[N]$ --- and $x$\bp\ --- which has the value assigned to $y$ in \textsf{input}[$N$], written $y[N]$.
Similarly, (\ref{ex:PSL2QKVL}) specifies for the attribute \textsf{key} at cell $n$ the two non-null state variable values $z:C_z, \; x\bp\;:x[n]$.
For cell $N$'s \textsf{query} to match cell $n$'s \textsf{key}, both non-null-valued state variables must match: for $z$ values to match, we must have $z[N] == C_z$, and for $x$\bp\ values to match, we must have $y[N] == x[n]$.
Thus these two values in $\q[N]$ and $\k[n]$ implement the desired Condition.

Any number of equality constraints may be present in the Condition of a PSL production, and for each such constraint, we simply insert into the state structure for \textsf{query} and \textsf{key} the state-variable values given in (\ref{ex:PSL2QKVL}).
This is amply exemplified in (\ref{tbl:parse}) -- (\ref{tbl:gen}).

The \textsf{value} attribute is straightforward: for a pair of cells $n, N$ for which the Condition is met, (\ref{ex:PSL2QKVL}) specifies that $\textsf{value}[n]$ is the state structure in which state variable $u$ is assigned the value that $w$ has in \textsf{input}[$n$] --- this is notated ``$u:w[n]$''.
When the cells $n, N$ meet the Condition, the Action uses this state structure to determine \textsf{output}[$N$], which is the same as \textsf{input}[$N$] except that the value of state variable $u$ in this structure is set to $w[n]$, over-writing any value $u$ may have in \textsf{input}[$N$].
As with the Condition, the Action of a production may have any number of assignments of the form $u[N] := w[n]$, and for each one, $u:w[n]$ is inserted into \textsf{value}[$n$].

As mentioned in the Comment column, the translations in (\ref{ex:PSL2QKVL}) also cover the cases in which $y[N]$ is replaced with a constant or a transformed value $y[N]@F$ in a Condition, or $w[n]$ is replaced by a constant in an Action. 
In fact the specifications in (\ref{ex:PSL2QKVL}) produce a translation for any PSL production having the form (\ref{ex:prodForm}).
The extensions in (\ref{ex:addexpr}) are handled as follows.

\textit{Inequality conditions.} When an inequality condition in a PSL production $z[N] \;!\!= C_z$ is translated into QKVL, \textsf{q} is specified as $z : z$, and \textsf{k} as $z \;!\!= C_z$. 
The semantics of `!=' here is that this \textsf{key} does not match a \textsf{q}[$N$] for any position $N$ where $z[N] = C_z$. 
Similarly, the PSL inequality condition $z[n] \;!\!= C_z$ is translated into QKVL by specifying \textsf{k} as $z : z$ and \textsf{q} as $z \;!\!= C_z$; this \textsf{q} fails to match \textsf{k}[$n$] at any position $n$ where $z[n] = C_z$.

\textit{Operators in Conditions.} Already treated in (\ref{ex:PSL2QKVL}). 

\textit{Repeat blocks.} Just like PSL, QKVL allows a sequence of layers to be tagged as a repeat block which functions just as in PSL.

This covers all the PARSE and GEN productions given in (\ref{tbl:parse}) -- (\ref{tbl:gen}).

\section{TPF, Implementational Level: DAT, a Discrete-Attention-Only Transformer Network    \label{sec:impl}}
We are finally ready to produce our Discrete-Attention-only Transformer network, DAT, which computes ICL functions in \F. 
We need to convert the discrete symbolic instructions of QKVL into matrices of numerical weights that generate the vectors \textbf{q, k, v}.
For this we need an embedding of the discrete state structures of QKVM in SSS into a vector space of transformer hidden states, $V_{\rm{SSS}}$.

\subsection{Embedding Abstract Machine Cell-State Structures as Transformer-Cell Vector States  \label{sec:implReps} }

\begin{exe}
    \ex Embedding state structures \textsf{s} $\in$ SSS as vectors $\mathbf{s} \in V_{\rm{SSS}}$: Fully local (1-hot) case
    \label{ex:FLSSSemb}
    \begin{xlist}
        \ex \textbf{s} is a concatenation of vectors $\mathbf{v}_x$ over all state variables $x$. Within $\mathbf{s}$, $\mathbf{v}_x$ begins with neuron $c_x^{\rm{b}}$ and ends with neuron $c_x^{\rm{e}}$: this is the \textit{x-register} $V_x$ within \textbf{s}.
        \ex The possible values $\{\textsf{x}_i\}_{i=1}^{d_{x}} \equiv \textsf{V}_x$ of variable $x$ are encoded as 1-hot (i.e., localist) vectors $\{\vec{\mathsf{x}_{\textit{i}}}\}_{i=1}^{d_{x}} \subset \mathbb{R}^{d_x}$. 
        We can order the neurons so that $\vec{\mathsf{x}_{\textit{i}}}$ is the $i^{\rm{th}}$ coordinate vector:
        $[\vec{\mathsf{x}_{\textit{i}}}]_j = \delta_{ij}$.
        \ex The vector encoding of the binding of variable $x$ to the value $\textsf{x_i} \in \textsf{V}_x$ is $\vec{\textit{x}:\textsf{x_i}}$; this is the 1-hot vector  $\vec{\mathsf{x}_{\textit{i}}}$ located in the register $V_x$ for variable $x$ (from neuron $c_x^{\rm{b}}$ to neuron $c_x^{\rm{e}})$.
        \ex If there are $M$ state variables encoded in the hidden state, $\{v_m\}_{m=1}^M$, and  variable $v_m$ has $d_{v_m}$ possible values, then $V_{SSS} \equiv \bigoplus_{m=1}^M V_{v_m} \cong \bigoplus_{m=1}^M
        \mathbb{R}^{d_{v_m}} \cong \mathbb{R}^D $ where $D \equiv \sum_{m=1}^M d_{v_m}$.  
        \ex This can be analyzed as a tensor product representation (TPR) with 1-hot embeddings of the state variables serving as TPR \textit{role vectors} and 1-hot embeddings of state variable values serving as TPR \textit{filler vectors}.
    \end{xlist}
\end{exe}

\noindent For explanation of the TPR analysis of this embedding, see \TPR.
The TPR analysis is helpful for showing how our analysis of DAT generalizes from the fully local embedding of (\ref{ex:FLSSSemb}) to more distributed embeddings (\ref{ex:distrSSSemb}).
Although we do not explicitly use these distributed embeddings for the hand-programming reported here, the analysis applies equally well to these embeddings, and is likely to prove necessary for probing for state vectors in embeddings that are learned by transformer models: these are expected to be distributed, not local (see discussion in Sec.~\ref{sec:LMs}).
In the body of the paper we will restrict attention to the simplest case, the fully local embedding, which is visualized in (\ref{ex:SSSvis}).
This is essentially the `disentangled residual stream' used in the RASP implementations of Lindner et al.~(\citeyear{lindner2023tracr}) and Friedman et al.~(\citeyear{friedman2023learning}).

\begin{exe}
    \ex Fully local embedding of SSS visualized   \label{ex:SSSvis}
\end{exe}

\newcommand{\bc}{$\bigcirc \cdot\!\cdot\!\cdot \bigcirc$}
\begin{center}
\tiny
\renewcommand{\arraystretch}{1.5}
\addtolength{\tabcolsep}{-3pt}
\resizebox{\linewidth}{!}{%
\begin{tabular}{|*{15}{c|}}
\multicolumn{4}{r}{$c_f^{\rm{b}} \hspace{3mm} c_f^{\rm{e}}$} \\
\multicolumn{4}{r}{$\downarrow \hspace{5mm} \downarrow$} \hspace{.5mm}\\
\hline
\bc & \bc & \bc & \bc & \bc & \bc & \bc & \bc & \bc & \bc & \bc & \bc & \bc & \bc & \bc \\
\hline
position &  symbol &  region &  field &  type &  index &  prev\_pos  &  prev\_sym  &  prev\_reg  &  prev\_fld  &  prev\_type & prev\_ind   &  position\bp  &  symbol\bp  &  $\cdots$     \\
\centering$p$& \centering$s$ & \centering$r$ & \centering$f$ & \centering$t$ & \centering$d$ & \centering$p$*  & \centering$s$*  & \centering$r$*  & \centering$f$*  & \centering$t$*  & \centering$d$*  & \centering$p$\bp  & \centering$s$\bp  & \centering$\cdots$     
\end{tabular}
}
\end{center}

\begin{exe} 
\ex Properties of the fully local embedding of SSS in $V_{\rm{SSS}}$ \label{ex:SSSlocal}
    \begin{xlist} 
        \ex In this embedding, a state variable $x$ (a TPR role) corresponds to a subspace $V_x$ of $V_{\rm{SSS}}$: the subspace spanned by the unit vectors for dimensions $c_x^{\rm{b}}$ through $c_x^{\rm{e}}$. 
        \ex The subsequence of neurons encoding a variable $x$ --- spanning from neuron $c_x^{\rm{b}}$ through neuron $c_x^{\rm{e}}$ --- constitutes the $x$-register within the overall vector $\mathbf{s}$.
        \ex When two state variables $x$ and $y$ have the same set of possible values $\textsf{V} = \{\textsf{v}_1, \ldots, \textsf{v}_d \}$, there is an isomorphism between $V_x \cong \mathbb{R}^d$ and $V_y \cong \mathbb{R}^d$: for any value $\textsf{v}_i \in \textsf{V}$, $\vec{\textit{x}:\textsf{v_i}} \in V_x$ corresponds to $\vec{\textit{y}:\textsf{v_i}} \in V_y$, these being the same vector $\vec{\mathsf{v}_{\textit{i}}}$ (the embedding of $\textsf{v}_i$ --- the $i^{\rm{th}}$ coordinate vector --- in $\mathbb{R}^d$) located within the $x$ and $y$ registers, respectively. This is what it means for $x$ and $y$ to ``have the same value''. (In the TPR analysis, $x$ and $y$ are roles bound to the same filler $\textsf{v}_i$; see \TPR).
    \end{xlist}
\end{exe}

\begin{exe}
    \ex Distributed embeddings summary --- from \TPR
    \label{ex:distrSSSemb} \\
    There are two types of distributed embeddings which are isomorphic, under orthogonal linear transformations, to the fully local embedding of (\ref{ex:FLSSSemb}), and under the TPR analysis take the identical form, with only these differences:
    \begin{xlist} 
        \ex Semi-local orthonormal embedding: identical, except the filler vectors $\vec{\mathsf{v}_{\textit{i}}}$ are orthonormal (but not necessarily 1-hot).
        There are localized registers, but the vector embeddings of values for the state variables $\{\vec{\mathsf{v}_{\textit{i}}}\}_{i=1}^{d_{x}}$ are, in general, dense distributed --- not 1-hot --- vectors.
        \ex Fully distributed orthonormal embedding: identical, but the role vectors as well as the filler vectors are orthonormal (and not necessarily 1-hot). 
        When the role vectors are not 1-hot, there are no localized registers: all state variables' embeddings are superimposed over the complete set of neurons spanning $V_{\rm{SSS}}$.
        In this case, the hypothesis that a learned hidden representation has the form specified here requires special methods for uncovering full distributed TPRs: see the discussion of the \textsc{Discover} technique in Sec.~\ref{sec:LMs}.
    \end{xlist}
\end{exe}

\subsection{The DAT Architecture   \label{sec:DATarch} }
To perform ICL, the DAT is given as input an appropriate prompt: a symbol string of variable length $P$, with $\textsf{s[p]} \in \textsf{S}$ being the type of the symbol in position $\textsf{p} \in \{1, \ldots, P\} \equiv \textsf{P}$, and $\textsf{S}$ the vocabulary of symbol types.
The input layer of DAT, like that of a decoder-only transformer, extends beyond the $P$ \textit{prompt cells} hosting the prompt; the layer extends up to a fixed total of $T$ cells, with the cells following the initial $P$ prompt cells --- the \textit{continuation cells} --- ready to host the continuation string that constitutes the model's output.

\begin{exe}
    \ex Discrete-Attention-only Transformer (DAT) state dynamics: Parallel processing of the prompt  \label{ex:a}
    \begin{xlist}
        \ex Machine states
        \begin{xlist} 
            \ex Layers $1, \ldots, L$, each a sequence of subnetworks called \textit{cells} with positions $1, \ldots, T$
            \ex Each cell has a state specified by 5 \textit{attribute vectors}, each in a vector space $V_{\rm{SSS}}$:
            \begin{itemize}
                \item input vector $\mathbf{i}$ 
                \item query, key, value vectors $\mathbf{q, k, v}$
                \item output vector $\textbf{o}$
            \end{itemize}
            \ex The query, key and value vectors for any cell are affine transformations of its input vector. The weights in these transformations are the parameters specifying the layer (they are constant throughout a layer).
            \ex Unlike standard transformers, there are no MLP or LayerNorm sublayers; DAT is an attention-only transformer architecture (with a single attention head).   
        \end{xlist}
        \ex Inputs in layer 1
        \begin{xlist}
            \ex The layer-1 prompt cell with position $\textsf{p}$ has input vector $\textbf{i}^{(1)}[\textsf{p}] = \vec{\textit{s}:\textsf{s[p]}} +             \vec{\textit{p}:\textsf{p}} + \vec{\textit{a}:1}$ where $\textsf{s}[\textsf{p}]$ is the type of the symbol at position \textsf{p}, and $a$ is the \textsf{parse} flag which is initialized to 1 for each prompt cell.
            (These are the cells subject to the productions of PARSE given in (\ref{tbl:parse}).)
            \ex 
            The input vector of the layer-1 cell carrying the final symbol of the prompt has an additional variable set, the end-of-prompt flag variable $z$ which is set to \textsf{EOP}: $\vec{\textit{z}:\textsf{EOP}}$.
            \ex Note that in the local embedding, $\vec{\textit{s}:\textsf{s[p]}}$ is the 1-hot vector embedding of  $\textsf{s}[\textsf{p}]$ in the $s$-register $V_s$ and $\vec{\textit{p}:\textsf{p}}$ is the 1-hot vector embedding of $\textsf{p}$ in the $p$-register $V_p$: summation of these vectors in two orthogonal subspaces is equivalent to concatenation.      
        \end{xlist}
        \ex Cell-state update computation: Discrete Attention (single-headed)
            \begin{xlist}
                \ex \textit{Attention vectors; DATmax.} \label{ex:attnWts} In any layer, cell $N$'s output vector \textbf{o}[\textit{N}] is determined by (i) the cell's input vector \textbf{i}[$N$] and (ii) the value vector \textbf{v}[\textit{n}] of a cell $n$ with a key vector \textbf{k}[$n$] that matches the non-null variable values encoded in the query vector \textbf{q}[$N$] of cell $N$; this $n$ is given by the function $\alpha$ defined as follows. 
                For all $N = 1, ..., T$, let the raw and normalized query-key dot-product vectors
                $\delta[N], \ \hat{\delta}[N] \in \mathbb{R}^T$  have components, for $n = 1, ..., T$:
                \\
                $
                \mbox{\hspace{1 cm}}    \mathbf{\delta}[N]_n \equiv \mathbf{q}[N] \cdot \mathbf{k}[n]
                \mbox{\hspace{1 cm}}    \mathbf{\hat{\delta}}[N]_n =  \mathbf{\delta}[N]_n /  || \mathbf{q}[N]  ||^2
                $
                \\
                Then, for the following reason, we define%
                \\
                $
                \mbox{\hspace{1 cm}}    \alpha(N) \equiv \min \big\{n \in \{1,...,T\}  \ \big\|\  \hat{\delta}[N]_n = 1 \big\}$  
                \\
                Because of the 1-hot encodings we are using in $V_{SSS}$, $\mathbf{q}[N] \cdot \mathbf{k}[n]$ is the number of variable registers set in \textbf{q}[$N$] to non-null values for which the corresponding variable registers in \textbf{k}[$n$] have the same value; the total number of such non-null values in \textbf{q}[$N$] is just $|| \mathbf{q}[N] ||^2$, which means that $\mathbf{\hat{\delta}}[N]_n \leq 1$, with equality holding only when the desired perfect matching occurs between the query and key vectors.
                Hence the requirement that $\hat{\delta}[N]_n = 1$.%
                \footnote{The DAT programs produced by our compiler will always yield $|| \mathbf{q}[N]  || = || \mathbf{k}[n]  ||$
                for all $N, n$ in the same layer. Thus we can equivalently define $\mathbf{\hat{\delta}}[N]_n =  \hat{\mathbf{q}}[N] \cdot \hat{\mathbf{k}}[n]$ while normalizing both \textbf{q} and \textbf{k} to unit length.
                For extension of these mathematical properties of attention in DAT to non-one-hot encodings, see \TPR. \label{fn:qklength}
                }
                If no perfectly matching $n$ exists, $\alpha[N]$ is undefined, and attention at $N$ will return $\mathbf{0}$; if multiple exact matches occur, $\alpha[N]$ is (by default) the first/leftmost match: hence the min over $n$. (In the non-default case of a layer $\ell$ with $\texttt{right\_match}(\ell) = \texttt{T}$, in the definition of $\alpha$, min is replaced with max.)%
                \footnote{In future work exploring in-weights learning in the DAT, it may improve gradient propagation to implement left- (or right-)mostness differentiably as follows.
                Suppose that before computing final attention weights for $N$ we preprocess the normalized key-query dot-product vector $\hat\delta[N]$ by setting to 0 all values at $n$ with $\hat\delta[N]_n < 1$, e.g. via $\hat\delta[N] - \textrm{ReLU}(\vec{1} - \hat\delta[N])$; then it remains to pick the left- (or right-)most of the remaining $\hat\delta[N]$ values (if any).
                Consider the leftmost case. 
                We add to $\hat\delta[N]_n$ a quantity $\lambda(n)$ which smoothly decreases as $n$ increases and which has an overall magnitude small enough to ensure that the addition of $\lambda$ will break exact ties in $\hat\delta[N]$ but never alter the location of $\arg\!\max_n \{\hat\delta[N]_n + \lambda(n)\}$. 
                Using a new state variable $\lambda$, this could be arranged by inserting, for small enough $\epsilon$, $\lambda(n) : \epsilon / n$ into \ssf{key}[$n$] and $\lambda : 1$ into \ssf{query}[$N$] for all $N$ ($n$ is the numerical value of the position variable $\ssf{input}[n].p$).
                An analogous maneuver could yield right-mostness via a new state variable $\rho$ assigned small values that smoothly \textit{increase} with $n$. 
                Then each production could select for left- or right-most attention by specifying either $\lambda : 1$ (and $\rho$ set to \textsf{null}) or $\rho : 1$ ($\lambda$ \textsf{null}), respectively, in its particular query.
                With this modification in place, DATnorm could be differentiably approximated by softmax with a low temperature parameter, returning at $N$ a nearly-1-hot vector of attention weights, `hot' at $\arg\!\max_n \{\delta[N]_n + \lambda(n)\}$ (or $\arg\!\max_n \{\delta[N]_n + \rho(n)\}$), the left- (or right-)most location $n$ of a 1 in $\hat{\delta}[N]_n$.
                }
               So $\alpha$ implements exactly the attentional selection process of the QKV machine (\ref{ex:QKVMparsing}c).
               Thus the attention-weight vector computed in standard transformers by the softmax over normalized query-key dot-products is  replaced in DAT by 
                \\
                $\mbox{\hspace{1 cm}}$    DATmax$(\hat{\delta}[N]) \equiv$  \textbf{1-hot}($\alpha(N)$)   
                \\
                where \textbf{1-hot}($n$) is the one-hot vector in $\mathbb{R}^T$ with value $1$ at location $n$.
                \ex \textit{Inequality conditions.} Recall from Sec.~\ref{sec:PSL2QKVL} that a PSL production with an inequality condition $z[N] \;!\!= z_i$ is translated into QKVL by specifying \textsf{q} as $z : z_i$, and \textsf{k} as $z \;!\!= z_i$; this \textsf{key} does not match a \textsf{query} for which the value of $z$ is $z_i$. 
                In DAT, for all cells $n$ in the layer implementing this production, the activity of each neuron $\mu$ in the $z$ register of \textbf{k}[$n$] is set to $[\mathbf{k}[n]]_\mu = 1 - [\vec{\mathsf{v}_{\mathit{i}}}]_\mu$, the ones'-complement of $\vec{\mathsf{v}_{\mathit{i}}}$.
                Within the $z$-register, this will yield a raw-attention dot-product $\mathbf{\delta}[N]_n \equiv \mathbf{q}[N] \cdot \mathbf{k}[n]$ of 0 when 
                $\mathbf{q}[N] = \vec{\mathsf{v}_{\mathit{i}}}$, but a dot product of 1 whenever $\mathbf{q}[N] = \vec{\mathsf{v}_{\mathit{j}}}$ with $j \neq i$.
               
                \ex  \textit{DATnorm.}  \label{ex:DATnorm} Given a vector $\mathbf{u} \in V_{SSS}$, define DATnorm(\textbf{u}) so that for each state variable $x$, the register subspace $V_x$ --- hosted by neurons $c_x^{\rm{b}}$ through $c_x^{\rm{e}}$ --- is independently normalized within DATnorm(\textbf{u}) to the 1-hot vector with activity 1 at the neuron in position $\arg\!\max_{i \in V_x} \{\mathbf{u}_i\}$, where `$i \in V_x$' abbreviates `$c_x^{\rm{b}} \leq i \leq c_x^{\rm{e}}$'; if \textbf{u}_x = \textbf{0}, we set DATnorm(\textbf{u}_x) = \textbf{0}. As used in the DAT architecture in (\ref{ex:oN}), when $\mathbf{u} \neq \mathbf{0}$, there will never be a tie for the maximum element of a vector $\{\mathbf{u}_i\}_{i \in V_x}$ used as an argument to $\arg\!\max$ for computing DATnorm. 
                Note that DATnorm operates within each register independently, quite unlike LayerNorm.
                Thus the register structure within DAT's hidden (attribute) vectors impose real  structure on DAT computation.
                \ex \textit{The output vector} \label{ex:oN} \textbf{o}[\textit{N}] of a cell $N$ is given by
                \\ 
                $\mbox{\hspace{1 cm}} \mathbf{o}[N] = \textrm{DATnorm}\big(\mathbf{i}[N] + \kappa \mathbf{v}[n] \big)
                   \hspace{1cm} n \equiv \alpha(N)$
               \\ where $\mathbf{v}[n]$ is the \textsf{value} attribute vector of cell $n$; $\alpha$ is defined in (\ref{ex:a}c.i), and if it is undefined, we have simply $\mathbf{o}[N] = \textrm{DATnorm} \big( \mathbf{i}[N] \big) = \mathbf{i}[N]$.
               We can choose any $\kappa  > 1$; we let $\kappa = 2$.
                (As in standard transformers, the term $\mathbf{i}[N]$ is the contribution of a residual connection which adds the input vector to the output.)
                Three cases exist for each state variable $x$: 

                \leftskip 60 pt \parindent=-30pt
                if $\mathbf{v}[n]$ does not have $x$ set to a non-null value, then --- within register $V_x$ --- $\mathbf{v}[$n$] = \mathbf{0}$ and therefore \textbf{o}[$N$] = \textbf{i}[$N$];
      
                \leftskip 60 pt \parindent=-30pt if $\mathbf{v}[n]$ has $x$ set to a non-null value $\mathsf{v}_k$, and in $\mathbf{i}[N]$, $x$ has a different non-null value $\mathsf{v}_j$, then within register $V_x$, \textbf{o}[$N$] = $\vec{\mathsf{v}_{\mathit{k}}}$, the 1-hot embedding of $\textsf{v}_k$. 
                This is because, in the $x$ register, $\mathbf{i}[N] + \kappa \mathbf{v}[n]$ has value 1 on the $j^{th}$ neuron (from $\mathbf{i}[N]$) and value $\kappa = 2 > 1$ on the $k^{th}$ neuron (from $\mathbf{v}[n]$), so the DATnorm operation resets the former to 0 and the latter to 1.%
                
                \leftskip 60 pt \parindent=-30pt  if $\mathbf{v}[n]$ has $x$ set to a non-null value $\mathsf{v}_k$ that is the same value as $\mathbf{i}[n].x$, then $\mathbf{i}[N] + \kappa \mathbf{v}[n] = (1 + 2)\vec{\mathsf{v}_{\mathit{k}}}$ which is similarly reset by DATnorm to simply $\vec{\mathsf{v}_{\mathit{k}}}$. 
                                
                \leftskip 30pt Thus this implements the update rule for the QKV machine (\ref{ex:QKVMparsing}c).
        \end{xlist}
         \ex Cell updating sequence 
        \begin{xlist}
            \ex The prompt cells are all updated in parallel; the input attribute vector \textbf{i}^{(2)}[\textsf{p}] of the cell in position \textit{p} of layer 2 is the output attribute vector \textbf{o}^{(1)}[\textsf{p}] for the cell in position $p$ of layer 1.
            This is repeated for all layers.
            \ex A sub-sequence of layers can constitute a repeat block, in which case the layers in that block are evaluated in sequence repeatedly until a termination condition specified for that block is met.
        \end{xlist}
    \end{xlist}
\end{exe}

\begin{exe}
    \ex DAT Machine state dynamics: Autoregressive generation of the continuation [as for QKVM (\ref{ex:QKVMgen})].  \label{ex:DATgen}
    \begin{xlist}
        \ex Given a prompt of length $P$, the level-1 states of cells with $p = 1, ..., P$ are determined by the prompt (\ref{ex:a}b.i). 
        These are processed in parallel according to (\ref{ex:a}).
        The continuation cells are updated autoregressively.
        The level-1 state of cell $P+1$ is the level-$L$ state of cell $P$ (except that $p := P + 1$).
        The values of all state variables (except $p$) are copied from $P$ to $P+1$, not just the \textsf{symbol} variable $s$.
        \ex Cell $P+1$ is processed through all $L$ layers, and then the level-1 state of cell $P+2$ is set equal to the level-\textit{L} state of cell $P+1$ (except that $p := P + 2$). This process iterates until a termination condition is met (e.g., the generation of a special termination symbol).
        \ex The generated continuation string is read from the level-1 continuation cells as the sequence of symbols that are embedded as the vectors $\mathbf{i}^{(1)}[\textsf{p}].s$, $\textsf{p} = P, P+1, ...$
    \end{xlist}
\end{exe}

A DATL program specifies a sequence of layers, some of which may be marked as forming a repeat block with a specified termination condition.
Each layer $\ell$ is specified by a set of query/key/value weights, which determine each of a cell $n$'s attribute vectors $\mathbf{q}[n], \mathbf{k}[n], \mathbf{v}[n]$ through an affine transformation of the cell's input vector $\mathbf{i}[n]$ via a matrix $\mathbb{W}^{(\ell)}_{\mathbf{q, k, v}}$ and a bias vector $\mathbf{b}^{(\ell)}_{\mathbf{q, k, v}}$ that implement the mappings $\texttt{W}_{q,k,v}$ of (\ref{ex:homogen}).%
\footnote{
The current default size of each bias vector in the $\texttt{dat\_explorer}$ simulator, Sec.~\ref{sec:tfexplorer}, is 3936 (41 registers each of size 96); the size of each matrix is [3936, 3936]. }%

\subsubsection{Compiling from QKVL to DATL\label{sec:QKVL2DATL}}
As spelled out in (\ref{ex:DATLtable}), compiling a QKVL program produces a set of affine-transformation weights (in a matrix and bias vector): for each layer, these transform a cell's input vector \textbf{i} into its query/key/value attribute vectors \textbf{q}, \textbf{k}, \textbf{v}.
Compiling proceeds layer by layer as follows.  
A QKVL instruction for a layer, a \textit{target-variable} : \textit{source-variable} pair such as $x : y$, is compiled into a set of weights that can be loaded into the DAT Transformer.
Weights are built up starting from zero, adding for each QKVL instruction a contribution to the weight matrices and biases.
$[\mathbb{W}_{\mathbf{k}}]_x^y$ denotes the block submatrix (within the key-generating matrix $\mathbb{W}_{\mathbf{k}}$) which maps the subspace $V_y$ of $V_{SSS}$ embedding values of a source variable $y$ into the subspace $V_x$ embedding values of the target variable $x$.
$\mathbf{b_k}$ is the constant bias term in the affine transformation mapping \textbf{i} to \textbf{k}:
$\mathbf{k} = \mathbf{i} \mathbb{W}_{\mathbf{k}} + \mathbf{b_k}$.

\pagebreak

\begin{exe}
\ex Compiling from QKVL to DATL, for attribute \textbf{k} (attributes \textbf{q}, \textbf{v} operate identically)
\label{ex:DATLtable}

\small
\begin{tabular}{*{5}{l}}
        &   QKVL     & DAT result & attribute vector   & contribution to weights $\mathbb{W}_{\mathbf{k}}, \mathbf{b}_{\mathbf{k}}$  \\
\cline{2-5}
\\ 

a.    &   $x : y$     & $\mathbf{k}[n].x := \mathbf{i}[n].y$    & $\mathbf{k}[n] := \mathbf{i}[n] \: \mathbb{W}_{\mathbf{k}}$    & $[\mathbb{W}_{\mathbf{k}}]_x^y = \mathbb{I}_{d_x}$ \\
& & & & (block identity matrix $V_y \rightarrow V_x$) \\
& & & & $d_x$ = \# possible values for $x$ = $d_y$ \\
b. & $x : y@F$ & $\mathbf{k}[n].x := \big(\mathbf{i}[n].y\big) \: \mathbb{F} $ & $\mathbf{k}[n] := \mathbf{i}[n] \: \mathbb{W}_{\mathbf{k}}$ & $[\mathbb{W}_{\mathbf{k}}]_x^y = \mathbb{F}$ \\
& & & & $\mathbb{F}$ = matrix implementing \\
& & & & \mbox{\hspace{6mm}} value-transform $F$
\\
c. & $x : \textsf{x}_i$ &  $\mathbf{k}[n].x := \vec{\mathsf{x}_{\mathit{i}}}$ & $\mathbf{k}[n] := \mathbf{b}_{\mathbf{k}} $ & $[\mathbf{b}_{\mathbf{k}}]_x =  \vec{\mathsf{x}_{\mathit{i}}}$ \\
& & & & the $x$ register $V_x$ of $\mathbf{b}_{\mathbf{k}}$ is
$\vec{\mathsf{x}_{\mathit{i}}}$ \\
d. & $x$ != $\textsf{x}_i$ &  $\mathbf{k}[n].x := \vec{1} - \vec{\mathsf{x}_{\mathit{i}}}$ & $\mathbf{k}[n] := \mathbf{b}_{\mathbf{k}} $ & $[\mathbf{b}_{\mathbf{k}}]_x =  \vec{1} -\vec{\mathsf{x}_{\mathit{i}}}$ \\
& & & & %
the ones'-complement of $\vec{\mathsf{x}_{\mathit{i}}}$ \\
e. & $x$ != $y$ &  $\mathbf{k}[n].x := \vec{1} - \mathbf{i}[n].y$ & $\mathbf{k}[n] := \mathbf{i}[n] \: \mathbb{W}_{\mathbf{k}}$  & $[\mathbb{W}_{\mathbf{k}}]_x^y = \mathbf{1} \hspace{-1mm} \mathbf{I} - \mathbb{I}_{d_x}$ \\
& & & & the ones'-complement of $\mathbb{I}_{d_x}$
\end{tabular}
\end{exe}
\normalsize
\vspace{-5mm}
\noindent $\vec{1}$ is the vector with 1 in every position; thus $\vec{1} - \vec{\mathsf{x}_{\mathit{i}}}$ is the ones'-complement of $\vec{\mathsf{x}_{\mathit{i}}}$, the vector with 0 in position $i$ and 1 in every other position: this matches the vector embedding of an \textit{x} value $\vec{\mathsf{x}_{\mathit{j}}}$ if and only if $j \neq i$.
$\mathbf{1} \hspace{-1mm} \mathbf{I}$ is the matrix of all 1s. 
(Note \ref{fn:linearF}, page \pageref{fn:linearF}, points out that the $\mathbb{F}$ needed for (\ref{ex:DATLtable}b) always exists, given a linearly-independent, e.g., one-hot, encoding of variable values.)
\subsubsection{DAT Operation\label{sec:datoperation}}
Implementation details of the operation of the DAT, viewed as a modification of a standard transformer, are presented in \datop.

\subsubsection{The\ \texttt{dat\_explorer} Simulator\label{sec:tfexplorer}}
The TPF is implemented in a publicly-accessible python package on \texttt{github};%
\footnote{\url{https://github.com/microsoft/discrete_attn_transformer/}}
it consists of several programs: \texttt{psl\_compiler}, \texttt{weights\_compiler}, \texttt{dat\_interpreter}, \texttt{dat\_transformer}, and \texttt{dat\_explorer}.
The \texttt{dat\_explorer} enables users to load a PSL program, compile it into a QKVL program, compile that into the numerical weights for a DAT, enter a prompt and run the DAT, visualizing each of its resulting cell states (for each column and layer) using the symbolic interpretation of vectors and matrices.  

Here is a screenshot of the program on the propositional-logic inference prompt (\ref{ex:pr2}):

\noindent \includegraphics[scale=.33]{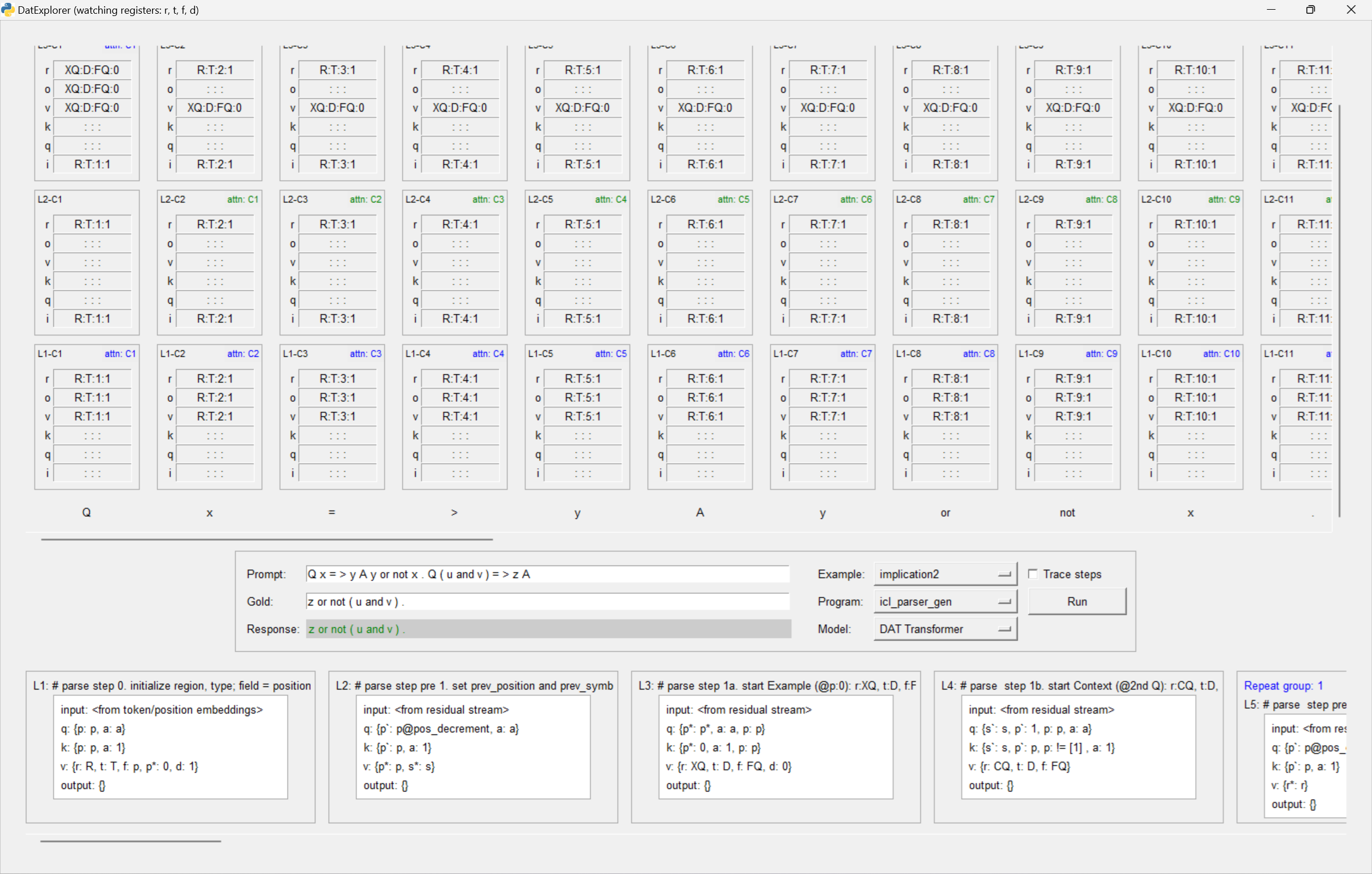}

Visualization features include:

\begin{itemize}
    \item pull-down menu for selecting a PSL program for producing a prompt continuation
    \item scrollable view of columns and layers of the DAT running the compiled program, displaying each cell as a box containing a row for each cell attribute
    \item a selectable set of watch-registers that are displayed symbolically in each cell of a layer
    \item dynamically built tooltips to see all registers of a cell row
    \item locus of attention $\alpha(N)$ for each cell $N$  (\ref{ex:a}c.i)
    \item currently-active prompt, with generated and gold continuation
    \item scrollable view of the running QKVL code (by production)
\end{itemize}

\subsubsection{Experiments\label{sec:exptsDAT}}
We tested our DAT implementation of the \PG\ algorithm using the following prompts: \\
\newline
\vspace{\baselineskip}
\noindent
\begin{tabular}{|l|p{8cm}|p{4cm}|}
\hline
Name & Prompt           & Gold Continuation \\
\hline
tgt 1 & Q the big bear ? a red car A the big bear \$ . Q some baby cub ? one small house A & some baby cub \$ . \\ \hline
tgt 2 & Q - his green bird A his green bird . Q - some light monkey A & some light monkey . \\ \hline
j loves m & Q john loves mary A mary hugs john . Q sue loves bill A & bill hugs sue . \\ \hline
act2pass & Q x V y A y Was V By x . Q u V w A & w Was V By u . \\ \hline
act2pass 2                       & Q x V y A y Was V -en By x . Q u1 u2 V w1 w2 w3 A &  w1 w2 w3 Was V -en By u1 u2 . \\ \hline
swap & Q B V D E A D E V B . Q F G H V K L A & K L V F G H . \\ \hline
cross mult & Q ( x / y ) /// (( u // v )) A ( x * v ) / (( y ** u )) . Q ( a / b ) /// (( c // d )) A & ( a * d ) / (( b ** c )) . \\ \hline
implica & Q x =\textgreater\ y A y or not x . Q ( u and v ) =\textgreater\ z A & z or not ( u and v ) . \\ \hline
implica 2 & Q x = \textgreater\ y A y or not x . Q ( u and v ) = \textgreater\ z A & z or not ( u and v ) .  \\ \hline
\end{tabular}

\noindent
The correct completion sequence for all of these prompts was predicted by the DAT.%
\footnote{
\textit{Note regarding} cross mult: It is currently not possible to give the template in the natural form \c{blue}{\Q\ ( x / y ) / (u / v ) \A\ ( x * v ) / ( y * u )}. 
Multiple distinct symbols for parentheses and division --- \c{blue}{(}, and \c{blue}{((}, etc. --- must be used because, in the current initial state of work in the TPF, the space \F\ of ICL functions studied constrains prompts so that every symbol type occurs in only one field type within a single Q- or A-region (\ref{ex:Fspec}c.i), and each field type occurs only once within a given Q- or A-region (\ref{ex:Fspec}b.i).
This is also manifest in (\ref{ex:pr3}).
Note that `**' is a variant of `*' intended to denote multiplication, not exponentiation.
} %
In addition to the above examples,\textit{ DAT achieved 100\% accuracy} when tested on the first 1000 examples from each of the following splits of the \texttt{1\_shot\_rlw} task: \texttt{test}, \texttt{ood\_lexical}, \texttt{ood\_cons\_len\_7}, and \texttt{ood\_cons\_count\_7}.
Recall that learned models struggled with these OOD tests (Sec.~\ref{sec:tgttraining}).

\section{Universality of the Framework: TPF is Turing-Complete    \label{sec:Turing}}
How general is the framework we have developed?
Specifically, we return to the question raised in Sec.~\ref{sec:univ}: how general is the class of functions that can be computed by PSL programs and thereby with DAT transformer networks?%
\footnote{We thank Rick Lewis for pointing the way towards the results presented here.}
By the classic theory-of-computation definition of `computable', any computable function $f$ can be computed by some Turing Machine TM$_f$, specified by a machine-instruction table governing the evolution of the machine's internal finite-state controller, which, conditioned by the current control state and the symbol on the tape currently being read by the machine's read/write head, writes a symbol and moves the head one position left or right.
We adopt the particular TM formalism in which each cell in this table (corresponding to a specific state/row, symbol/column pair [$\sigma_0, S_0$]) is an instruction of the form (\ref{ex:TMcell}). 

\pagebreak

\begin{exe}
   \ex Turing-table instruction \label{ex:TMcell}
   \begin{xlist}
       \ex If the machine state is $\sigma_0$ and the symbol at the current head position is $S_0$,
       \ex then write the symbol $S_1$, change the state to $\sigma_1$, and move the head on the tape one position in direction $\delta$ (= `L' or `R'; left or right).
   \end{xlist}
    \textit{Abbreviated:}  $\sigma_0, \: S_0 \Rightarrow \sigma_1, \: S_1, \: \delta$
\end{exe}

\subsection{DAT(TM) Implementation of a Given TM   \label{sec:DATTMf}}
Once we see how to express a TM-table instruction in PSL, TPF lets us compile a PSL program implementing all TM$_f$'s instructions into DAT(TM$_f$): a DAT that exactly emulates TM$_f$.
We first show how the general instruction (\ref{ex:TMcell}) can be expressed as a sequence of five productions in PSL.%
Treating the entire instruction table of TM$_f$ as a list of such instructions (arbitrarily ordered), and translating each instruction into the corresponding five productions, gives a PSL program for computing $f$.
The entire set of production is one large repeat block: productions keep applying until none are able to apply.
The TM tape is realized as the sequence of cells in the Production System Machine PSM (Sec.~\ref{sec:algI}), one cell of the PSM for each cell of the tape.
The three state variables we use in the PSM are given in (\ref{ex:TMvars}).

\begin{exe}
    \ex State variables for PSM realizing a Turing Machine
    \label{ex:TMvars}
    \begin{xlist}
        \ex $s[N]$ = current symbol-type in tape position $N$
        \ex $c[N] = 0$ if $N$ is \textit{not }the current head position (otherwise 1, L, or R).
        \ex $\sigma[N]$ = current TM state (for all $N$)
    \end{xlist}
\end{exe}

As illustrated in (\ref{ex:TMex}) below, the Turing-table instruction (\ref{ex:TMcell}) [$\sigma_0, \: S_0 \Rightarrow \sigma_1, \: S_1, \: \delta$] is translated into the following five productions; if the direction of head movement $\delta$ is L,
\textit{Production} $P_2$(L) will be activated; otherwise, $P_2$(R).
Here, writing values to state variables is not autoregressive --- the PSM cell being updated, $N$, is wherever the TM head is located (wherever $c[N] \neq 0$ at the time of update) --- and attention is not causal --- crucially, $n > N$ holds often; e.g., see (\ref{ex:P2L}).

\begin{exe}
    \ex \textit{Production P_1.} In the currently active tape position, update the state and symbol
    \label{ex:P1}
    \begin{xlist}
        \ex Condition: $c[N] == 1, \sigma[N] == \sigma_0, s[N] == S_0$ 
        \ex Action: set $c[N] := \delta, \sigma[N] := \sigma_1, s[N] := S_1$ 
    \end{xlist}
\end{exe}

\begin{exe}
    \ex \textit{Production P_2}(L). Move head left, update state there 
    \label{ex:P2L}
    \begin{xlist}
        \ex Condition: $c[n] ==$ L, $p[n] == p[N]@pos\_increment$
        \ex Action: set $c[N] := 1, \sigma[N] := \sigma[n] $ 
    \end{xlist}
\end{exe}

\begin{exe}
    \ex \textit{Production P_2}(R). Move head right, update state there 
    \label{ex:P2R}
    \begin{xlist}
        \ex Condition: $c[n] ==$ R, $p[n] == p[N]@pos\_decrement$
        \ex Action: set $c[N] := 1, \sigma[N] := \sigma[n] $ 
    \end{xlist}
\end{exe}

\begin{exe}
    \ex \textit{Production P_3.} Remove moved-head mark%
    \footnote{The Condition utilizes an option in the PSL syntax described in (\ref{ex:in}), where ``$x[n]$ in [L, R]'' does the work of two productions, identical except that in one production, the Condition includes ``$x[n] ==$ L'' while in the other, it includes ``$x[n] ==$ R''.
    In the QKV Machine, this can be achieved in a single layer in which the \textsf{key} is 2-hot, with a 1 at the loci of both \sf{L} and \sf{R}, allowing a `perfect match' with a \textsf{query} containing either $s : L$ or $s : R$. 
    (The normalization of \textbf{k} however needs to count only 1 possible match, not 2, since a perfect match is \textit{either} \sf{L} \textit{or} \sf{R}, not both, in \textbf{q}.)
    }
    \label{ex:P3}
    \begin{xlist}
        \ex Condition: $c[N]$ in [L, R]
        \ex Action: set $c[N] := 0$ 
    \end{xlist}
\end{exe}

\begin{exe}
    \ex \textit{Production P_4.} Broadcast new state globally 
    \label{ex:P4}
    \begin{xlist}
        \ex Condition: $c[n] == 1$ 
        \ex Action: set $\sigma[N] := \sigma[n] $ 
    \end{xlist}
\end{exe}

The input to the PSM is determined as follows.
The sequence of symbols $s[N]$ in the input to the PSM is the sequence of symbols on the input tape of TM$_f$.
The TM-state variable $\sigma[N]$ is set to the TM's start state, for all positions $N$.
If the TM head starts at position $N_0$ on the TM tape, in the initial state of the PSM, $c[N]$ is initialized to 1 for $N = N_0$ and to 0 for all other $N$. 

A minimal illustration of the operation of a single TM instruction --- on a tape containing only 3 positions, with the head in the middle position --- is provided in (\ref{ex:TMex}). 

\begin{exe}
    \ex Example of a Turing-Machine update 
    \label{ex:TMex}
    \medskip \\
   TM-table-cell instruction: $\sigma_0, \: \sf{B} \Rightarrow \sigma_1, \: \sf{X}$, R
    \begin{xlist}
        \smallskip
        \ex 
        \begin{tabular}{*{4}{c|}l}
        \cline{2-4} 
        $s$ & \sf{A} & \sf{B} & \sf{C} & start \\
        $\sigma$ & $\sigma_0$ & $\sigma_0$ & $\sigma_0$ & \\
        $c$ & 0 & 1 & 0 & \\
        \cline{2-4}
        \end{tabular}
        \smallskip
        \ex 
        \begin{tabular}{*{4}{c|}l}
         & & $N$ & & $P_1$ \\
        \cline{2-4} 
        $s$ & \sf{A} & \sf{X} & \sf{C} &  \\
        $\sigma$ & $\sigma_0$ & $\sigma_1$ & $\sigma_0$ & \\
        $c$ & 0 & R & 0 & \\
        \cline{2-4}
        \end{tabular}
        \smallskip
        \ex
        \begin{tabular}{*{4}{c|}l}
         & & $n$ & $N$ & $P_2$(R) \\
        \cline{2-4} 
        $s$ & \sf{A} & \sf{X} & \sf{C} &  \\
        $\sigma$ & $\sigma_0$ & $\sigma_1$ & $\sigma_1$ & \\
        $c$ & 0 & R & 1 & \\
        \cline{2-4}
        \end{tabular}
        \smallskip
        \ex
        \begin{tabular}{*{4}{c|}l}
         & & $N$ &  & $P_3$ \\
        \cline{2-4} 
        $s$ & \sf{A} & \sf{X} & \sf{C} &  \\
        $\sigma$ & $\sigma_0$ & $\sigma_1$ & $\sigma_1$ & \\
        $c$ & 0 & 0 & 1 & \\
        \cline{2-4}
        \end{tabular}
        \smallskip
        \ex
        \begin{tabular}{*{4}{c|}l}
         & $N$ & $(N)$ & $n (N)$ & $P_4$ \\
        \cline{2-4} 
        $s$ & \sf{A} & \sf{X} & \sf{C} &  \\
        $\sigma$ & $\sigma_1$ & $\sigma_1$ & $\sigma_1$ & \\
        $c$ & 0 & 0 & 1 & \\
        \cline{2-4}
        \end{tabular}
    \end{xlist}
\end{exe}

Thus we have:

\begin{exe}
    \ex $DAT(TM_f)$ Theorem: TM$_f$ in DAT weights
    \label{ex:DAT(TM)} \\
    DAT(TM$_f$) implements TM$_f$ using a repeat block of five layers for each cell of the TM$_f$ instruction table.
\end{exe}

In DAT(TM$_f$), the TM program for $f$ is implemented in the weights; next we see how such a program can be implemented instead in activations: a kind of ICL.

The universality result (\ref{ex:DAT(TM)}) establishes that, despite the focus here on the particular Templatic Generation Task defined in Sec.~\ref{sec:fun}, the TPF has potentially broad relevance for mechanistic interpretation of transformer computation (see Sec.~\ref{sec:enhancingTf}).

\subsection{DAT(UTM): DAT Implementation of a Universal TM   \label{sec:DATUTM}}
The approach presented in Sec.~\ref{sec:DATTMf} could be used to implement a TM that is universal: one that can take as input a tape that includes the TM table for computing a function $f$ as well as an argument string \textbf{\textit{s}}, and can write on the tape the value $f(\textbf{\textit{s}})$.
However, the random-access memory capability of the TPF allows a natural way to directly construct a more efficient Turing-Universal version of the DAT: DAT(UTM).
For DAT(UTM), the cells of the PSM are used, like the tape of any universal TM, to store the TM table for computing $f$ as the initial segment of the prompt, followed by the argument string $\textbf{\textit{s}}$ as the remainder of the prompt.
The completion generated by the machine is then $f(\textbf{\textit{s}})$.
This is the TM analog of the use of ICL for TGT, with the prefix of the prompt encoding a TM program rather than a text-generation template.
The analogy is even closer for the task of Nested Function Evaluation presented below (\ref{ex:NFE}).

In detail, for computing $f(\textbf{\textit{s}})$, the prompt is specified as in (\ref{ex:UTM}).
Each single entry in the TM$_f$ table is stored in the initial state-variable values of a single cell of the PSM (these are never overwritten); these state variables are $\Sigma0, \Sigma1$, \textit{S0}, \textit{S1}, and $\Delta$.
The only deviation from the standard DAT architecture defined above is that the prefix of the initialization (the input layer) --- where the instructions of TM$_f$ are written --- assigns values to state-variables other than $s$ and $p$; the remainder of the input layer --- the suffix of the prompt --- is set as usual for PSM, assigning values only to state-variables $s$ and $p$ to provide the argument symbol string $\textbf{\textit{s}}$ for computing $f(\textbf{\textit{s}})$.

\begin{exe}
    \ex Universal Turing-Machine input prompt for PSM(UTM) for computing $f(\textbf{\textit{s}})$
    \label{ex:UTM}
    \begin{xlist}
        \ex Prefix (program)   
        \begin{xlist}
            \ex For each TM$_f$ instruction: $\sigma_0, \: S_0 \Rightarrow \sigma_1, \: S_1, \: \delta$ (\ref{ex:TMcell}),
            \ex assign initial state-variable values of a single cell of the PSM prompt: \\
            \mbox{\hspace{1cm}} $\Sigma0 : \sigma_0 , \;$ \textit{S0} : S$_0, \; \Sigma1 : \sigma_1$, \textit{S1} : S$_1, \; \Delta : \delta$
        \end{xlist}
        \ex Suffix (data) \\
        Encode the argument-string $\textbf{\textit{s}}$ as usual, assigning the state-variable $s[n]$ the value of the symbol type for the $n^{\textrm{th}}$ position $p[n]$ in \textit{s}.
        \end{xlist}
\end{exe}

We then use the same productions as for the fixed-function TM$_f$ in the previous section (\ref{ex:P1} -- \ref{ex:P4}), except that the one production that encodes the content of a TM$_f$ table instruction, $P_1$, is replaced by the production $U_1$ (\ref{ex:U1}) which `looks up' the relevant instruction parameters within the activations in the current prompt rather than having those parameters built into the weights of all the layers implementing $P_1$ for each of the instructions of the TM$_f$ table.
\begin{exe}
    \ex \textit{Production U_1.} Same as \textit{Production P_1} (\ref{ex:P1}) but with states, symbols, and head-movement-direction looked-up from the prompt-prefix cell $n$ encoding the TM instruction in which the current-state and current-symbol parameters $\Sigma0, S$0 match those of the currently-updating prompt-suffix cell $N$ (emulating the current TM-head position)
    \label{ex:U1}
    \begin{xlist}
        \ex Condition: $c[N] == 1, \: \Sigma0[n] == \sigma[N], \: $ \textit{S0}$[n] == s[N]$ 
        \ex Action: $\sigma[N] := \Sigma1[n], \: s[N] := S\textit{1}[n], \: c[N] := \Delta[n]$
    \end{xlist}
\end{exe}

This establishes a second universality result for TPF:%
\footnote{During the finalization of this paper, Schuurmans et al.~(\citeyear{schuurmans2024autoregressive}) presented the construction of a Universal Turing Machine using a pre-trained LLM rather than a hand-programmed transformer (and 2027 rather than 5 productions).
It's possible that their method for inducing an LLM to execute formal productions given in the prompt might be adapted to the type of UTM-emulation-by-ICL presented here.
}

\begin{exe}
    \ex $DAT(UTM)$ \textit{Theorem: TM in DAT activations}
    \label{ex:DAT(UTM)} \\
    DAT(UTM) is a version of DAT that implements a UTM in a single repeat block of five layers.
\end{exe}

\section{Discussion and Future Work: Mechanistic Interpretation and Enhancement of Transformers \label{sec:fut}}
\subsection{What have we Learned?  \label{sec:lessons} \label{sec:disc}}
A number of findings were previewed in the theoretical summary (\ref{ex:correspondences}), which we reprise here, adding pointers to the relevant discussion in the paper.

\begin{exe}
    \ex How can ICL in a transformer perform symbolic templatic text generation? 
    \label{ex:corrRefs}
    
    Via the following [transformer element] $\sim$ [symbolic element] correspondences: 
    \begin{xlist}
        \ex a cell's residual stream $\sim$ a variable-value structure (\ref{ex:SSSvis})
        \begin{xlist}
            \ex a subspace of the residual stream $\sim$ a state variable
            \ex a vector component within a variable's subspace $\sim$ a value of that variable 
        \end{xlist}

\pagebreak

        \ex a layer's internal connections $\sim$ a production  
        (\ref{ex:PSL2QKVL} -- \ref{ex:DATLtable})
        \begin{xlist}
            \ex query-key matching in attention $\sim$ evaluating the condition of the layer's production
            \ex value vectors $\sim$ the production's action        
            \ex query-key matching on a subspace corresponding to a goal $\sim$ conditional branching for goal-directed action (Sec.~\ref{sec:combin})
        \end{xlist}
        \ex a nested set of structural variables $\sim$ hierarchical data structure (\ref{ex:PRmatrix})
       \begin{xlist}
            \ex sharing the value of a level-$l$ structural variable $\sim$ in the same (type of) level-$l$ constituent
            (adopted from Hinton, 2023\nocite{hinton2023represent})
       \end{xlist}
       \ex  a sequence of structural-variable values (at the `field' level) $\sim$ the sequence of abstract roles defining a template (\ref{ex:SWfull})
    \end{xlist}
\end{exe}

A high-level summary of our results, with pointers to relevant sections of the paper, is given in (\ref{ex:DATprops}).

\begin{exe}
    \ex A type of transformer --- a Discrete-Attention-only Transformer (DAT) can: \label{ex:DATprops}
    \begin{xlist}
        \ex encode a cell state as a vector-embedded structure that encodes the values for a set of symbolic state-variables (Sec.~\ref{sec:implReps}); 
        \ex employ this same structure for queries, keys and values (\ref{ex:QKVLang}); 
        \ex encode, in a cell’s state, abstractions like the role of a symbol in a parse of the prompt, via a set of structural variables;
        encode a hierarchical parse structure in the state of an entire layer (Sec.~\ref{sec:SWPSM}); 
        \ex use simple weight matrices to generate \textbf{q, k, v} vectors that implement symbolic productions: condition/action rules that read and write values of state variables (Sec.~\ref{sec:QKVL2DATL});
        \ex implement a \textbf{higher-level symbolic programming language for transformer computation}, PSL (Sec.~\ref{sec:algI});
        \ex implement programs in this language which parse the prompt, e.g., as a template (Sec.~\ref{sec:PARSE});
        \ex implement programs in this language for generating text that is sensitive to this parse structure, including a type of ICL (Sec.~\ref{sec:GENPSL}). 
    \end{xlist}
\end{exe}

\subsection{Mechanistic Analysis of Symbol Processing in Trained Transformer Models   \label{sec:LMs} }
The TPF results on how to construct a transformer to perform templatic text generation through ICL --- answering our questions (\ref{ex:ourQs}) --- suggests many hypotheses about how \textit{trained} transformers perform this task; these hypotheses are all precisely specified and so formally testable.
The hypotheses fall into two groups: those based on the specific algorithm for TGT, \PG, that we have programmed into our system, Sec.~\ref{sec:SpecHypTGT}, and those based on the general ICL framework formalized by TPF, Sec.~\ref{sec:GenHypICL}.
The universality results of Sec.~\ref{sec:Turing} encourage us to apply these latter hypotheses to networks trained for tasks other than TGT --- e.g, LMs.

\subsubsection {Specific Hypotheses Concerning TGT\label{sec:SpecHypTGT}}

\begin{exe}
    \ex Specific \PG-algorithm-derived hypotheses about how trained transformers perform TGT \label{ex:LMhypTGT}
    \begin{xlist}
        \ex Performing the operation of a single production --- two classes of testable hypotheses for each of 29 productions $P$:
        \label{ex:LMhypTGTProd}
        \begin{xlist}
            \ex attention pattern: the loci of some head's maximal attention for symbols in some layer are fit by the attention pattern produced by $P$;
            \ex attribute vectors: the \textbf{i, q, k, v} vectors of some head in some layer, when projected to an appropriate subspace, are fit, under an appropriate linear transformation, by the corresponding attribute vectors implementing $P$ in the DAT.
        \end{xlist}
        \ex Encoding the parse:  
        \label{ex:LMhypTGTparse}
        \begin{xlist}
            \ex For each constituent, on some subspace of the residual stream, the projection on that subspace is the same for all symbols within that constituent.
            \ex  For each constituent type, on some subspace of the residual stream, the projection on that subspace is the same for all symbols within constituents of that type.
        \end{xlist}
    \end{xlist}
\end{exe}

(\ref{ex:LMhypTGTProd}-i) can be used as a kind of screening method for picking out which attention heads at which layers are candidates for interpretation as a given production. 
For (\ref{ex:LMhypTGTProd}-ii), there is already publicly-available software to perform the
analysis. 
Exploiting the framework of Tensor Product Representations (TPRs, \TPR), the \textsc{Discover} package \cite{mccoy2019rnns} can take as input a set of hidden vectors from a `target model' under analysis, as well as a hypothesis about the symbol structures that these vectors embed: here, a set of roles (variables) that have been assigned fillers (values).
In the \textsc{Discover} technique (the `decoder' variant, McCoy, 2022, Sec.~7.3.1\nocite{mccoy2022PhD}), a trainable auxiliary neural network takes the hidden vector to be interpreted and decomposes it according to the symbol structure hypothesized as its interpretation: this auxiliary network learns a vector embedding for the hypothesized roles and their possible fillers, and a linear transformation to map the target hidden vector to a space of the dimensionality of a TPR constructed using those vector embeddings.
Optimization adjusts these embeddings, and the linear transformation, so that the structure resulting from decomposition best approximates the hypothesized symbol structure.
If the result is a good approximation, then the hypothesis is confirmed and, given the hypothesized role:filler structure for any target hidden vector, we have a closed-form equation for a TPR which is the component of the hidden state lying in the subspace targeted by the \textsc{Discover} interpretation. 

The closed-form equation for the hidden vector as a TPR makes it possible to manipulate that vector by subtracting a role:filler pair's contribution to that hidden vector and replacing it with the appropriate contribution for a different filler for that role, with the objective of making a controlled alteration of the output of the target model, thereby testing the causal role of the symbolic structure hypothesized as the interpretation of the hidden state.
For example, in Soulos et al. (\citeyear{soulos2020discovering}), studying the SCAN compositionality task, the output, initially correct for an original input \textsf{jump twice after run left}, changes to a new output that would be correct for \textsf{jump thrice after run left}, when the contribution of \textsf{twice} to the hidden vector is replaced with the contribution that would have been made by \textsf{thrice};  multiple such alterations can be performed in sequence to ultimately end up with the output appropriate for, say, \textsf{walk thrice after look right}.
That is, the TPR approximation gave a purely formal analysis of the target model's hidden encoding vector that provided a causally efficacious decomposition: altering the constituents in the internal encoding caused the model to decode it into just the novel output that is appropriate for the given altered constituents.

This same technique could be used to test the $29 \times 4$ fully-precisely-specified hypotheses stated in (\ref{ex:LMhypTGT}a-ii).
Note that the `decoder' variant of \textsc{Discover} finds, within the representational space of the target vector being interpreted, a subspace that can be linearly mapped to a TPR --- the TPR need not explain the entire representational space, which is appropriate here because we expect our algorithm's variable:value pairs to explain only part of the variance in the hidden vectors; the hidden vector may well encode many other properties as well: a LM's internal representation, for example, will likely contain considerable semantic information along with the semantics-free  information processed in our purely symbolic algorithm which provides the hypotheses (\ref{ex:LMhypTGT}a-ii).
This is particularly expected because even a trained network that does encode something like our productions will not dedicate an entire layer to a single production; successfully trained transformers of Sec.~\ref{sec:tgttraining} perform TGT with many fewer layers than the 29 deployed in our one-production- and one-head-per-layer compiled DAT.

The type of encoding structure hypothesized in (\ref{ex:LMhypTGTProd}) has in fact been identified 
in pre-trained LLMs performing variable binding \cite{davies2023discovering,wu2025how}, entity tracking \cite{feng2023language,dai2024representational}, and belief-tracking \cite{prakash2025language} tasks.
For instance, Feng \& Steinhardt~(\citeyear{feng2023language}) consider tasks like `P lives in WA. G lives in MD. Where does P live?'
The example region of the prompt creates a binding problem, where P's residence must be bound to WA, G's residence must be bound to MD, and accurately answering `where does $x$ live?' requires retrieving the correct binding for $x$.
Each of the two facts in the prompt involves a separate binding of person to place; Feng \& Steinhardt found that the two facts were in essence assigned two identifiers (analogous to `fact\_1', `fact\_2') and the entities in each fact were bound to their corresponding identifier in the residual stream --- by addition, with a particular orthogonal subspace dedicated to encoding the identifiers. 
These `identifier labels' are embedded in the residual stream just as are our `region labels', to which they directly correspond.

\subsubsection {General Hypotheses Concerning ICL\label{sec:GenHypICL}}
Rather than testing the specific \PG-derived hypotheses realized in our DAT for TGT (\ref{ex:LMhypTGT}), we can reformulate the hypotheses in more general terms and use existing methods to discover structures hidden within a trained model performing another ICL task.
\begin{exe}
    \ex General TPF-framework-derived hypotheses about how trained transformers perform ICL \label{ex:LMhypICL}
    \begin{xlist}
        \ex For some head in some layer, when projected to an appropriate subspace, the trained model's \textbf{i, q, k, v} vectors are fit, under an appropriate linear transformation, by an unknown (to-be-discovered) set of state-variable:value structures 
        \begin{xlist}
            \ex In this variant of (\ref{ex:LMhypTGTProd}), the variable:values are unknown structures to be discovered by analysis
            \ex  Here we can use an extension of the \textsc{Discover} technique in which there is no predetermined human-generated hypothesis for the role:filler structure of the target hidden states under analysis. 
            A second auxiliary network called \textsc{Role} is used to learn to assign roles and fillers for the interpretation of a target hidden state: here, the variables and values of a state structure.
            In its original application \cite{soulos2020discovering}, 
            this was successfully applied to an encoder-decoder RNN solving the \textsc{Scan} compositional-generalization task: \textsc{Role} learned to assign one of a set of initially meaningless roles to each symbol in an input sequence, such that the hidden vector produced by the target RNN's encoder network was accurately modeled as a linearly-transformed TPR built from: a learned set of vector embeddings for the roles; a learned set of vector embeddings for the input symbols; and a learned linear transformation of the resulting TPR.

        \end{xlist}
        \ex For each constituent, on some subspace of the residual stream, the projection on that subspace is the same for all symbols within the same constituent of an unknown (to-be-discovered) parse tree;  
        \begin{xlist}
            \ex In this variant of (\ref{ex:LMhypTGTparse}), the parse is not hypothesized a priori, but discovered empirically by the analyst, as in the work of Murty et al.~(\citeyear{murty2022characterizing,murty2023grokking}).
            \ex As mentioned in Sec.~\ref{sec:SpecHypTGT}, Feng \& Steinhardt (\citeyear{feng2023language}) discovered a case of this in which the constituents are distinct facts given in the prompt.
        \end{xlist}
        \ex  At some layer and for some head, there is systematic matching between subspaces of queries and keys at positions systematically linked by the trained transformer's attention mechanism; across different prompts, as the query varies, the key varies congruently: 
        \begin{xlist}
            \ex under an orthogonal transformation $\mathbb{T}$,%
            \footnote{$\mathbb{T}$ maps the subspace for register $y$ to the subspace for register $x$ (or $x$\bp).}
            implementing a to-be-discovered production's Condition (``$x[n] == y[N]$''):  $\mathbf{k}[n] = \mathbf{q}[N] \, \mathbb{T}$ on appropriately chosen subspaces of the $\mathbf{k, q}$ vector spaces ---
            \ex possibly with an orthogonality-imposing transformation $\mathbb{N}$ intervening: $\mathbf{k}[n] = \mathbf{q}[N] \, \mathbb{T \, N}$, (analogous to the one's-complement operation implementing ``$x[n]$ != $y[N]$''; $\mathbb{N}$ can be any anti-symmetric matrix, $\mathbb{N}^\top = - \mathbb{N}$).
        \end{xlist}
            \ex When decomposed using the subspaces discovered in (\ref{ex:LMhypICL}c), some weight matrices $\mathbb{W}_{\mathbf{q,k,v}}$ contain embedded orthogonal block sub-matrices, analogous to the embedded identity matrices $\mathbb{I}$ of DAT for copying values between variables (or an embedded $\mathbb{N}$ matrix for negating a variable).            
    \end{xlist}
\end{exe}

\medskip
\subsection{Enhancing TPF and Transformer Architectures \label{sec:xxx}}
Of the many directions for possible future research building on the work presented here, we mention one that significantly extends the framework TPF itself (Sec.~\ref{sec:compRec}), two that pursue how features of the standard transformer architecture might be incorporated into the DAT (Sec.~\ref{sec:enhancingDAT}), and three that apply what we have learned from TPF to suggest potential improvements to the standard transformer architecture (Sec.~\ref{sec:enhancingTf}). 

\subsubsection{ Extending the TPF: Composition and Recursion\label{sec:compRec}}
The task, TGT, defined by the functional-level description of the system studied here (\ref{ex:Fspec}), is to identify a single template, illustrated by a single example at the beginning of the input prompt, and to generate text by instantiating that template with new symbol strings filling its slots.
The powerful symbolic-computation operation of \textit{embedding} would, in this context, consist of multiple templates in a prompt, some embedded within the slots of others. 
To pursue this we are examining a variation of the task in which templates are replaced by symbolic-function definitions: in place of \sf{\Q\ swap x y \A\ y x} we have \mbox{\sf{swap ( x , y ) $\rightarrow$ y x}}. This extended task, \textit{Nested Function Evaluation} (\textit{NFE}), naturally allows embedding of functions/templates within others, as in the recursive example in (\ref{ex:NFE}).
\begin{exe}
    \ex Nested Function Evaluation: NFE 
    \label{ex:NFE}
    
    \textit{Prompt:}
    \\ \mbox{\hspace{4mm}}\sf{swap ( x , y ) $\rightarrow$ y x ; twice ( x ) $\rightarrow$ x x ; swap ( twice ( a b ) , swap ( c , d ) ) $\rightarrow$} 

    \vspace{2pt}
    \textit{Continuation:} 
    \\ \mbox{\hspace{4mm}}\gsf{swap ( twice ( a b ) $\rightarrow$ a b a b , swap ( c , d ) $\rightarrow$ d c ) $\rightarrow$ swap ( a b a b , d c ) }
    
\hspace{21mm} 
    \gsf{$\rightarrow$ d c a b a b}  
\end{exe}

Preliminary results of current work show that the methods developed here for TGT extend naturally to NFE. TPF can employ productions that cycle between two behaviors which we term \textit{copy} and \textit{eval}. The \textit{copy} behavior operates similarly to \CF: it simply repeats symbols from the prompt that are too complex to be resolved immediately. This process continues until a segment of the prompt is found which can be resolved using the templates already specified (e.g. \sf{twice ( a b )} in the example above). At this point the model switches to the \textit{eval} behavior, resolving the nested function according to the examples in the prompt. By iterating between these two behaviors, DAT can resolve complex functions through their simple constituents and thereby accommodate the two remaining capabilities of symbolic computation not captured by TGT itself: composition (\ref{ex:compos}) and recursion (\ref{ex:recur}). 
Presuming that these preliminary results hold up, the insights from the modestly-extended TPF suffice to understand how \textit{all} the fundamental compositional elements of symbolic computation (\ref{ex:props}) can be naturally embedded in the continuous numerical computation of which transformer-style networks are capable.

\subsubsection {Enhancing the DAT Architecture\label{sec:enhancingDAT}}
Certain features of the standard transformer architecture might profitably be incorporated into the DAT architecture in future work.

\begin{exe}
    \ex Enhancing DAT with architectural features of standard transformers  
    \label{ex:enhancingDAT}
    \begin{xlist}
        \ex \textit{Multiple attention heads.} A production might be implemented as one head within an extension of DAT deploying multi-headed attention. 
        When multiple productions' Conditions are met and their Actions conflict, symbolic production systems invoke a conflict-resolution procedure preventing multiple conflicting productions from acting simultaneously.
        In the multi-headed DAT, pre-programming of the production system could ensure that such conflict cannot arise, ensuring that such conflicting productions are separated into different layers, with the appropriate ordering.
        Non-conflicting productions could be implemented in different heads within a single layer.
        As long as there remains no MLP sublayer in the DAT, there is little computational difference between productions being implemented in multiple heads within a single layer or as single heads across multiple layers.
        \ex \textit{MLPs.} In our TGT formalization, there is no need to modify symbols or identify relations between them other than equality, so we have no need of MLPs in our DAT. 
        Of course it is generally presumed that MLPs play a large role in LMs and the work here does not speak to their role.
        Many analogy-inspired ICL tasks do incorporate symbol modification, and assume prior knowledge of symbol mappings, as in \sf{\Q\ x y z \A\ X Y Z} or \sf{\Q\ p q r s \A\ t u v w}.
        Addition of symbol-mapping sublayers into the DAT (e.g., an MLP) would open the door to studying this larger class of interesting ICL tasks and developing a general framework that correspondingly extends the version of TPF presented here.
    \end{xlist}
\end{exe}

\subsubsection {Enhancing the Transformer Architecture\label{sec:enhancingTf}}
The following considerations relating our DAT to trained transformers encourage several possible directions for future work on enhancing standard transformer architectures.

\begin{exe}
    \ex Enhancing standard transformers with architectural features of DAT
    \label{ex:enhancingTf}
    \begin{xlist}
        \ex \textit{Repeat blocks.}
        While DAT lacks the MLP and multi-headedness of standard transformers, it has an architectural feature lacking in standard transformers, repeat blocks.
        The power of such structure has been explored in the Universal Transformer \cite{dehghani2018universal}, the Looped Transformer \cite{fan2024looped,giannou2023looped} in which all layers together form a single repeat block, and adaptively at scale in Mixture-of-Recursions \cite{bae2025mixture}. 
        \ex \textit{Discretization.} \label{ex:discr} DATmax and DATnorm could provide important inductive biases for transformers learning to perform ICL in something like the way our discretely-functioning DAT does.
        \begin{xlist}
             \ex \textit{DATmax} discretizes the processing (attention) to the retrieval of values from a single (by default, leftmost) locus (\ref{ex:a}c.i).
             Csord\'as et al.~(\citeyear{csordas2021neural}) have shown substantial improvements in learning embedding-depth generalization on algorithmic tasks by similarly constraining attention.
             \ex \textit{DATnorm} discretizes the data to a single value per variable (\ref{ex:a}c.iv). DATnorm imposes a disciplined use of the residual stream to encode discretely-valued state variables --- a TPR, not necessarily deploying one-hot embeddings (see \TPR; this generalizes the `disentangled residual stream' notion of Friedman et al., 2023\nocite{friedman2023learning}). 
        \end{xlist}         
        These features of discreteness (\ref{ex:discr}) could be inserted into standard transformers softly, as biases implemented as regularization terms added to the loss function.
      
        \ex \textit{Residual-stream recurrence.} During generation, the input to a column of the DAT is not just the single `next symbol' predicted by the previous column: it is the full residual stream at the top of the previous column. 
        This allows direct propagation of computed features of tokens such as role in a parse (e.g., \textsf{field}); such recurrence gives much power to RNNs and can potentially do the same for enhanced transformer architectures --- at the cost of complicating (or preventing) training by teacher forcing (see also Fan et al., 2020\nocite{fan2020addressing}).%
        \footnote{
        Note that, unlike the task for standard transformer LMs, the generation task in TGT is deterministic, and our production system results in a unique symbol being assigned non-zero generation probability; thus symbol-generation in a given column does not need to know which of multiple potential symbols was randomly sampled for generation in the previous column. 
        }
    \end{xlist}
\end{exe}

\subsection{Unifying Semantics-Free and Semantics-Permeated Knowledge} \label{sec:unif}

As mentioned in Sec.~\ref{sec:sem-free}, the ultimate goal of this research program is to develop architectures that intimately unify both (i) TPF's type of formal, semantics-free processing and (ii) standard transformer processing.
As merely an illustration of the large space of possibilities for unification, here we present just one simple approach.
We propose a transformer architecture part of which is biased (through regularization terms in the loss function) towards DAT-like processing; the rest of the network is a typical transformer with no such bias.

\begin{exe}
    \ex Integrating DAT and standard-transformer processing: The Semi-formal Transformer architecture 
    \begin{xlist}
        \ex The residual stream $V$ is the direct sum of two subspaces, the \textit{biased} subspace $V_b$ and the \textit{unbiased} subspace $V_u$.
        \ex There are two types of attention heads: biased heads $H_b$ and unbiased heads $H_u$.
        \ex Only biased heads can write into $V_b$ (i.e., only they can have value vectors with non-zero components in $V_b$): $V_b$, like DAT, is constrained to host a TPR with subspaces encoding symbolic values for variables; writing into the residual stream is performed only by biased heads that use DATnorm to enforce the constraint that registers have a unique symbolic value.
        \ex All heads can read from the entire residual stream, i.e., queries and keys can contain vector components in both $V_b$ and $V_u$.
        \ex Attention for biased heads is subject to a regularization  term in the loss function that biases the model to weights that approximate DATmax, focusing attention towards a single source.
        \ex Layers contain an MLP sublayer that operates exclusively on $V_u$. Any LayerNorm operations that may be present also apply only on $V_u$.  
    \end{xlist}
\end{exe}

In this Semi-formal Transformer architecture, $V_b$ is processed by compositionality-respecting operations, but control decisions selecting what operations to perform are carried out by unconstrained neural processes: this division of labor has proved to be very effective as embodied in the Differentiable Tree Machine \cite{soulos2023differentiable,soulos2024compositional}.

For a particular task domain, TPF shows how we could use PSL to pre-program the $H_b$ with domain-useful, abstract formal operations; the unbiased portion of the model then uses powerful neural computation to determine how to deploy these formal operations.
A promising application domain for this new architecture is the processing of semi-formal documents, which feature natural-language descriptions of properties joined with logical combinators, such as rules for what constitutes an acceptable comment for an online forum, which must not have objectionable property A or B and must contain certain required elements C and D.
Standard LM heads determine whether the NL-described properties A--D are true or false of a target comment, writing these values into $V_b$, and formal heads are programmed to evaluate AND, OR and NOT combinations, given the LM-estimated truth values of these properties that they read from $V_b$.

\subsection{Wrapping Up} \label{sec:wrap}

The neurocompositionality hypothesis \cite{smolensky2022neurocompositional,smolensky2022neurocompositionalTR} proposes that human-level intelligence arises from a deep synergy between two principles previously viewed as incompatible: representations have compositional structure that drives their processing, and representations lie in a continuous space supporting similarity-based generalization and gradient-based learning.
We suspect that the astounding success of transformer-based AI results from these systems satisfying the neurocompositionality hypothesis.

The Transformer Production Framework (TPF) presented here (Secs.~\ref{sec:fun} -- \ref{sec:impl}) shows with complete precision and completeness how a characteristically powerful ability of transformers, templatic text generation through in-context learning (Box~\ref{box:ICL1}; Sec.~\ref{sec:Fdef}), can result from neural networks deploying representations that are continuous and compositional, and processing that, while numerical and continuous, exploits the compositional structure of the representations: these networks implement a kind of abstract program that has been a central model of human higher-level cognition --- a production system (\ref{ex:PS}).
This production-system level of description endows the networks that implement it with great power and generality --- it is Turing Universal (Sec.~\ref{sec:Turing}).

In addition to enabling hand-programming to create a fully mechanistically-interpretable concrete model capable of powerful ICL, TPF provides a rich set of precise, formally-testable hypotheses about the mechanisms that power ICL in transformers that are trained from data (Sec.~\ref{sec:LMs}).
If even a fraction of these hypotheses are confirmed in future work, significant light will have been shed on the inner workings of the black boxes powering contemporary generative AI.

Finally, while a number of the mechanisms deployed in this paper have been discussed to some extent in the literature --- e.g., sparse/discrete attention \cite{entmax,  meister2021sparseattentioninterpretable}, repeat blocks of layers \cite{csordas2021neural,csordas2024moeut} and structured residual streams \cite{mak2025residualmatrixtransformersscaling}, the TPF framework presented may prove beneficial for architecture design by allowing us to bridge the gap between theory and practice.  
Researchers can design tasks that require a certain form of abstract symbolic reasoning, and obtain an existence proof for their solvability by writing the generalizing program and obtaining the corresponding compiled DAT model. 

This opens up a number of interesting questions. If we are unable to write a generalizing program without a specific architectural component, is the same true of learned models? If we are able to write the generalizing program, but cannot correspondingly learn it --- why not? The TPF framework can be uniquely beneficial here, as it allows us to study potential solutions at an extremely granular level (down to the individual attention head if needed), and thereby bring theory and empirical validation far closer than might otherwise be possible, giving us a principled basis on which to study what components, and what combinations of them, contribute towards generalization.

As an example, both geometric attention (realized in TPF through DATmax) and sharp attention (realized in TPF through DATnorm) have separately been the subject of considerable study. However, the way they combine in the TPF framework (strict condition match followed by positional tie breaker) unlocks a new form of reasoning which had not, to the best of our knowledge, been discussed in the literature at time of writing. An attention mechanism based on --- and inspired by --- this combination has since been released \cite{opper2025trabetterlengthgeneralisation}, and enables novel forms of generalization, which are trivial for the DAT, but had previously been unattainable by contemporary transformer variants.

\section*{Acknowledgments}\label{sec:ack}
For insightful comments and suggestions we wish to thank members of the Johns Hopkins Neurocompositional Computing research group, the Yale Computational Cognitive Science group, and at Microsoft Research (MSR) Redmond, the Deep Learning, RiSE and SIGLR groups, especially Matthai Philipose and Nikolaj Bj$\o$rner; we also thank the participants and organizers of the 2024 Weinberg Institute Symposium on Cognitive Science at the University of Michigan \cite{smolensky2024weinberg}, especially Rick Lewis for discussion of the universality of PSL.
Thanks too to Geoff Hinton for discussion of encoding hierarchical structure in transformers, to Tal Linzen for discussion of syntactic performance of LLMs, and to Tom McCoy for discussion of the decoder version of DISCOVER. For valuable contributions to online resource release we are grateful to Swadheen Shukla at MSR.
Finally we acknowledge crucial support of internships at Microsoft for authors M.~Opper and A.~Davies. 

\bibliography{NeurocompArxivNew.bib}
\bibliographystyle{plainnat}

\appendix

\section{A Walk-Through of the Symbolic Computation Implicit in the \swap\ Task}
\label{sec:appWalk}

We now describe the role of this structure in the definition and execution of the task \swap, identifying the essential roles in the algorithm of the key properties of symbolic computation spelled out in (\ref{ex:props}).
In this informal section, we will make free use of the properties of the function class \F\ providing the functional-level description of our TPF system (\ref{ex:Fspec}): these properties are formally defined below in (\ref{ex:Fspec}), and (\ref{ex:parsing1}) indicates how particular steps in the algorithm are justified by that formal definition.

\begin{exe}
    \ex \swap\ instance (\ref{ex:SWinst}) full (\color{blue} prompt \color{black} + \color{orange}continuation \color{black}) structure: Target
    \label{ex:SWfull} \\
    \footnotesize 
    \resizebox{\linewidth}{!}{%
    \Tree 
    [.S 
        [.X 
            [.QR 
                [.FQ 
                    [.\underline{v}_{FQ} 
                        [.\sf{Q} ]]]
                [.F1 
                    [.v_{F1} 
                        \sf{B} \sf{C} ] ] 
                [.FV  
                    [.\underline{v}_{FV} 
                        [.\sf{V} ]]] 
                [.F2  
                    [.v_{F2} 
                        \sf{D} \sf{E} ]]] 
            [.AR 
                [.FA 
                    [.\underline{v}_{FA} 
                        [.\sf{A} ]]]
                [.F2 
                    [.v_{F2} 
                        \sf{D} \sf{E} ] ] 
                [.FV  
                    [.\underline{v}_{FV} 
                        [.\sf{V} ]]] 
                [.F1  
                    [.v_{F1} 
                        \sf{B} \sf{C} ]]]] 
        [.C 
            [.QR 
                [.FQ 
                    [.\underline{u}_{FQ} 
                        [.\sf{Q} ]]]
                [.F1 
                    [.u_{F1} 
                        \sf{F} \sf{G} ] ] 
                [.FV  
                    [.\underline{u}_{FV} 
                        [.\sf{V} ]]] 
                [.F2  
                    [.u_{F2} 
                        \sf{J} \sf{K} \sf{L} ]]] 
            [.AR 
                [.FA 
                    [.\underline{u}_{FA} 
                        [.\sf{A} ]]]
                [.\color{orange}F2 
                    [.\color{orange}u_{F2} 
                        \color{orange}\textsf{J} \color{orange}\textsf{K} \color{orange}\textsf{L} ] ] 
                [.\color{orange}FV  
                    [.\color{orange}\underline{u}_{FV} 
                        [.\color{orange}\textsf{V} ]]] 
                [.\color{orange}F1  
                    [.\color{orange}u_{F1} 
                        \color{orange}\textsf{F} \color{orange}\textsf{G} ]]]]]
    }
\end{exe}

\noindent Within the Q-region QR of the example X (subconstituent QR within constituent X), the substrings of the template (\ref{ex:SWtempl}) --- \sf{Q}, \textit{x}, \sf{V}, \textit{y} --- are treated as four subconstituents we will call \textit{fields}: they are labelled FQ, F1, FV, F2.
The \textit{delimiter fields} FQ and FV alternate in linear position with the \textit{constituent fields} F1 and F2.
The A-region AR contains the same fields as QR, but with F1 and F2 swapped, and with the initial field now FA rather than FQ.

The field structure found in the Q-region QR of the example string is identical to that of the QR in the continuation-cue string C (subconstituent QR within constituent C), but the values of F1 and F2 are different strings.
These new field values are then inserted into a copy of the field structure of the example A-region (AR within X) to generate the output, completing the A-region of the continuation (AR within C).

The previous paragraph is a high-level overview of the algorithm in our TPF framework that  will generate the continuation using the complete parsed template structure (\ref{ex:SWfull}): \PG.
This will be decomposed into an initial \textit{parsing algorithm} PARSE which generates the portion of the tree in (\ref{ex:SWfull}) that dominates the input provided by the prompt (blue symbols).
This will be followed by a \textit{generation algorithm} GEN that uses this parse to build the remainder of (\ref{ex:SWfull}) (the orange symbols). 
We begin by walking through the parsing algorithm.
\subsection{Parsing Algorithm PARSE: Informal Walk-Through \label{sec:parsalg}}

A first step towards building the \swap\ structure (\ref{ex:SWfull}) is parsing the prompt into four regions, shown in (\ref{ex:SWreg}): in order, they are the Q-region of the example, the A-region of the example, the Q-region of the cue, and the A-region of the cue, which contains only the symbol \sf{\A} initially, but will be extended by the generation algorithm. 
The start of each region is marked by a reserved delimiter symbol, either \sf{\Q} or \sf{\A}.
This illustrates property (\ref{ex:hier}); S is composed of two subconstituents of type X and C; the X and C subconstituents are in turn each composed of two subsubconstituents of type QR and AR.
Thus we have two distinct tokens of both type QR and type AR: property (\ref{ex:types}).
The beginning of each of the two type-QR constituents is signalled by a token of the atomic Q type (\ref{ex:symbols}); it is important that these two symbols are recognizably tokens of the same type: property (\ref{ex:ident}).
Throughout this section, we identify [in square brackets] the properties of symbolic computation listed in (\ref{ex:props}) which are crucial to performing this task.
We see that nearly all the properties in (\ref{ex:props}) are demanded by the task, showing that it provides a strong test of symbolic computational abilities.

\begin{exe}
    \ex \swap\ instance prompt (\ref{ex:SWprompt}): region structure
    \label{ex:SWreg} \\
    \footnotesize 
    \Tree 
    [.S 
        [.X 
            [.QR                  
                \sf{Q} 
                \sf{B} \sf{C}  
                \sf{V} 
                \sf{D} \sf{E} ]
            [.AR                  
                \sf{A} 
                \sf{D} \sf{E}  
                \sf{V} 
                \sf{B} \sf{C} ]]
        [.C 
            [.QR                  
                \sf{Q} 
                \sf{F} \sf{G}  
                \sf{V} 
                \sf{J} \sf{K} \sf{L} ] 
            [.AR 
                \sf{A} ]]]
\end{exe}

A next step is parsing the Q-region into fields, shown in (\ref{ex:SWXfield}); as already mentioned, these are the substrings into which the region must be divided to capture the templatic pattern (\ref{ex:SWtempl}) which the generation process will later need to fill. 
Each substring is taken to be the value assigned to a variable, the field that it fills.
The delimiter \sf{\Q} is the value of the delimiter field FQ.
\sf{V} is also a delimiter, the value of the field variable FV:\textit{ all delimiter fields have a constant value throughout the input structure} S.
That \sf{V} repeats in the Q-region of the cue signals that it is a delimiter: in the task we study, non-delimiter fields must change value between the example and the continuation-cue.
FV separates the two argument slots of the template: \textit{x} and \textit{y} in (\ref{ex:SWtempl}).
The first of these is the field F1, with value \sf{B C}; the second is F2, with value \sf{D E}.
These non-delimiter fields will be called \textit{constituent} fields: their values are not fixed; they are the open slots in the template, which take different values in the cue and in the example.

\begin{exe}
    \ex \swap\ instance prompt (\ref{ex:SWprompt}): example region field structure (\underline{delimiters} underlined)
    \label{ex:SWXfield} \\
    \footnotesize 
    \Tree 
    [.S 
        [.X 
            [.QR 
                [.FQ 
                    [.\underline{v}_{FQ} 
                        [.\sf{Q} ]]]
                [.F1 
                    [.v_{F1} 
                        \sf{B} \sf{C} ] ] 
                [.FV  
                    [.\underline{v}_{FV} 
                        [.\sf{V} ]]] 
                [.F2  
                    [.v_{F2} 
                        \sf{D} \sf{E} ]]] 
            [.AR 
                [.FA 
                    [.\underline{v}_{FQ} 
                        [.\sf{A} ]]]
                [.F2 
                    [.v_{F2} 
                        \sf{D} \sf{E} ] ] 
                [.FV  
                    [.\underline{v}_{FV} 
                        [.\sf{V} ]]] 
                [.F1  
                    [.v_{F1} 
                        \sf{B} \sf{C} ]]]] 
        [.C 
            [.QR                  
                \sf{Q} 
                \sf{F} \sf{G}  
                \sf{V} 
                \sf{J} \sf{K} \sf{L} ] 
            [.AR 
                \sf{A}
                ]]]
\end{exe}

In order that their boundaries be well-defined, constituent fields in the example Q-region must be separated by delimiter fields (which we mark by underlining).
Thus \sf{B C} and \sf{D E} are the values of constituent fields, which we are calling F1 and F2; they are separated by the delimiter field FV.
\sf{B C} is labeled as the value of F1 by being grouped under the node v$_{F1}$; likewise for \sf{D E} and v$_{F2}$.
That \sf{B C} and \sf{D E} are the values of constituent fields is signalled by their appearance as substrings of both the Q- and the A-regions of the example: these symbol-sequences retain their identities --- are detectable as identical --- across the different regions in which they appear (\ref{ex:ident}).
The values of constituent fields vary across inputs, but within a given Q/A pair within a single input --- i.e., separately within each of X and C --- these fields have the same values.
Like the Q-region of the example, the A-region of the example has v$_{F1} =$ \sf{B C} as the value of F1, and v$_{F2} =$ \sf{D E} as the value of F2: but the ordering of fields is different in the Q- and A-regions of the example.
These field sequences are respectively the subconstituent role sequences characteristic of constituents of type QR and AR [property (\ref{ex:roleset})].

Since the parsing of the Q-region of the example identifies the sequence of roles (here, fields) characteristic of type-QR constituents, the Q-region of the cue can be parsed using the same field sequence: this is shown in (\ref{ex:SWpromptfield}).

\begin{exe}
    \ex \swap\ instance prompt (\ref{ex:SWprompt}): complete field structure
    \label{ex:SWpromptfield} \\
    \footnotesize 
    \Tree 
    [.S 
        [.X 
            [.QR 
                [.FQ 
                    [.\underline{v}_{FQ} 
                        [.\sf{Q} ]]]
                [.F1 
                    [.v_{F1} 
                        \sf{B} \sf{C} ] ] 
                [.FV  
                    [.\underline{v}_{FV} 
                        [.\sf{V} ]]] 
                [.F2  
                    [.v_{F2} 
                        \sf{D} \sf{E} ]]] 
            [.AR 
                [.FA 
                    [.\underline{v}_{FQ} 
                        [.\sf{A} ]]]
                [.F2 
                    [.v_{F2} 
                        \sf{D} \sf{E} ] ] 
                [.FV  
                    [.\underline{v}_{FV} 
                        [.\sf{V} ]]] 
                [.F1  
                    [.v_{F1} 
                        \sf{B} \sf{C} ]]]] 
        [.C 
            [.QR 
                [.FQ 
                    [.\underline{u}_{FQ} 
                        [.\sf{Q} ]]]
                [.F1 
                    [.u_{F1} 
                        \sf{F} \sf{G} ] ] 
                [.FV  
                    [.\underline{u}_{FV} 
                        [.\sf{V} ]]] 
                [.F2  
                    [.u_{F2} 
                        \sf{J} \sf{K} \sf{L} ]]] 
            [.AR 
                [.FA 
                    [.\underline{u}_{FA} 
                        [.\sf{A} ]]]
                ]]]
\end{exe}

\noindent Given the sequence of fields of the Q-region of the cue, we can identify the values filling those fields.
The delimiter fields FQ, FV are matched to their fixed string values \sf{Q}, \sf{V} (the same in the cue as in the example); and \textit{the substrings between those delimiters are parsed as the new values of the constituent fields F1, F2 in the cue}: these are u$_{F1}$ = \sf{F G}, u$_{F2}$ = \sf{J K L}.
[Illustrating property (\ref{ex:roleset}), constituents of type QR have a characteristic sequence of roles (here,  `fields'): FQ F1 FV F2, whether embedded within an X or a C constituent (\ref{ex:ident}). 
Constituents of type AR have the role-sequence: FA F2 FV F1, not only in the given AR embedded within X, but also in the to-be-generated AR embedded within C.]
\subsection{Generation Algorithm GEN: Informal Walk-Through  \label{sec:GenAlg}}
Having identified the field sequence of an AR constituent from the AR within the example, the same sequence populates the AR in the cue: (\ref{ex:SWfield}) [properties (\ref{ex:ident}) and (\ref{ex:roleset})].
\\
\begin{exe}
    \ex Continued \swap\ instance (\ref{ex:SWinst}): field structure
    \label{ex:SWfield} \\
    \footnotesize 
    \resizebox{\linewidth}{!}{%
    \Tree 
    [.S 
        [.X 
            [.QR 
                [.FQ 
                    [.\underline{v}_{FQ} 
                        [.\sf{Q} ]]]
                [.F1 
                    [.v_{F1} 
                        \sf{B} \sf{C} ] ] 
                [.FV  
                    [.\underline{v}_{FV} 
                        [.\sf{V} ]]] 
                [.F2  
                    [.v_{F2} 
                        \sf{D} \sf{E} ]]] 
            [.AR 
                [.FA 
                    [.\underline{v}_{FQ} 
                        [.\sf{A} ]]]
                [.F2 
                    [.v_{F2} 
                        \sf{D} \sf{E} ] ] 
                [.FV  
                    [.\underline{v}_{FV} 
                        [.\sf{V} ]]] 
                [.F1  
                    [.v_{F1} 
                        \sf{B} \sf{C} ]]]] 
        [.C 
            [.QR 
                [.FQ 
                    [.\underline{u}_{FQ} 
                        [.\sf{Q} ]]]
                [.F1 
                    [.u_{F1} 
                        \sf{F} \sf{G} ] ] 
                [.FV  
                    [.\underline{u}_{FV} 
                        [.\sf{V} ]]] 
                [.F2  
                    [.u_{F2} 
                        \sf{J} \sf{K} \sf{L} ]]] 
            [.AR 
                [.FA 
                    [.\underline{u}_{FA} 
                        [.\sf{A} ]]]
                \color{orange}F2 
                \color{orange}FV 
                \color{orange}F1 ]]]
    }
\end{exe}

Within a single QR/AR pair, each field has the same value in both the QR and the AR subconstituents, so the new values of the  fields identified in the QR of the cue --- u$_{F1}$ and u$_{F2}$ (along with the fixed value of the delimiter field FV) --- must serve as the values of those same fields in the completion's AR (\ref{ex:SWfull2}) [illustrating (\ref{ex:roleseq})].

\begin{exe}
    \ex \swap\ instance (\ref{ex:SWinst}) full structure: Generated
    \label{ex:SWfull2} \\
    \footnotesize 
    \resizebox{\linewidth}{!}{%
    \Tree 
    [.S 
        [.X 
            [.QR 
                [.FQ 
                    [.\underline{v}_{FQ} 
                        [.\sf{Q} ]]]
                [.F1 
                    [.v_{F1} 
                        \sf{B} \sf{C} ] ] 
                [.FV  
                    [.\underline{v}_{FV} 
                        [.\sf{V} ]]] 
                [.F2  
                    [.v_{F2} 
                        \sf{D} \sf{E} ]]] 
            [.AR 
                [.FA 
                    [.\underline{v}_{FQ} 
                        [.\sf{A} ]]]
                [.F2 
                    [.v_{F2} 
                        \sf{D} \sf{E} ] ] 
                [.FV  
                    [.\underline{v}_{FV} 
                        [.\sf{V} ]]] 
                [.F1  
                    [.v_{F1} 
                        \sf{B} \sf{C} ]]]] 
        [.C 
            [.QR 
                [.FQ 
                    [.\underline{u}_{FQ} 
                        [.\sf{Q} ]]]
                [.F1 
                    [.u_{F1} 
                        \sf{F} \sf{G} ] ] 
                [.FV  
                    [.\underline{u}_{FV} 
                        [.\sf{V} ]]] 
                [.F2  
                    [.u_{F2} 
                        \sf{J} \sf{K} \sf{L} ]]] 
            [.AR 
                [.FA 
                    [.\underline{u}_{FA} 
                        [.\sf{A} ]]]
                [.\color{orange}F2 
                    [.\color{orange}u_{F2} 
                        \color{orange}\textsf{J} \color{orange}\textsf{K} \color{orange}\textsf{L} ] ] 
                [.\color{orange}FV  
                    [.\color{orange}\underline{u}_{FV} 
                        [.\color{orange}\textsf{V} ]]] 
                [.\color{orange}F1  
                    [.\color{orange}u_{F1} 
                        \color{orange}\textsf{F} \color{orange}\textsf{G} ]]]]]
    }
\end{exe}

\noindent This determines the continuation string \gsf{J K L V F G}.

If the continuation string is to be produced one symbol at a time --- as in the transformer implementation in our TPF system --- generating the first symbol \gsf{J}\ involves generating the next field, \g{F2}, and unbinding it [property (\ref{ex:unbinding})] to get the first symbol of its value string in the \textit{Q-region of the cue}, \g{u$_{F2}$}; this is then bound to the new instance of \g{F2}\ in the \textit{A-region of the cue} [property (\ref{ex:binding})].
This is the first of our two generation operations: starting the next field, \NF.
The second operation, continuing a field that has already been initiated --- \CF\ --- generates the next symbol in the current field: \gsf{K}. 
Choosing the appropriate operation is a case of conditional branching conditioned on representational structure [property (\ref{ex:condBranch})].

\CF\ is in fact the operation performed by an \textit{induction head} \cite{olsson2022context}: predict that the symbol following some symbol $\Sigma$ in the continuation will be of the same type as the symbol following the most recent previous token matching the type of $\Sigma$.
While it is known how to implement \CF\ in a transformer using induction attention heads, \NF\ is a considerably more abstract operation: predict that the \textit{field} following $\Sigma$ in the continuation will be of the same type as the \textit{field} following the previous occurrence of $\Sigma$'s field type --- within the \textit{example A-region}, not the \textit{cue Q-region}.
Given the next field type, generating the next symbol requires determining the value string for that field type --- within the \textit{cue} Q-region, not the \textit{example} Q-region.

To generate the next continuation symbol, the generation algorithm we give below (see also Sec.~\ref{sec:GEN}) decides whether to apply \NF\ or \CF; for this, it must determine whether the value string for the field currently being generated has been completed: if so, \NF\ is called for; otherwise, \CF\ is needed.
This is a structure-sensitive choice [property  \ref{ex:condBranch}].
The results of processing the subconstituents (fields) of AR with \NF\ and \CF\ are composed together in the field-sequence prescribed by the parse structure of AR [property \ref{ex:compos}].

For \NF\ to generate the next field in the AR of C under construction, the last-generated field must be matched with a field in the AR of the example X; the field following that matching field within X gives the field type that needs to be generated next in the continuation.

\section{PSL and QKVL Programs for Templatic Parsing and Generation with the \PG\ Algorithm}\label{sec:appPSLalg}

\begin{exe}  
    \ex PARSE  
     \label{tbl:parse}
\end{exe}

\begin{center}
\tiny
\begin{longtable}[htbp]{p{7mm} @{}p{3cm}| p{2.5cm} p{2.3cm} | p{12mm} p{10mm} @{}p{10mm} }
\hline
& & \multicolumn{2}{| c |}{PSL Production} & \multicolumn{3}{| c }{QKVL instructions} \\
\cline{3-7}
P\# & gloss & Condition & Action & q & k & v\\
 \hline \hline
 & & \multicolumn{1}{| l |}{\textbf{all have parse[N] == 1}} & 
 & \multicolumn{1}{ l |}{\squeeze{.95}{\textbf{all have a:a}}} 
 & \multicolumn{1}{ l |}{\squeeze{.95}{\textbf{all have a:1}}} &
\\
\cline{3-3}
\cline{5-6}

\endhead
\hline
\endfoot

0 & initialize region, type; \hfill\break field = position &    
    position[N] == position[N] &
    \begin{tabular}{@{}l}
        region[N] = R\_INIT\\
        type[N] = T\_INIT\\
        field[N] = position[N]\\
        prev\_position[N] = 0  \\      
        index[N] = 1
    \end{tabular}
& p:p & p:p & r:R, t:T, f:p, p*:0, d:1
\\
\hline

pre-1 & set prev\_position and \hfill\break prev\_symbol &    
    position[n] == \hfill\break\mbox{position[N]@pos\_decrement}   &
    \begin{tabular}{@{}l}
        prev\_position[N] = \\ \mbox{\hspace{2mm}position[n]} \\
        prev\_symbol[N] =  \\ \mbox{\hspace{2mm}symbol[n]}
    \end{tabular}
& p\bp:p@pos\_ decrement & p\bp:p & p*:p, s*:s
\\
\cline{3-7}

1a & start Example (at p:0): \hfill\break r:XQ, t:D, f:FQ, d:0 &    
    \textsf{prev\_position[N] == 0} &
    \begin{tabular}{@{}l}
        region[N] = XQ \\
        type[N] = D \\
        field[N] = FQ \\
        index[N] = 0 \\
    \end{tabular}
& p*:p*, p:p & p*:0, p:p & r:XQ, t:D, f:FQ, d:0
\\
\cline{3-7}

1b & start Cue (at 2nd Q): \hfill\break r:CQ, t:D, f:FQ &    
   \begin{tabular}{@{}l}
        symbol[n] == symbol[N] \\
        position[n] == 1  \\
        position[N] != 1 
    \end{tabular} &
    \begin{tabular}{@{}l}
        region[N] = CQ \\
        type[N] = D \\
        field[N] = FQ \\
    \end{tabular}
& s$^\backprime$:s, p\bp:1, p:p & s\bp:s, p\bp:p, p!=1 & r:CQ, t:D, f:FQ
\\
\hline

&   \textit{start repeat block} \\
\cline{2-2}

pre-2a & set prev\_region &    
    position[n] == \hfill\break\pn   &
    prev\_region[N] =\hfill\break \mbox{\hspace{2mm}region[n]} &
    p\bp:p@pos\_ decrement & p\bp:p & r*:r
\\
\cline{3-7}

2a & propagate r:XQ to 1st delimiter (t:D; starts r:CQ) &
   \begin{tabular}{@{}l}
       prev\_region[N] == XQ \\
       region[N] == R\_INIT
   \end{tabular}    &
   region[N] = XQ
& r*:r*, r:r, p:p & r*:XQ, r:R, p:p  & r:XQ
\\
\cline{2-2}
&   \textit{end when NO\_CHANGE}  \\
\hline

&   \textit{start repeat block} \\
\cline{2-2}

pre-2b & set prev\_region &    
    position[n] == \hfill\break\pn   &
    prev\_region[N] =\hfill\break \mbox{\hspace{2mm}region[n]} &
    p\bp:p@pos\_ decrement & p\bp:p & r*:r
\\
\cline{3-7}

2b & propagate r:CQ to input end &
   \begin{tabular}{@{}l}
       prev\_region[N] == CQ \\
       region[N] == R\_INIT
   \end{tabular}    &
   region[N] = CQ
& r*:r*, r:r, p:p & r*:CQ, r:R, p:p  & r:CQ
\\
\cline{2-2}
&   \textit{end when NO\_CHANGE}  \\
\hline

3a & start r:XA (f:FA) at A in current r:XQ &
    \begin{tabular}{@{}l}
        symbol[N] == A \\
        region[N] == XQ
    \end{tabular}    &
    \begin{tabular}{@{}l}
        region[N] = XA \\
        type[N] = D \\
        field[N] = FA
    \end{tabular}      
& s:s, r:r, p:p & s:A, r:CQ, p:p & r:XA, t:D, f:FA
\\
\cline{3-7}

3b & start r:CA (f:FA) at A in current r:CQ &
    \begin{tabular}{@{}l}
        symbol[N] == A \\
        region[N] == CQ
    \end{tabular}    &
    \begin{tabular}{@{}l}
        region[N] = CA \\
        type[N] = D \\
        field[N] = FA
    \end{tabular}      
& s:s, r:r, p:p & s:A, r:XQ, p:p & r:CA, t:D, f:FA
\\
\cline{3-7}

\hline
&   \textit{start repeat block} \\
\cline{2-2}

pre-4 & set prev\_region &    
    position[n] == \hfill\break\pn   &
    prev\_region[N] =\hfill\break \mbox{\hspace{2mm}region[n]} &
    p\bp:p@pos\_ decrement & p\bp:p & r*:r
\\
\cline{3-7}

4 & propagate XA to 1st t:D (starts r:CQ) &
   \begin{tabular}{@{}l}
       prev\_region[N] == XA \\
       region[N] == XQ  \\
       type[N] == T\_INIT
   \end{tabular}    &
   region[N] = XA
 & r*:r*, r:r, t:t, p:p & r*:XA,t:T, r:XQ,  p:p & r:XA
\\
\cline{2-2}
&   \textit{end when NO\_CHANGE}  \\
\hline

5a & symbol in XQ later repeated in CQ: set t:D  &
   \begin{tabular}{@{}l}
       region[n] == CQ \\
       region[N] == XQ  \\
       symbol[n] == symbol[N] 
   \end{tabular}    &
   type[N] = D
& r\bp:CQ, r:r, s\bp:s & r\bp:r, r:XQ, s\bp:s & t:D
\\
\cline{3-7}

5b & symbol in CQ that repeats from XQ: set t:D, same field  &
   \begin{tabular}{@{}l}
       region[n] == XQ \\
       region[N] == CQ  \\
       symbol[n] == symbol[N] 
    \end{tabular}    &
    \begin{tabular}{@{}l}
        type[N] = D \\
        field[N] = field[n]
    \end{tabular}
& r\bp:XQ, r:r, s\bp:s & r\bp:r, r:CQ, s\bp:s & t:D, f:f
\\
\cline{3-7}

5c & symbol in XA that repeats from CQ: set t:D, same field  &
    \begin{tabular}{@{}l}
       region[n] == CQ \\
       region[N] == XA  \\
       symbol[n] == symbol[N] 
    \end{tabular}    &
    \begin{tabular}{@{}l}
        type[N] = D \\
        field[N] = field[n]
    \end{tabular}
& r\bp:CQ, r:r, s\bp:s & r\bp:r, r:XA, s\bp:s & t:D, f:f
\\
\hline

6 & identical untyped symbols in X have the same C field  &
    \begin{tabular}{@{}l}
        symbol[n] == symbol[N] \\
        region[n] == XQ  \\
        region[N] == XA  \\
        type[N] == T\_INIT
    \end{tabular}    &
    \begin{tabular}{@{}l}
        type[N] = C \\
        field[N] = field[n] 
    \end{tabular}
& s\bp:s, r\bp:XQ, r:r, t:t & s\bp:s, r:XA, r\bp:r, t:T & t:C, f:f
\\
\hline

7 & unset types in XA are delimiters  &
   \begin{tabular}{@{}l}
        region[N] == XA  \\
        type[N] == T\_INIT
    \end{tabular}    &
   type[N] = D
& r:r, t:t, p:p & r:XA, t:T, p:p & t:D
\\
\cline{3-7}

7' & remaining unset types are constituents  &
   type[N] == T\_INIT  &
   type[N] = C
& t:t, p:p & t:T, p:p & t:C
\\
\hline

pre-8 & set prev\_region, prev\_type, \hfill\break prev\_field &    
    position[n] == \hfill\break\pn   &
    \begin{tabular}{@{}l}
       prev\_region[N] = \\ \mbox{\hspace{2mm}region[n]} \\
       \squeeze{0.95}{prev\_type[N] = type[n]}\\
     \squeeze{0.95}{prev\_field[N] = field[n]}
    \end{tabular}    
& p\bp:p@pos\_ decrement & p\bp:p & r*:r, t*:t, f*:f
\\
\cline{3-7}

8 & field sequence is the same in XQ and CQ &
    \begin{tabular}{@{}l}
        prev\_region[n] == XQ \\
        region[n] == XQ \\
        prev\_type[n] == D \\
        type[n] == C \\
        region[N] == CQ \\
        prev\_type[N] == D \\
        type[N] == C \\
        prev\_field[n] == prev\_field[N]
    \end{tabular}    &
    field[N] = field[n]     
& r*\bp:XQ, r\bp:XQ, t*\bp:D, t\bp:C, r:r, t*:t*, t:t, f*\bp:f* 
& r*\bp:r*, r\bp:r, t*\bp:t*, t\bp:t, r:CQ, t*:D, t:C, f*\bp:f* 
& f:f
\\

\hline
&   \textit{start repeat block} \\
\cline{2-2}

pre-9 & set prev\_field &    
    position[n] == \hfill\break\pn   &
    prev\_field[N] =\hfill\break \mbox{\hspace{2mm}field[n]} &
    p\bp:p@pos\_ decrement & p\bp:p & f*:f
\\
\cline{3-7}

9 & constituent fields change only at t:D &
   \begin{tabular}{@{}l}
       prev\_type[N] == C \\
       type[N] == C  
   \end{tabular}    &
    field[N] =\hfill\break \mbox{\hspace{2mm}prev\_field[N]}
 & t*:t*, t:t,\hfill\break p:p & t*:C, t:C,\hfill\break p:p & f:f*
\\
\cline{2-2}
&   \textit{end when NO\_CHANGE}  \\
\hline

10 & change in f $\Rightarrow$ d:0 &
   prev\_field[N] != field[N]   &
   index[N] = 0
& f*:f*, p:p & f*!=f, p:p & d:0
\\
\hline

11 & set parse = 0 at prompt's last token &
   z\_{temp}[N] == EOP   &
   parse[N] = 0
& z:z, p:p & z:EOP, p:p & a:0
\\

\hline
\hline
\end{longtable}
\end{center}
\begin{exe}    
    \ex GENERATE
    \label{tbl:gen}
\end{exe}

\begin{center}
\tiny
\begin{longtable}[htbp]{p{7mm} @{}p{3cm}| p{2.5cm} p{2.3cm} | p{12mm} p{10mm} @{}p{10mm} }
\hline
& & \multicolumn{2}{| c |}{Production} & \multicolumn{3}{| c }{QKV instructions} \\
\cline{3-7}
G\# & gloss & Condition & Action & q & k & v\\
 \hline \hline
 & & \multicolumn{1}{| l |}{\textbf{all have parse[N] == 0}} & 
 & \multicolumn{1}{ l |}{\squeeze{.95}{\textbf{all have a:a}}} 
 & \multicolumn{1}{ l |}{\squeeze{.95}{\textbf{all have a:0}}} &
\\

\cline{3-3}
\cline{5-6}
\endhead

\hline
\endfoot

0 & set end = 0, x\_temp = 0 globally &
   position[N] ==\hfill\break \mbox{\hspace{2mm}position[N]} &
   \begin{tabular}{@{}l}
       end[N] = 0 \\
       x\_temp[N] = 0  
   \end{tabular}       
& p:p & p:p & e:0 
\\
\hline

pre-1 & update prev\_symbol and \hfill\break prev\_field &
   position[n] == \hfill\break\pn & 
   \begin{tabular}{@{}l}
       prev\_symbol[N] = \\ \mbox{\hspace{2mm}symbol[n]} \\
       \squeeze{0.95}{prev\_field[N] = field[n]}  
   \end{tabular}    
& p\bp:p@pos\_ decrement & p\bp:p & s*:s, f*:f
\\
\cline{3-7}

1   & 
\textbf{IF} symbol in CQ matching current symbol is not field-final \textbf{THEN} [\CF] copy next symbol in CQ, set end[N]=1, x\_temp[N]=0 to prevent application of remaining productions &
   \begin{tabular}{@{}l}
        region[n] == CQ \\
        prev\_symbol[n] == symbol[N] \\
        prev\_field[n] == field[N] \\
        index[n] != 0 
    \end{tabular} &
    \begin{tabular}{@{}l}
        end[N] = 1 \\
        x\_temp = 0 \\
        region[N] = CA \\
        \squeeze{0.95}{symbol[N] = symbol[n]} \\
        field[N] = field[n] \\
        type[N] = type[n] \\
        index[N] = index[n]
    \end{tabular}
& r\bp:CQ, s*\bp:s, f*\bp:f, d\bp!=0
    & r\bp:r, s*\bp:s*, f*\bp:f*, d\bp:d 
    & e:1, x:0, r:CA, s:s, f:f, t:t, d:d
\\
\hline

pre-2 & update prev\_field and \hfill\break prev\_region  &
   position[n] == \hfill\break\pn & 
   \begin{tabular}{@{}l}
       \squeeze{0.95}{prev\_field[N] = field[n]} \\
       prev\_region[N] = \\ \mbox{\hspace{2mm}region[n]}  
   \end{tabular}    
& p\bp:p@pos\_ decrement & p\bp:p & f*:f, r*:r
\\
\cline{3-7}

2   & \textbf{ELSE} (end[N] = 0) [\NF] find final position of field in XA matching current field, assign following field label to y\_temp[N] &
   \begin{tabular}{@{}l}
        end[N] == 0 \\
        region[n] == XA \\
        prev\_field[n] == field[N] \\
        index[n] == 0 \\
        prev\_region[n] == XA
    \end{tabular} &
    y\_temp[N] = field[n]
& e:e, r\bp:XA, f*\bp:f, d\bp:0, r*\bp:XA
    & e:0, r\bp:r, f*\bp:f*, d\bp:d, r*\bp:r* 
    & y:f
\\
\hline

3   & 
\textbf{ELSE} con't: \textbf{IF} find initial position of field y\_temp[N] in CQ, copy that symbol and its structural variables (except r) and set flag x\_temp:=1 to block P3' &
   \begin{tabular}{@{}l}
       end[N] == 0 \\
       region[n] == CQ \\
       field[n] == y\_temp[N] \\
       index[n] == 0
    \end{tabular} &
    \begin{tabular}{@{}l}
        x\_temp[N] = 1 \\
        region[N] = CA \\
        \squeeze{0.95}{symbol[N] = symbol[n]} \\
        field[N] = field[n] \\
        type[N] = type[n] \\
        index[N] = index[n]
        \end{tabular}
& e:e, r\bp:CQ, f\bp:y, d\bp:0
    & e:0, r\bp:r, f\bp:f, d\bp:d 
    & x:1, r:CA, s:s, f:f, t:t, d:d
\\
\hline

3'   & 
\textbf{ELSE} (x\_temp=0): find initial position of the field y\_temp[N] in XA, copy that symbol and its structural variables (except r) &
   \begin{tabular}{@{}l}
       end[N] == 0 \\
       x\_temp[N] == 0 \\
       region[n] == XA \\
       field[n] == y\_temp[N] \\
       index[n] == 0
    \end{tabular} &
    \begin{tabular}{@{}l}
        region[N] = CA \\
        \squeeze{0.95}{symbol[N] = symbol[n]} \\
        field[N] = field[n] \\
        type[N] = type[n] \\
        index[N] = index[n]
        \end{tabular}
& e:e, x:x,r\bp:XA, f\bp:y, d\bp:0 
    & e:0, x:0, r\bp:r, f\bp:f, d\bp:d 
    & r:CA, s:s, f:f, t:t, d:d  
\\
\hline
\hline

\end{longtable}
\end{center}

\section{Further Related Work} \label{sec:appRelW}

Here we widen the discussion of related work from that initiated in Sec.~\ref{sec:rel}.

\subsection{ICL Mechanisms \label{sec:iclmechanisms}}
\textbf{Bayesian learning:} Xie et al.~(\citeyear{xie2022bayesianinference}) posit that ICL can emerge in models when pre-training documents possess long-range coherence derived from a document-level `concept'.  They study knowledge-based ICL examples with 0-64 shots that must predict a single correct output token. They explain that models must infer a latent concept from the prompt's $n$-shot input/output pairs to predict the correct output token, describing this as an implicit form of Bayesian inference. 
In our work, the `concept' providing coherence to a particular prompt is not semantically defined, but an abstract symbolic template, which governs the generation of an entire output string; this template must be inferred from each prompt's single example. We analyze the inference process involved in completing such prompts, but not the conditions enabling such a process to be learned. 
\\
\\
\textbf{Gradient descent:} Dai et al.~(\citeyear{dai2023gradientdescent}) report finding a duality between attention in transformer models and gradient descent, and posit that in-context learning can be understood as implicit fine-tuning by adjusting the model weights using attention in the feedforward inference process.  The contribution to attention from the in-context material can be viewed as an adjustment to the trained attention-governing weights, with the value vectors playing the role of back-propagated error signals in a fine-tuning update restricted to query- and key-generating weights. They study ICL classification tasks, with up to 32-shot examples.  Our work is focused on how a pre-determined algorithm for sophisticated symbol-processing can be applied using 1-shot examples to predict an entire output sequence, not just a classification label.
\\
\\
\textbf{CSCG:} Swaminathan et al.~(\citeyear{Swaminathan2023schemalearning}) view in-context learning through a different sequence-learning model called the clone-structured causal graph (CSCG), using the mechanisms of schema learning, recall, and rebinding.  They posit and confirm that similar mechanisms could exist in transformers. They study both knowledge-based classification tasks as well as abstract sequence-prediction tasks.

\subsection{Influence of Pre-Training Data \label{sec:effectspretraining}}
\textbf{Training data requirements for ICL:}
Chan et al.~(\citeyear{chan2022datadistributional}) establish 3  requirements for LM-training data to give rise to a model exhibiting ICL: data burstiness (not uniformly distributed), data with a large number of rarely occurring classes, and data with dynamic meaning.  Using the Omniglot dataset, they show that ICL usually competes with ``in-weights learning'', with one of the two winning.  

\subsection{ICL Function Learning \label{sec:iclfunclearning}}
\textbf{Simple numerical functions:} Garg et al.~(\citeyear{garg2022can}) train decoder-only transformers from scratch to perform ICL in simple numerical function classes: linear functions,  sparse linear functions, decision trees, and two-layer ReLU neural networks (using 20--40 shot examples of 20-dimensional inputs).  They show that ICL achieves performance comparable to an optimal least squares estimator, and is robust to certain types of train-test distribution shift. 
\\
\\
\textbf{Language learning}: Akyürek et al.~(\citeyear{akyurek2024icllearning}) introduce a new task for studying ICL --- stochastic languages --- and compare the performance of transformers to state space models and their variants.  The find that transformers perform best on this task, and posit that induction heads are important.  They show improvements on the other models when the equivalent of induction heads are added.
This lends credibility to the hypothesis that, for trained models, counterparts of our operations of \CF, identical to induction heads, and \NF, the abstraction of induction heads from symbol matching to syntactic category matching, may have an important role.

\subsection{ICL Learning Process \label{sec:icllearnprocess}}
\textbf{Generalization and Stability:} Li et al.~(\citeyear{li2023transformersasalgorithms}) study the generalization and stability of transformers pre-trained using multi-task learning (MTL) with $n$-shot-style examples.  Through proofs with mild assumptions, they obtain generalization bounds for ICL that depend on the stability of the transformer algorithm, showing that as the ICL prompt length increases, the ICL predictions become more stable.  They also find that the transfer risks on unseen examples depend on the number of examples and complexity of the MTL tasks, but, surprisingly, not on the complexity of the transformer architecture.
\\
\\
\textbf{ICL algorithms for linear regression:} Akyurek et al.~(\citeyear{akyurek2022whatlearningalgorithm}) study linear regression, exploring whether ICL uses one of three known algorithms to learn the linear function latent in the examples in a given ICL prompt.  They identify 4 operations over the hidden states of a transformer layer  that can be implemented in a single transformer layer (move-subvector, matrix-multiply, scalar-divide, affine-transform) and prove by construction (programming the transformer with these 4 instructions) that a transformer can emulate 3 classical solutions to linear regression: gradient descent, closed-form ridge regression, and exact least-squares regression.  They then pre-train transformers on linear regression tasks with an ICL-style objective and examples, and compare the resulting behavior to the 3 previously programmed models.  They find that their trained transformers transition between the classical algorithms as depth and dataset noise vary, and converge to optimal Bayesian decisions as the width and number of layers grow. 
\\
\\
\textbf{PAC in-context learnability:} Wies et al.~(\citeyear{wies2023learningabilityoficl}) extend the PAC framework to prove, under mild assumptions, that ICL efficiently `learns' tasks through examples.  Their pre-training data is a mixture of multiple latent downstream tasks, presented in $n$-shot-style prompts, with consistent delimiters in each prompt. They conclude that ``in-context learning is more about identifying the task than about learning it'' [p. 1].  They find polynomial sample complexity in the number of shots per prompt.  

\subsection{Improving ICL \label{sec:improvingicl}}
\textbf{Meta training}: Min et al.~(\citeyear{min2022metaicl}) fine-tuned a GPT-2 large transformer tasks from 142 NLP datasets reformatted as ICL style tasks.  This resulted in increased performance over baselines on the test set (containing 52 unique target tasks), sometimes beating models with 8x larger parameter count.  Performance increased with the number and diversity of the fine-tuning tasks.  Best performance resulted from fine-tuning with both instructions and the meta-training tasks.
\\
\\
\textbf{Meta ICL:} Coda-Forno et al.~(\citeyear{coda-forno2023metaiclinllms}) introduced a technique to improve ICL by preceding the normal $n$-shot examples of a task in the prompt with $K$ $n$-shot examples from other, related tasks.  They study this technique using the pre-trained GPT-3 model and 2 tasks: 1-dimensional regression and a 2-armed bandit.  Through analysis of the ICL activations, they show how these  $K$ additional-task examples reshape the model's prior over expected tasks.
\\
\\
\textbf{Reasoning module:} Bhatia et al.~(\citeyear{bhatia2023tart}) analyze ICL failure cases, relative to task specific fine-tuning, and posit that ICL has all the information it needs --- the right representations for the task --- but fails due to its inability to perform simple probabilistic reasoning over the representations to predict the next token.  They create a separate, task agnostic reasoning module (a decoder-only transformer), trained only on synthetic logistic regression tasks.  After training, the models are composed by feeding the output of the base model through a PCA/averaging layer and then to the reasoning module.  An additional variant called leave-one-out (LOO) embedding improves the model further.  They also demonstrate that the reasoning module is not just model- and task-agnostic, but also modality-agnostic, by using it for binary classification tasks with audio and image inputs.

\section{Templatic Generation Task (TGT) Grammar   \label{sec:tgtgrammar} }
The grammar describing Templatic Generation tasks is shown here in Backus-Naur form:

\begin{verbatim}
<tgt>               ::= <example> <cue>
<example>           ::= Q <question> A <answer>
<cue>               ::= Q <question> A 
<question>          ::= <dc sequence>
<answer>            ::= <dc sequence>

<dc sequence>       ::= <constituent list> [ <delimiter> ]
<constituent list>  ::= <constituent> [ <delimiter> <constituent list> ]

<delimiter>         ::= <symbol> [ <delimiter> ] 
<constituent>       ::= <symbol> [ <constituent> ]

Additional constraints: 
    - no symbol can be repeated within the <question>, within the <answer>, 
      or within the <cue>.
      
    - symbols in the <constituents> of the <cue> cannot overlap with  
      symbols in the <constituents> of the <question>.

\end{verbatim} 
     
\section{PSL Grammar   \label{sec:pslgrammar} }
The grammar for the PSL Language is shown here in Backus-Naur form:

\begin{verbatim}
<program>         ::= <declarations> <statements>
<declarations>    ::= <declaration> [ <declarations> ]
<declaration>     ::= <register map> | <constant map> | <system map> | <watch list>

<register map>          ::= registers "{" <register entry list> "}"
<register entry list>   ::= <register entry> [ "," <register entry list> ]
<register entry>        ::= <register name> ":" <register short name>
<register name>         ::= <unquoted string>
<register short name>   ::= <quoted string>

<constant map>          ::= constants "{" <constant entry list> "}"
<constant entry list>   ::= <constant entry> [ "," <constant entry list> ]
<constant entry>        ::= <constant name> [ ":" <quoted string> ]
<constant name>         ::= <unquoted string>

<system map>            ::= system "{" <system entry list> "}"
<system entry list>     ::= <system entry> [ "," <system entry list> ]
<system entry>          ::= <system register name> ":" <register name>
<system register name>  ::= symbol | position | output | parse | eop

<watch list>            ::= watch "[" <register name> [ "," <register name> ] "]"

<statements>   ::= <statement> [ <statements> ]
<statement>    ::=  <causal attn statement> | <where statement> | <repeat statement>

<causal attn statement> ::= causal_attn ":" <boolean value>
<boolean value>         ::= true | false

<where statement>       ::= <where variant> <conditions> ":" <assignments>
<where variants>        ::= where | where_lm | where_rm
            
<conditions>            ::= <condition> [ and <conditions> ]
<condition>             ::= <simple condition> | "(" <conditions> ")"    

<simple condition>      ::= <bool compare> | <bool in>
<bool compare>          ::= <left cond> <compare op> <right cond>   
<left cond>             ::= <register name> "[" <register index> "]"
<register index>        ::= N | n 
<compare op>            ::= "==" | "!="
<right cond>            ::= <constant name> | <right reg> [ <weight func> ]
<right reg>             ::= <register name> "[" <register index> "]" 
<weight func>           ::= "@" <weight function> 
<weight function>       ::= pos_increment | pos_decrement

<bool in>               ::= <left cond> <in op> "[" <constant list> "]"
<in op>                 ::= in | not in
<constant list>         ::= <constant name> [ "," <constant list> ]

<assignment list>       ::= <assignment> [ <assignment list> ]
<assignment>            ::= <assign left> "=" <assign right>
<assign left>           ::= <register name> "[" N "]" 
<assign right>          ::= <constant name> | <right reg>

<repeat statement>      ::= repeat <statements> until <stop condition>
<stop condition>        ::= NO_CHANGE | <conditions>
\end{verbatim}

\section{DAT Operation    \label{sec:datop} }
This appendix summarizes how the DAT Transformer operates (during inference, i.e., in-context learning; in-weights-learning has been left for future work).
The explanation takes as a starting point a conventional transformer, and describes the alterations that lead to the DAT.

\subsection{High-Level Architecture}
\begin{itemize}
    \item We start with a normal decoder-only transformer.
    \item We remove the Feedforward module in each layer.
    \item In each column of the transformer, the residual stream, and vectors for inputs, queries, keys, and values, are maintained as \texttt{n\_registers} registers, each of which is a 1-hot (or 0-hot) vector over the vocabulary symbols.
    \item We load and freeze all Multi-Head Attention (MHA) weights (DAT only operates in inference mode).
\end{itemize}

\ignore{      %
\subsection{DATnorm  \label{sec:DATnorm2}}

See \ref{sec:DATnorm} \footnote{Sometimes called \texttt{PashaMax} in the codebase.}

\begin{itemize}
\item The input is of shape: $[\text{seq\_len}, \text{n\_registers}, \text{d\_register}]$. 
\item PashaMax ensures that each register is set to the argmax 1-hot (or 0-hot) of the highest probability logit.
\end{itemize}

}    %

\subsection{Transformer-Level Operation}

\begin{itemize}
\item As part of the DAT initialization, the tensor weights (compiled from QKVL code that was compiled from PSL code) are loaded into the Q, K, V weights (matrices and bias vectors) for each layer. These weights are then frozen.

\item The `\texttt{forward()}' method of the transformer contains an additional parameter `\texttt{input embeddings}' that is used as follows:
\begin{itemize}
    \item When the given prompt is being processed (the initial prompt, before any new columns have been generated), `\texttt{input embeddings}' is set to None. 
    \item When the remainder of the prompt is being processed, `\texttt{input embeddings}' for the current 
    column is set to the output of the DAT for the previous column.
\end{itemize}

\item When the `\texttt{forward()}` method of the transformer is called, we build the embeddings for all columns as follows:
\begin{itemize}
    \item Compute the symbol embeddings using a frozen matrix that translates each vocab index into its 1-hot vector representation.
    \item Compute a ``one'' embedding (using the vocab index for the symbol ``1'').
    \item Compute an ``eop'' embedding (using the vocab index for the symbol ``EOP'').
    \item Compute ``pos'' embeddings (using the vocab index for each position number, e.g., ``42'').
    \item Initialize the ``src'' embeddings to all zeros (for all columns and all registers within each column).

    \item Finally, we set registers on the src embeddings:
    \begin{itemize}
        \item Set the ``s'' register to the symbols embeddings.
        \item Set the ``p'' register to the pos embeddings.
        \item Set the ``a'' register (for the parse flag) to the one embeddings.
        \item Set the ``z'' register for the last column to the eop embedding.
    \end{itemize}

    \item If `input\_embeddings` (from processing the prefix columns) have been passed to `forward()`, we set src embeddings to the concatenation of (src embeddings, input\_embeddings).

    \end{itemize}

    \item We process each layer of the transformer as follows:

    \begin{itemize}
    \item If the layer is the first layer of a repeat block, we capture the input value of all of the registers in each column.
    \item We process the layer using the input to produce an output.
    \item If the layer is the last layer of a repeat block, we compare the register values of the output to the register values of saved-off input, for each column. If any register of any column has changed, we continue processing with the first layer of the repeat block.
    \item If the registers have not changed, or if the current layer was not the last layer of a repeat block, we continue processing with the next layer.
    \end{itemize}
\end{itemize}

\subsection{Layer-Level Operation}
MHA is performed with the following changes:
\begin{itemize}
    \item The \texttt{softmax()} and \texttt{dropout()} of the standardly-computed attention weights are replaced by DATmax, i.e., the computation of $\alpha[N]$ given in (\ref{ex:attnWts}).
    \item Values of input, query, key, value, MHA output, and attn\_weight tensors are captured for later diagnostics and visualization (see screenshot in Sec.~\ref{sec:tfexplorer}).
    \item The standard \texttt{dropout()} and \texttt{LayerNorm()} used to add the output of MHA to the residual stream are replaced by DATnorm, as given in (\ref{ex:DATnorm}).
\end{itemize}

\section{TGT Testing Prompt Prefix   \label{sec:tgtprompt} }
When testing LLMs on our TGT dataset (Sec.~\ref{sec:expts}), the following \texttt{sys\_prompt} was prepended to each Q/A string in the dataset:

\begin{verbatim}
    You are a helpful assistant; please complete the following abstract  
    pattern exactly once. The pattern contains an example question/answer 
    pair, followed by a second question and a missing answer. Do not output  
    anything except the final answer. Pay close attention to all special 
    characters. 
\end{verbatim}

\noindent
One of the task variants also appends the TGT grammar to the above system-prompt:

\begin{verbatim}
    The grammar for these patterns can be described as follows: 
    <tgt>               ::= <example> <cue>
    <example>           ::= Q <question> A <answer>
    <cue>               ::= Q <question> A 
    <question>          ::= <dc sequence>
    <answer>            ::= <dc sequence>
    <dc sequence>       ::= <constituent list> [ <delimiter> ]
    <constituent list>  ::= <constituent> [ <delimiter> <constituent list> ]
    <delimiter>         ::= <symbol> [ <delimiter> ] 
    <constituent>       ::= <symbol> [ <constituent> ]    
\end{verbatim}

\section{LLM Testing Details    \label{sec:llmtesting} }
This appendix includes the results of exploratory testing on the GPT-4 LLM Transformer, using variants of the core task: 1\_show\_rlw.

To test the effect of different number of constituents in the core task, we varied the constituent count as shown in Table~\ref{tab:argct}.  Increasing the count had the most impact on performance of all the variants tested (ranging from 0.96 to 0.23 accuracy).  This test shows that even the best LLM starts to fail as the TGT task complexity increases.%
\footnote{
The `\texttt{ood}' in the file names here refers to test items that are out of the training distribution for models trained from scratch, discussed in \trainscratch.
These same test items are used here to test LLMs, where we of course do not control the training distribution, so `\texttt{ood}' should not be taken literally in this context. 
However, it is true that except for the \texttt{1\_shot\_eng} split, all the test items use symbols that are random-letter `words' (rlw), and these symbols may never have been encountered during LM training.
}

\begin{table}[H]
\centering
\caption{Experiment results showing accuracy with varying numbers of constituents.} \begin{tabular}{|l|l|l|l|l|l|}
\hline
\textbf{Task} & \textbf{Split} & \textbf{Con Count} & \textbf{Con Length} & \textbf{Prompts} & \textbf{Accuracy} \\ \hline
1\_shot\_rlw & test & 1 & 1 & 100 & 0.96 \\ \hline
1\_shot\_rlw & test & 2 & 1 & 100 & 0.97 \\ \hline
1\_shot\_rlw & ood\_cons\_count\_3 & 3 & 1,2,4 & 100 & 0.62 \\ \hline
1\_shot\_rlw & ood\_cons\_count\_5 & 5 & 1,2,4 & 100 & 0.35 \\ \hline
1\_shot\_rlw & ood\_cons\_count\_7 & 7 & 1,2,4 & 100 & 0.26 \\ \hline
1\_shot\_rlw & ood\_cons\_count\_10 & 10 & 1,2,4 & 100 & 0.23 \\ \hline
\end{tabular}
\label{tab:argct}
\end{table}

To test the effect of different number of symbols per constituent, we varied the constituent length as shown in Table~\ref{tab:arglen}.  Here we see a striking contrast with Table~\ref{tab:argct} --- the model performance drops only slowly as this aspect of the task complexity increases.  Notice that with the value 2 and 3, the model performed perfectly.

\begin{table}[H]
\centering
\caption{Experiment results showing GPT-4 accuracy with varying constituent lengths.} \begin{tabular}{|l|l|l|l|l|l|}
\hline
\textbf{Task} & \textbf{Split} & \textbf{Con Count} & \textbf{Con Length} & \textbf{Prompts} & \textbf{Accuracy} \\ \hline
1\_shot\_rlw & test & 1 & 1 & 100 & 0.96 \\ \hline
1\_shot\_rlw & test & 1 & 2 & 100 & 1.00 \\ \hline
1\_shot\_rlw & ood\_cons\_len\_3 & 1,2,4 & 3 & 100 & 1.00 \\ \hline
1\_shot\_rlw & ood\_cons\_len\_5 & 1,2,4 & 5 & 100 & 0.98 \\ \hline
1\_shot\_rlw & ood\_cons\_len\_7 & 1,2,4 & 7 & 100 & 0.96 \\ \hline
1\_shot\_rlw & ood\_cons\_len\_10 & 1,2,4 & 10 & 100 & 0.89 \\ \hline
\end{tabular}
\label{tab:arglen}
\end{table}

For the next test, we wanted to see if model performance would increase with the number of shots (input/output sample pairs) in each example.  Table~\ref{tab:shots} shows the results.  4 shots seems to be the optimal, with performance dropping on both sides of that number.

\begin{table}[H]
\centering
\caption{Experiment results showing accuracy with varying number of shots.} \begin{tabular}{|l|l|l|l|p{6cm}|}
\hline
\textbf{Task} & \textbf{Split} & \textbf{Runs} & \textbf{Accuracy} & \textbf{Notes} \\ \hline
1\_shot\_rlw & test & 100 & 0.7500 & baseline core task \\ \hline
3\_shot\_rlw & test & 100 & 0.8600 & \\ \hline
4\_shot\_rlw & test & 100 & 0.9600 & dynamically created from 5\_shot\_rlw \\ \hline
5\_shot\_rlw & test & 100 & 0.9500 & \\ \hline
7\_shot\_rlw & test & 100 & 0.9300 & dynamically created from 10\_shot\_rlw \\ \hline
10\_shot\_rlw & test & 100 & 0.9200 & \\ \hline
\end{tabular}
\label{tab:shots}
\end{table} 

The next test covers 4 variants that we thought might help the performance of the model (Table~\ref{tab:engl}).  Using English words in place of the 2 letter random words improved performance significantly, as did adding the grammar of the TGT dataset to the system prompt.  Using uppercase RLW words (vs. lowercase in core task) hurt performance significantly.

\begin{table}[H]
\centering
\caption{Experiment results showing the impact of varying task aspects on accuracy.} \begin{tabular}{|l|l|l|l|p{5cm}|}
\hline
\textbf{Task} & \textbf{Split} & \textbf{Runs} & \textbf{Accuracy} & \textbf{Notes} \\ \hline
1\_shot\_rlw & test & 100 & 0.7500 & baseline core task \\ \hline
1\_shot\_eng & test & 100 & 0.8800 & uses English words vs. RLW \\ \hline
1\_shot\_rlw & test & 100 & 0.8000 & adding grammar to prompt \\ \hline
1\_shot\_rlw & ood\_lexical & 100 & 0.5400 & uses uppercase RLW words \\ \hline
\end{tabular}
\label{tab:engl}
\end{table}

For our final set of tests, we want to test the ability of the model to generalize outside of the N-shot distribution on the cue we trained it with.  For this, we used the 5\_shot\_rlw task.  The first row of Table~\ref{tab:5shot} shows the baseline performance of the model using 5 shots.  We then tried the 3 out of distribution cues shown in the table.  It can be seen that the model performed significantly worse in these cases, but did not fail as typical non-LLM models tend to do in OOD generalization.

\begin{table}[H]
\centering
\caption{Experiment results showing how differing cue/answer distributions affect accuracy in N-shot tasks.} \begin{tabular}{|l|l|l|l|p{5cm}|}
\hline
\textbf{Task} & \textbf{Split} & \textbf{Runs} & \textbf{Accuracy} & \textbf{Notes} \\ \hline
5\_shot\_rlw & test & 100 & 0.9500 & baseline 5\_shot task \\ \hline
5\_shot\_rlw & dyn\_ood\_lexical & 100 & 0.8500 & N-shots are lowercase rlw; cue is uppercase RLW \\ \hline
5\_shot\_eng & dyn\_ood\_cons\_len & 100 & 0.8400 & N-shots are cons length=1,2,4; cue is cons length=7 \\ \hline
5\_shot\_rlw & dyn\_rev\_cons\_len & 100 & 0.7100 & N-shots are cons length=7; cue is cons length=1,2,4 \\ \hline
\end{tabular}
\label{tab:5shot}
\end{table}

\section{RASP}    \label{sec:RASPApp}
In this appendix, we summarize key insights and features of the Restricted Access Sequence Processing (RASP) language and other related work in the same vein as the TPF framework presented here.

First developed by \citet{weiss2021rasp}, RASP was introduced as a computational model of the transformers. Instead of neural network primitives, the RASP language uses sequence operations as primitives that are conceptually aligned with transformer components. \citet{weiss2021rasp} demonstrated that their handed coded RASP programs are able to solve several sequence-manipulation tasks, but were unable to prove their realizability in transformers. To address this limitation, \citet{lindner2023tracr} proposed a compiler named \texttt{Tracr} that compiles (a subset of) RASP programs into transformer models, thereby proving that for a given program a corresponding model exists which can implement it. These broadly correspond to the PSL language and the realization of its programs as DAT models presented in this work. 

Although conceptually related RASP and PSL differ in some important respects, which end up having important downstream consequences. The most important of these lies in the choice of atomic \textbf{atomic data structure} that underlies each approach. At a high level this leads to the following differences at multiple interrelated levels of analysis:

\begin{itemize}
    \item Functional level: RASP takes a \textit{procedural} perspective on representing transformer computation, while PSL takes a \textit{declarative} perspective. This gives rise to computations that incorporate architectural variations beyond the vanilla transformers in PSL, specifically, recurrence and layer looping.
    \item Algorithmic level: \texttt{Tracr} utilizes the residual stream to encode intermediate variables in the computation procedure specified in RASP, while QKVL uses the inherently defined SSS variables as the residual stream, where variables are stored, modified, and passed along the computation dynamics.
    \item Implementational level: (among abundant differences) \texttt{Tracr} has to restrict RASP to avoid arbitrary boolean operation composition on selectors, while DAT does not have such a constraint.
\end{itemize}

A full overview of the differences is presented in section 1.3. However, for the benefit of the unfamiliar reader we provide a primer on RASP and Tracr in the following subsections.

\subsection{The RASP Programming Language} \label{sec:RASPlang}

Instead of the TGT considered in the current work, the RASP language is based on the idea of \textit{sequence manipulation}, with the goal of characterizing how a transformer encoder could perform multi-step logical inferences over input expressions. That is, RASP programs capture the transformations on the input sequence as a whole at each step of computation through a transformer encoder. Therefore, RASP intuitively focuses on mapping the primitive transformer components, i.e.,  attention and feed-forward computations, into primitives in the programming language.

There are two basic types of RASP operations that respectively correspond to the two main components in the transformer architecture: the element-wise operations (corresponding to the MLP module) and the \texttt{select-aggregate} operations (corresponding to the attention module). An input string \texttt{abc} is converted into a sequence through the two built-in sequence operators (\textit{s-op}s): \texttt{tokens(``abc'') = [``a'', ``b'', ``c'']} and \texttt{indices(``abc'') = [0, 1, 2]}. S-ops are functions that map an input string to a sequence of the same length. S-ops can be composed with arithmetic and logical operators (\texttt{+, -, \%, if, >, <}, etc.): for instance, \texttt{(tokens if (indices \% 2 == 0) else ``-'')(``hello'') = ``h-l-o''}. Constant values are treated as s-ops as well, with a single value broadcasting across all positions in order to maintain the sequence status: for instance, \texttt{length(``abc'') = [3, 3, 3]}. Therefore, the composition of s-ops is conceptualized as element-wise operations in RASP, which matches the intuition behind the MLP layers.

On the other hand, the attention mechanism is conceptualized in RASP as a two step, \texttt{select-aggregate} operation. In a standard transformer, attention scores are obtained by the `$QK$ circuit', which uses the dot product between queries and keys to determine how each position is weighted. The dot product part is captured by the \texttt{select} operation, which takes as input a key sequence $k$, a query sequence $q$, and a binary predicate $p$. It outputs a selection matrix named a \textit{selector} that describes whether condition $p(k,q)$ holds for each $(k,q)$ token pair. 
For instance (reprinted from \citealp{weiss2021rasp}):
\begin{align*}
        S \equiv \texttt{select}([0,1,2],[1,2,3],<) = 
        \begin{bmatrix}
            \textbf{T} & F & F\\
            \textbf{T} & \textbf{T} & F\\
            \textbf{T} & \textbf{T} & \textbf{T}
        \end{bmatrix}
\end{align*}

Next, the `$OV$ circuit' in a transformer produces the output by weighting the symbols in the value vector according to the attention scores. The weighting and production steps are handled by the \texttt{aggregate} operation, which takes as input a selector and a sequence; it outputs another sequence that averages for each row of the selector the values of the sequence in its selected columns --- the ``averaging over the binary values in each selector row'' part does the weighting and the ``producing the output given the input sequence'' part does the value-vector-based production part.
For instance, using the selector $S$ above (also from \citealp{weiss2021rasp}):
\begin{align*}
        \texttt{aggregate}(S,[10,20,30]) = 
        [10, 15 ,20]
\end{align*}

Thus, the \texttt{select-aggregate} operation composes a key, a query, and a value s-op into an output s-op. Through a combination of element-wise operations and \texttt{select-aggregate} operations, a RASP program composes primitive s-ops into a final s-op, which maps the input string into a sequence of the same length. As is mentioned in Section~\ref{sec:progtransformers}, it is demonstrated that RASP programs could be designed to solve several sequence manipulation tasks such as: reversing, histogram, double-histograms, sorting, ranking by frequency, etc. Here, we walk through the RASP solution to a simple task of reversing (e.g., \texttt{reverse(``abcde'')=``edcba''}). The RASP program for the reversing task is presented as follows (reprinted from Figure 4 of \citealp{weiss2021rasp}):

\begin{lstlisting}
opp_index = length - indices - 1;
flip = select(indices, opp_index, ==);
reverse =  aggregate(flip, tokens);
\end{lstlisting}
\medskip

This program requires a 2-layer transformer with 1 head per layer. The first layer computes line 1 of the program, where the \texttt{select-aggregate} mechanism is required to compute the \texttt{length} s-op. Specifically, instead of a primitive, \texttt{length} is formally defined as \texttt{length = 1 / aggregate(select(\textbf{1},\textbf{1},==), indicator(indices==0))}. Then, given input string \texttt{``abcde''}, \texttt{indicator(indices==0) = [1, 0, 0, 0, 0]}. It follows that:

\begin{align*}
        \texttt{select\_all} \equiv \texttt{select}(1,1,==) = 
        \begin{bmatrix}
            \textbf{T} & \textbf{T} & \textbf{T} & \textbf{T} & \textbf{T}\\
            \textbf{T} & \textbf{T} & \textbf{T} & \textbf{T} & \textbf{T}\\
            \textbf{T} & \textbf{T} & \textbf{T} & \textbf{T} & \textbf{T}\\
            \textbf{T} & \textbf{T} & \textbf{T} & \textbf{T} & \textbf{T}\\
            \textbf{T} & \textbf{T} & \textbf{T} & \textbf{T} & \textbf{T}\\
            \textbf{T} & \textbf{T} & \textbf{T} & \textbf{T} & \textbf{T}
        \end{bmatrix}
\end{align*}

The \texttt{select-aggregate} attention operation in layer 1 thus returns an s-op where each entry is the reciprocal of the input length:

\begin{align*}
        \texttt{aggregate}(\texttt{select\_all}, \texttt{indicator(indices==0)}) = 
        [0.2, 0.2, 0.2, 0.2, 0.2]
\end{align*}

The following two element-wise operations are then applied on the output of \texttt{aggregate} to compute the \texttt{opp\_index}. They are composed into a single MLP layer, and the output s-op is stored in the residual stream for later use.
\begin{itemize}
    \item Taking the reciprocal on the output of the attention block: \texttt{length = 1 / [0.2, 0.2, 0.2, 0.2, 0.2] = [5, 5, 5, 5, 5]};
    \item Get a reversed s-op of the \texttt{indices}: \texttt{opp\_index = length - indices - 1 = [5, 5, 5, 5, 5] - [0, 1, 2, 3, 4] = [1, 1, 1, 1, 1] = [4, 3, 2, 1, 0]};
\end{itemize}

The second layer computes line 2 and 3 of the program, where line 2 specifies the selector and line 3 specifies the \texttt{aggregate} operation. No element-wise operations are needed at this layer. Given the s-op representing the reversed indices, it is natural to define the attention pattern that selects input characters in the reversed order:

\begin{align*}
        \texttt{flip} \equiv \texttt{select(indices, opp\_index,==)} = 
        \begin{bmatrix}
            F & F & F & F & \mathbf{T}\\
            F & F & F & \mathbf{T} & F\\
            F & F & \mathbf{T} & F & F\\
            F & \mathbf{T} & F & F & F\\
            \mathbf{T} & F & F & F & F
        \end{bmatrix}
\end{align*}

Then, the final \texttt{aggregate} operation applies the \texttt{flip} selector on the input s-op to obtain the final output:
\begin{align*}
        \texttt{aggregate}(\texttt{flip}, \texttt{tokens}) = 
        \texttt{[e, d, c, b, a]}
\end{align*}

Notice that each RASP program decides the number of layers and the number of attention heads per layer needed for executing the program. See Appendix~\ref{sec:tracr} for more details.

The following table presents rough functional correspondences between elements in RASP, the standard transformer architecture, and PSL.

\begin{center}
\small
\resizebox{\linewidth}{!}{%
\begin{tabular}{|p{3cm} p{3cm} p{3cm} p{4cm}|} 
 \hline
 \textbf{RASP} & \textbf{Function} & \textbf{Transformer} & \textbf{PSL} \\
 \hline\hline
 \textit{s-op} & string $\rightarrow$ sequence & hidden states encoded in the residual stream & cell states in \texttt{SSS} \\ 
 \hline
 \textit{selector} & (s-op $k$, s-op $q$, predicate $p$) $\rightarrow$ selector & attention matrices & Production Condition \\
 \hline
 element-wise operations & s-op $\rightarrow$ s-op & MLP modules & N.A. \\
 \hline
 \texttt{select-aggregate} operations & (selector $S$, s-op $v$) $\rightarrow$ s-op & attention heads & Production (evaluating Condition and taking Action) \\
 \hline
\end{tabular}
}
\end{center}

\subsection{The \texttt{Tracr} Compiler} \label{sec:tracr}
Given the structure of the RASP programming language, the \texttt{Tracr} compiler first compiles RASP into an intermediate representation that operates on subspaces of the residual steam, which conceptually corresponds to the QKVL in TPF, and then compiles the intermediate representations into weights and matrices that realize a decoder-only transformer model, which conceptually corresponds to the DAT in TPF. We summarize the steps of compiling RASP programs into the intermediate level representation below.

\begin{enumerate}
    \item Given a RASP program, create a computational graph with each node representing a RASP expression (either an s-op or a selector) and each edge representing a RASP operation (either element-wise or \texttt{select-aggregate}). This graph encodes the dependencies between how s-ops are composed, the order of which is directly translated into attention and MLP block arrangement.

    \item Given a pre-determined vocabulary and a maximal sequence length, for each node in the graph, treat it as a variable and infer the set of all possible values it can take. Allocating a subspace of the residual stream (by assigning a set of basis vectors) to this variable, such that every variable has a subspace orthogonal to the subspaces of other variables. This achieves a disentangled residual stream, so that every operation reads from and writes to a dedicated subspace. \label{2ndPt}

    \item Given the inferred values in step \ref{2ndPt}, translate each individual node into a transformer block. 
    \begin{itemize}
        \item \texttt{Tracr} supports two types of representations: for a categorical variable, each sequence token is represented as a one-hot vector; for a numerical representation, each sequence token is represented as a scalar value in a one-dimensional space.

        \item The element-wise operations are translated into MLP blocks based on manually engineered heuristics, with categorical variables handled by lookup tables and numerical variables handled by piecewise linear approximations: discretizing the range of possible values into buckets (with the granularity chosen to minimize approximation error).

        \item The select-aggregate operations are translated into attention blocks. A selector is represented by a $W_{QK}$ matrix, and an aggregate operation is represented by a $W_{OV}$ matrix, with only hard attention and categorical inputs allowed.
    \end{itemize}

    \item Translate the entire computational graph into an arrangement of transformer blocks. With the goal of finding the smallest possible model, first find out the total number of layers needed by computing the longest path from input to a given node, which denotes the number of attention and MLP modules needed to compute the represented steps. Then, arrange the nodes into alternating attention and MLP blocks without violating any dependency in the graph.

    \item Embed each s-op (both the input s-ops to an operation and the output s-ops from an operation) into its own orthogonal subspace, and construct the residual stream space as the direct sum of all components’ input and output spaces. 
\end{enumerate}

\subsection{Points of Contrast}

Despite various similarities between \{RASP, \texttt{Tracr}\} and \{PSL, QKVL, DAT\}, here we focus on several fundamental points of contrast between the two approaches. The choice of different atomic data structures between RASP and PSL leads to drastic difference in conceptualizing the Transformer computations each language is designed to represent, across multiple levels of analysis. In RASP, the atomic representation unit is a sequence operator (\textit{s-op}), a sequence of symbols with no internal structure within each symbol. The other basic representation in RASP, selectors, are binary matrices built on a pair of \textit{s-op}s together with a predicate. In contrast, as is defined in (40b), each computation step of PSL is represented by a layer / sequence of cell states, where each cell state encodes a discrete set of variables such as \texttt{region} ($r$), \texttt{field} ($f$), \texttt{index} ($d$), in addition to \texttt{position} ($p$) and \texttt{symbol} ($s$). That is, PSL carries a richer data structure compared to RASP, as each cell in a PSL state contains multiple registers in addition to the symbol information, while each unit in the s-op sequence carries only the symbol information. The richness of information in PSL, represented by the state-structure space (SSS), induces the following distinctions compared to RASP.

\subsubsection{Functional level: Procedural vs. Declarative Perspectives}
In RASP, the two types of basic operations, element-wise operations and select-aggregate operations, are defined over \textit{s-op}s that represent sequence-wise computation: transforming an \textit{s-op} (and a selector) into another \textit{s-op}. This sequence manipulation nature of RASP programs displays a procedural perspective: instead of targeting specific positions or symbols in a sequence, RASP specifies computations / transformations on entire sequences, as there is no inherently built-in operations on a more fine-grained level. In compilation, \texttt{Tracr} also follows this perspective by tracing a RASP program into a computational graph, where nodes are intermediate \textit{s-op} representations and edges are sequence-transforming operations. The nature of variables associating to specific nodes in the whole computational graph reflects its procedural character.

On the other hand, a PSL program consists of a sequence of productions, where each production is specified by a \textit{Condition} and an \textit{Action}. Although the sequence of productions still specifies the procedure of how PARSE and GEN should take place at the higher level, the variables encoded in each cell state at every step of computation are associated only to individual symbol positions to support for production condition checking and action application, instead of being sensitive to the computation steps in the whole program. That is, those variables are static properties of the symbols, independent from the step of production the machine is conducting. This reflects the declarative perspective PSL takes on how atomic information is stored and represented. 

Moreover, PSL allows for additional computations that incorporate architectural variations beyond the vanilla Transformer, which RASP cannot achieve by design. PSL allows for features other than \textit{symbol} and \textit{position} to persist across timesteps and positions i.e, the layer-1 state of cell $P+1$ can be set to be the layer-$L$ state of cell $P$. This ability of representing recurrence enables PSL to model computations representable by architectures like the feedback transformer \cite{fan2020addressing} and the staircase transformer \cite{ju2022staircase}. Another built-in property of PSL is that it allows for layer looping: repeated application of particular layers. Thus, PSL can also model computations representable by universal transformers \cite{dehghani2018universal} and MOE-Universal transformers \cite{csordas2024moeut}. RASP, on the other hand, cannot represent such computations.

\subsubsection{Algorithmic level: Treatments on the Residual Stream}
After tracing a RASP program into a computational graph, \texttt{Tracr} allocates subspaces in the residual stream for storing all nodes in the graph, i.e., all the intermediate \textit{s-op}s needed for later steps of computation. This treatment of the residual stream again reflects the procedural perspective RASP and \texttt{Tracr} take: the residual stream is used as the storage space for reading and writing intermediate variables at the representation unit of sequences, and the what has been written into the subspaces reflect the computational traces or activation states the program goes through in executing the program.

Similar to RASP, the residual stream in PSL is also partitioned into subspaces that encode individual variables (or registers, features, attributes) at each position. However, unlike the sequence-wise nature the RASP framework takes, each production in PSL can simultaneously retrieve and check multiple variables at various positions and modify the values of variables targeting specific positions. Accessing and modifying a single variable at any step of the computation is independent from which step the program is current executing, which again reflects the declarative perspective that PSL takes on encoding individual variable information. 

RASP could only read and write the content of full subspaces that encode the entire \textit{s-op}s. Thus, each selector, i.e., attention pattern, could only represent the pairwise comparison of two attributes represented by the two \textit{s-op}s it takes. However, the cell state structures enables the flexibility for PSL to read and write variables at the level of unit sequence positions, allowing for simultaneous comparison of multiple sets of attributes when constructing a single attention pattern. At the algorithmic level, QKVL further refines the complexity of representation unit by specifying each cell state with five cell attributes: \texttt{query, key, value, input}, and \texttt{output}, where each attribute itself is a state structure in $\text{SSS}$ that encodes the lower level variables, as is specified in (68). This nested structure at each symbol position in QKVM further exhibits the data-container, declarative character that TPF takes. Again, representing information as symbol sequences versus as data containers give rise to the disparity in the fine-grainedness of information access.

\subsubsection{Implementational level: Constraining Boolean Operator Compositions}
A further distinction comes at the implementational level. Both RASP and PSL allow for compositions of 'selectors' (in RASP) or conditions (in PSL), allowing layers to harness multiple features at once and thereby efficiently represent programs of increasing complexity. However, in terms of realising this functionality, i.e., compiling such programs into a corresponding transformer, the two frameworks differ. Tracr (the compiler for RASP) cannot compile programs containing operator composition. Though \citet{lindner2023tracr} provide some discussion of how this could be achieved in Appendix B of their paper, their compiler does not support such, more complex, operations. On the other hand, PSL programs involving complex compositions of conjunctions and negations of conditions are readily compilable into DATs. This makes representing sophisticated programs far simpler, which is further complemented by the DAT explorer discussed in the next section.

\subsubsection{Additional Contribution: the Transformer Explorer Visualization Tool}
Besides the conceptual differences discussed above, a further concrete contribution is the visualization tool, Transformer Explorer, that accompanies PSL (Sec.7.2.3 in the main body). Transformer Explorer offers an interactive GUI that enables visualizing the state of the residual stream at each step of computation, given a prompt that represents a TGT input sequence. Specifically, the GUI displays, via symbolic interpretation of activation vectors, each cell state for all positions across all layers, including the changes in values of each register in each cell state. This enables researchers to keep track of feature activations and modifications at each stage of computation, which improves interpretability and is crucial for debugging and understanding more complex programs. To our current knowledge, visualization tools are not available for prior works.  

In sum, TPF provides both a theoretical framework and the necessary existence
proof of its realizability through compilable models, together with tooling that
makes the process easy and interpretable for researchers.

\section{QKVL File Description}    \label{sec:qkvlgrammar}
The QKVL file is in JSON format.  At the top level, it consists of a single dictionary containing system maps and an entry for all the productions from the PSL program: \\

\begin{tabular}{|l|l|}
\hline
Dictionary Key                    & Dictionary Value \\
\hline

register\_map   & A dictionary of register names and their associated short names \\  
constants\_map  & A dictionary of constant names and their optional associated strings \\ 
system\_map     & A dictionary of system reserved registers and their associated register names \\ 
watch\_list     & A list of registers to be watched (for \texttt{dat\_explorer}) \\ 
weights         & A list of blocks (each either a production dictionary or a repeat-dictionary) \\  \hline
\end{tabular}
\\

\noindent See Sec.~6.1.1 for explanation of the term ``weights'' here: these are symbolic instructions, not numerical weights, but they will later be compiled into numerical weight matrices for implementation in the DAT.

\vspace{.5cm} 
\noindent A production-dictionary represents a PSL production. It contains the following dictionary keys and values:  

\medskip

\begin{tabular}{|l|l|}    \hline
Dictionary Key     & Dictionary Value \\  \hline
layer\_comment     & the comment associated with this block  \\  
causal\_attn       & specifies if causal attention in effect for block \\ 
right\_match       & specifies if right match attention in effect for block \\  
weights            & the weights dictionary for this block    \\  \hline
\end{tabular}
\\
\medskip

\noindent A weights-dictionary represents a production's conditions (in the `q' and `k' dictionary keys) and its action (in the `v' dictionary key):  

\medskip

\begin{tabular}{|l|l|}
\hline
Dictionary Key           & Dictionary Value \\
\hline
q         & a registers dictionary for the query weights        \\
k         & a registers dictionary for the key weights       \\
v         & a registers dictionary for the value weights \\
\hline
\end{tabular}
\\
\medskip

A registers-dictionary represents destination/source pairs.  The dictionary keys are the destination registers 
and the associated dictionary values are the source registers or constants.  Short names (usually the first letter 
of the register name) are used for both destination and source registers (except ``index'' and ``parse'' are respectively abbreviated ``d'' and ``a'').  Each register (e.g., ``p'') 
can be optionally followed by 1 of 3 register modifiers:
\begin{verbatim}
   - p            (the value of p in the current column "N")
   - p*           (the value of p in the previous column "N - 1")
   - p`           (the value of p in the column "n")
   - p*`          (the value of p in the column "n - 1")

\end{verbatim}
If the source value is a register, it can be followed by an optional function, for example:
\begin{verbatim}
    - p@pos_increment
\end{verbatim}
The currently supported function names are: \texttt{@pos\_increment}, \texttt{@pos\_decrement}.\\

\noindent
By default, source values in the k- and v-registers dictionaries represent the ``=='' operator.  

\noindent
Source values for the ``!='' operator take the form:
\begin{verbatim}
    - ["!=", <register or constant>]
\end{verbatim}

\noindent
Source values for the ``in'' operator take the form:
\begin{verbatim}
    - ["in", <comma separated list of constants>]
\end{verbatim}

\noindent
Here is a sample of a production dictionary:  
\begin{verbatim}
    {
        "layer_comment": "// parse  step 1b. start Cue ",
        "causal_attn": false,
        "right_match": false,
        "weights": {
            "q": {
                "s`": "s",
                "p`": "1",
                "p": "p",
                "a": "a"
            },
            "k": {
                "s": [
                        "in",
                        "-",
                        ".",
                        "A"
                    ],
                "p`": "p",
                "p": [
                    "!=",
                    "1"
                ],
                "a": "1"
            },
            "v": {
                "r": "CQ",
                "t": "D",
                "f": "FQ"
            }
        }
    }
\end{verbatim}
A repeat dictionary represents a the use of the ``repeat'' keyword in PSL to repeat a group of blocks
until the specified value changes.  It consists of 3 key/value pairs:
\\
\medskip

\begin{tabular}{|l|l|}
\hline
Key                    & Value \\
\hline
layer\_comment         & the comment closest to the start of the repeat block         \\
until                  & the stopping condition for the repeat processing  \\
weights                & the list of weight dictionaries for the inner blocks         \\
\hline
\end{tabular}
\\
\medskip

\noindent Currently, the only condition supported for ``until'' is ``NO\_CHANGE'', meaning the blocks are repeated until the value of all registers in all columns after processing the last inner block matches their values before executing the first inner block.
\\

Here is a sample of a repeat dictionary: 
\begin{verbatim}
    {
        "layer_comment": "// repeat pre_2a, 2a. propagate XQ rightward",
        "until": {},
        "weights": [
            {
                "layer_comment": "// parse  step pre_2a. set prev_region",
                "weights": {
                    "q": {
                        "p`": "p@pos_decrement",
                        "a": "a"
                    },
                    "k": {
                        "p`": "p",
                        "a": "1"
                    },
                    "v": {
                        "r*": "r"
                    }
                }
            },
            {
                "layer_comment": "// parse  step 2a. propagate XQ",
                "weights": {
                    "q": {
                        "r*": "r*",
                        "r": "r",
                        "a": "a",
                        "p": "p"
                    },
                    "k": {
                        "r*": "XQ",
                        "r": "R",
                        "a": "1",
                        "p": "p"
                    },
                    "v": {
                        "r": "XQ"
                    }
                }
            }
        ]
    }

\end{verbatim}

\section{Compiling a PSL program into a QKVL instruction file}    \label{sec:Apppsltoqkvl}

\subsection{Register Abbreviation}

Register names are translated to their abbreviation using the ``registers'' map specified in the PSL program.  In addition, if the position-index of the register is ``\texttt{n}'',  a backquote (``\texttt{`}'') is appended to the abbreviation.%
\footnote{In Production P5b of (99), the PSL Condition ``region[$n$] == XQ, region[$N$] == CQ'' becomes in QKVL \ssf{query}: $\{r : r, r^\backprime : \textrm{XQ}\}$, \ssf{key}: $\{r : \textrm{CQ}, r^\backprime : r\}$.
These match when \ssf{query}[$N$] == \ssf{key}[$n$], i.e., if and only if  $r[N] ==$ CQ and $r[n] ==$ XQ.
}%
Also, if ``\texttt{@function}'' is specified after the register index, then ``\texttt{@function}'' is appended to the abbreviation. 
The constants for initial values, ``\texttt{R\_INIT}'' and ``\texttt{T\_INIT}'', are abbreviated to ``\texttt{R}'' and ``\texttt{T}'', and the variables named \texttt{w\_temp} are abbreviated to ``\texttt{w}'' (for $w = x, y, z$).

\subsection{Production Block Processing  \label{sec:prodBlock}}

Each \verb|<production block>| is processed as follows:

\begin{enumerate}
    \item The comment closest to the beginning of the production block in the PSL is captured.
    
    \item Empty $q$, $k$, and $v$ register dictionaries are created.
    
    \item Each condition in the condition list is added to the $q$ and $k$ register dictionaries as follows:
    
    \begin{itemize}
        \item The left-hand --- target --- register name is translated to its abbreviation.
        
        \item The right-hand --- source --- register name or constant is translated to its abbreviation. If the comparison operator of the condition is ``!='', the right-hand abbreviation is translated to the following list of strings: [``!='', \verb|<right abbreviation>|].
        
        \item If the index of the left-hand register is ``n'' or ``*:n'', then:
        
        \begin{itemize}
            \item The left/target abbreviation becomes the dictionary key, and the right/source abbreviation becomes the dictionary value; they are added to the $k$-register dictionary. Then, to the $q$-register dictionary, the left/target abbreviation is added as both the dictionary key and value, stripping off the backquote of the value, if any.
        \end{itemize}
        
        \item Otherwise:
        
        \begin{itemize}
            \item The left/target abbreviation becomes the dictionary key, and the right/source abbreviation becomes the dictionary value; they are added to the $q$-register dictionary. Then, to the $k$-register dictionary, the left/target abbreviation is added as both the dictionary key and dictionary value, stripping off the backquote of the value, if any.
        \end{itemize}
    \end{itemize}
    
    \item Each assignment in the assignment list is added as a dictionary key/value pair to the $v$-register dictionary.
    
    \begin{itemize}
        \item The left-hand/target register name is translated to its abbreviation.
        
        \item The right-hand/source register name or constant is translated to its abbreviation.
        
        \item Using the left-hand/target abbreviation as the dictionary key and the right-hand/source constant or abbreviation as the dictionary value, the pair is added to the $v$-dictionary.
    \end{itemize}
    
    \item A weights-dictionary is created to hold the $q$-, $k$-, and $v$-register dictionaries (dictionary keys are ``q'', ``k'', and ``v'', and the dictionary values are the associated register dictionaries).
    
    \item A production-dictionary is created consisting of a ``layer\_comment'' dictionary key/value and a ``weights'' dictionary key/value (to hold the weights-dictionary).
\end{enumerate}

\subsection{Repeat Block Processing}

Each repeat block is processed as follows:

\begin{enumerate}
    \item The comment closest to the start of the repeat block is captured.
    
    \item The ``until'' condition is captured.
    
    \item Each production block within the repeat scope is processed as described in Sec.~\ref{sec:prodBlock}, resulting in a list of production dictionaries.
    
    \item A new repeat-dictionary is created with the following dictionary key/values:

\begin{tabular}{|l|l|}
\hline
Key           & Value \\
\hline
layer\_comment         & the comment for the repeat block          \\
until       & the condition for terminating the repeat process          \\
weights         & the list of production-dictionaries          \\
\hline
\end{tabular}
\end{enumerate}
The final result is a list of production and repeat dictionaries that are converted to JSON format and output to a file.

\section{Tensor Product Representations}    \label{sec:TPR}
\renewcommand{\v}[1]{\overrightarrow{#1}}
\newcommand{\vi}[2]{\overrightarrow{\mathsf{#1}_#2}}
A general method for encoding symbol structures as neural activity vectors is \textit{Tensor Product Representations} (TPRs) \cite{smolensky1987analysis,smolensky1990tensor}.%
\footnote{
Related methods, many under the rubric of \textit{Vector Symbolic Architectures}
\cite{kleyko2022survey,schlegel2022comparison}, might well work as well as TPRs here, provided sufficiently accurate means are available for detecting perfect matching for Conditions and perfect value-changing for Actions.
}
Any symbol structure type can be decomposed as a set of \textit{structural roles}, and a particular token of that type is defined by assigning \textit{fillers} to these roles.
In the special case relevant to TPF, the state-variable structure (37) is defined by roles which are the state variables, and fillers that are the values that these variables can be assigned.
Retaining the notation of the main text of this paper, we can write \textsf{r:f} for the \textit{binding} of the role \textsf{r} to the filler \textsf{f}.
The particular state-variable structure \textsf{S} shown in (37) is defined by the bindings $\beta(\textsf{S}) = \{ r : \textrm{XQ}, \; f : \textrm{FQ}, \; s : \textrm{Q}, ... \}$

Let us group roles/variables by their set of possible fillers/values.
The group $\gamma \in 1, \ldots, \Gamma$ contains the $n_\gamma$ roles $\textsf{R}^\gamma = \{ \textsf{r}^\gamma_1, \ldots, \textsf{r}^\gamma_{n_\gamma} \}$, all of which take fillers in the same set, $\textsf{F}^\gamma = \{ \textsf{f}^\gamma_1, \ldots, \textsf{f}^\gamma_{m_\gamma} \}$.
(For TGT, one group would be the set of region-valued variables $\{r, r^*, r^{\backprime}, r^{*{\backprime}}\}$.)
To create a TPR, 
we adopt row-vector embeddings of roles $\mathbb{E}_{\textrm{R}} : \textsf{R}^\gamma \rightarrow \v{\textsf{R}^\gamma}; \textsf{r}^\gamma_i \mapsto \v{\mathsf{r}^\gamma_i}$ and fillers $\mathbb{E}_{\textrm{F}} : \textsf{F}^\gamma \rightarrow \v{\textsf{F}^\gamma_}; \textsf{f}^\gamma_j \mapsto \v{\mathsf{f}^\gamma_j}$.
Then the TPR for a particular symbol structure $\textsf{S}$ defined by the bindings 
$\beta(\textsf{S}) = \{ \textsf{r}^\gamma_i:\textsf{f}^\gamma_{j(i)} \ | \ \gamma = 1, \ldots, \Gamma; \ i = 1, \ldots, n_\gamma; \ j(i) \in 1, \ldots, m_\gamma \}$  
is the tensor 
$\textbf{T}_{\textsf{S}} \equiv \bigoplus_{\gamma=1}^\Gamma \sum_{i=1}^{n_\gamma} \v{\textsf{r}^\gamma_i} \otimes \v{\textsf{f}^\gamma_{j(i)}} \in \bigoplus_{\gamma=1}^\Gamma \v{\textsf{R}^\gamma} \otimes \v{\textsf{F}^\gamma} \equiv \mathcal{V}$
or equivalently the matrix 
$\textbf{T}_{\textsf{S}} \equiv \bigoplus_{\gamma=1}^\Gamma \sum_{i=1}^{n_\gamma} {\v{\textsf{r}^\gamma_i}}^\top  \v{\textsf{f}^\gamma_{j(i)}}$.
(This `matrix' is essentially a sequence of $\Gamma$ matrices, each operating on a different subspace $\v{\textsf{R}^\gamma} \otimes \v{\textsf{F}^\gamma} $ of the overall direct-sum vector space $\mathcal{V}$.)

We now focus on a single subspace defined by a value of $\gamma \in 1, \ldots, \Gamma$, and leave the superscript $\gamma$ implicit, e.g., writing $n$ for $n^\gamma$.

TPRs are designed so that compositional operations on entire structures can be performed in parallel by operating on them with linear transformations \cite{smolensky2012symbolic} --- but also so that it is possible to accurately extract fillers of individual roles when needed.
In the standard case --- lossless TPR encoding --- the role embeddings are chosen to be linearly independent, which guarantees the existence of a set of \textit{dual vectors}  $\{\vi{r}{i}^+\}$ defined so that $\vi{r}{i}^+ \cdot \vi{r}{j}  = \delta_{ij}$.
Then to unbind role $\vi{r}{i}$ it suffices to compute $\vi{r}{i}^+ \; \mathbb{T}_{\textsf{S}} \equiv \widehat{\vi{f}{i}}$. 
It is straightforward to verify that this is an accurate extraction, that $\widehat{\vi{f}{i}} = \vi{f}{i}$: 

$\widehat{\vi{f}{i}} = \vi{r}{i}^+ \; \mathbb{T}_{\textsf{S}} = \vi{r}{i}^+ \left( \sum_{j=1}^n \vi{r}{j}^\top \vi{f}{j} \right)  = \sum_{j=1}^n \left( \vi{r}{i}^+ \cdot \vi{r}{j} \right) \vi{f}{j}  = \sum_{j=1}^n \delta_{ij} \vi{f}{j}  = \vi{f}{i} $ 

This error-free extraction occurs whenever the role embeddings are linearly independent; in the general case, these are dense vectors and the TPR is fully distributed --- every element of the tensor contributes to encoding the filler of every role; there are no separable `registers' for the roles.

Note that in the special case that the linearly independent role embeddings are orthonormal, then $\vi{r}{i}^+ = \vi{r}{i}$. 
This is the fully distributed orthonormal embedding of (69b).

One-hot role embeddings are yet a further special case of orthonormality (since fully dense vectors can also be orthonormal).
In this special case, $\mathbb{T}_{\textsf{S}}$ is simply the matrix in which the $k^{\textrm{th}}$ row is $\vi{f}{k}$.
For the state-variable structure of TPF, the $k^{\textrm{th}}$ row of $\mathbb{T}_{\textsf{S}}$ is exactly the  register for the $k^{\textrm{th}}$ state variable, containing the vector $\vi{f}{k}$ which is the embedding of the value of that variable; this is the semi-local case of (69a).
In the yet further special case when the filler embeddings are also 1-hot, this reduces to exactly the fully local embedding of (68), visualized in (67).

We now show that the DAT can implement a PSL program correctly without using the 1-hot, fully local embedding of state-variable structures deployed in the main text (and the current software).
We will assume a TPR embedding of state structures, and require only that the role embeddings are orthonormal, and that the filler embeddings are normalized.

To work, what DAT requires of its state-variable-structure representation is only that: for Conditions, we can accurately identify perfectly matching \textsf{query} and \textsf{key} vectors; and for Actions, that we can accurately adjust state variables according to the demands encoded in the \textsf{value} vectors.

As for the Condition requirement, the dot product of the TPRs for two SSS structures \textsf{S} and $\textsf{S}'$ (a \textsf{query} and a \textsf{key}, in the DAT case), when we adopt orthonormal (not necessarily 1-hot) role embeddings, is%
\footnote{
Here we have used the identity 
$\left( \vec{u} \otimes \vec{v} \right) \cdot
\left( \vec{x} \otimes \vec{y} \right) =
\left( \vec{u} \cdot \vec{x} \right)
\left( \vec{v} \cdot \vec{y} \right)$
as well as the orthonormality condition on roles
$\vi{r}{i} \cdot \vi{r}{j} = \delta_{ij}$
}

$\textbf{T}_{\textsf{S}} \cdot \textbf{T}_{\textsf{S'}} =
\left( \sum_{i=1}^M \vi{r}{i} \otimes \vi{f}{i} \right) \cdot \left( \sum_{j=1}^M \vi{r}{j} \otimes \vi{f'}{j} \right)
= \sum_{i=1}^M \vi{f}{i} \cdot \vi{f'}{i} 
= \sum_{i=1}^M  \cos ( \vi{f}{i} , \vi{f'}{i}) \! \parallel \! \vi{f}{i} \! \parallel \, \parallel \! \vi{f'}{i} \! \parallel
$

\noindent If all filler vectors are normalized, $\parallel \! \vi{f}{i} \! \parallel \; = 1$, then each term in this dot product reduces to $\cos ( \vi{f}{i} , \vi{f'}{i})$, which is $\leq 1$, with equality holding only when $\vi{f}{i} = \vi{f'}{i}$.
Let the total number of non-null values (non-zero filler vectors) in the query be $m$.%
\footnote{As noted in note 12 of the main text, for the DAT (i.e., transformer weights compiled from a PSL program), the set of state-variables assigned non-null values is always the same for \textsf{query} and \textsf{key}, thus avoiding any match issue arising from variables that are null-valued in one but not the other.
}
Then dividing the dot product by $m$ gives a value less than 1 unless $\vi{f}{i} = \vi{f'}{i}$ for all roles $i$ with non-zero fillers.
So let us define the \textsf{query} and \textsf{key} vectors \textbf{q} and \textbf{k} to be the TPRs for their respective defining state-variable structures \textsf{S} and $\textsf{S}'$ in SSS, divided by $\sqrt{m}$.
Then their dot product is

$\mathbf{q} \cdot \mathbf{k} = 
\left( \textbf{T}_{\textsf{S}}/\sqrt{m} \right) \cdot \left( \textbf{T}_{\textsf{S'}}/\sqrt{m} \right) =
\left( \textbf{T}_{\textsf{S}} \cdot \textbf{T}_{\textsf{S'}} \right) / m$

\noindent So just as for the 1-hot encoding in the body of the paper (70c-i), requiring this dot product to be 1 enforces perfect matching of all non-null variable values specified in the query.

\medskip

For implementing a production's Action, to set the value of variable \textsf{r} in the TPR residual stream $\mathbb{O}$ to \textsf{f}, we use 

$\mathbb{O} \mapsto \mathbb{O} + \vec{\textsf{r}}^\top \left[ \vec{\textsf{f}} - \vec{\textsf{r}}^+ \mathbb{O} \right]$

\noindent In this adjustment, the first term inserts the value \textsf{f} into  \textsf{r} while the second removes the existing value of \textsf{r} (if any).%
\footnote{
Instead, it would also be possible to adapt the approach used in the main text (70c-iv), adding to the old value of the variable an encoding of the new value up-weighted by $\kappa$, and then applying DATnorm to eliminate all but the most-active value.
In the non-1-hot case, DATnorm is defined by extracting the filler vector (value) of each role vector (variable), finding the closest element in the filler-vector dictionary, and setting the variable to that value, as shown above, removing the current value.
This version of DATnorm may be useful for other purposes, but for applying the Action of a production, the method proposed in the text of this Appendix is clearly simpler.
}

\medskip

Since the DAT could function perfectly well with non-1-hot encoding of state variables, given a trained transformer capable of performing ICL, it is entirely possible that it is performing computation quite similar to our DAT, using distributed (dense) encodings.
Testing this hypothesis can be pursued through the steps detailed in Sec.~9.2.

\section{Exploratory training from scratch and testing}     \label{sec:trainscratch}
This appendix includes the results for models trained from scratch on the TGT dataset and then tested on various in and out of distribution splits.  \\

The following sequence to sequence models were tested: vanilla transformer (encoder/decoder), nano\_gpt (decoder only), nano\_gpt\_attn\_only (no MLP layers), cnn, lstm\_attn (LSTM with attention), and mamba.  \\

\noindent
Here are the architecture hyperparameters:
\begin{table}[ht]
\centering
\caption{Architecture hyperparameters.}
\begin{tabular}{|l|c|c|c|c|c|c|}
\hline
\textbf{Model} & \textbf{hidden} & \textbf{filter} & \textbf{layers} & \textbf{heads} & \textbf{state size} & \textbf{bidir} \\
\hline
transformer         & 512 & 512 & 3 + 3 & 1 &  &  \\ \hline
nano\_gpt           & 512 & 512 & 6     & 1 &  &  \\ \hline
nano\_gpt\_attn\_only & 512 & 512 & 6     & 1 &  &  \\ \hline
cnn                 & 512 & 512 & 3 + 3 &   &  &  \\ \hline
lstm\_attn          & 512 &     & 3 + 3 &   &  & false \\ \hline
mamba               & 512 & 512 & 18    &   & 16 &  \\ \hline
\end{tabular}
\end{table}

\noindent
Here are the training hyperparameters:
\begin{table}[ht]
\caption{Training hyperparameters}
\centering
\resizebox{\linewidth}{!}{%
\begin{tabular}{|l|c|c|c|c|c|c|}
\hline
\textbf{Model} & \textbf{LR} & \textbf{weight decay} & \textbf{steps} & \textbf{early stop} & \textbf{batch size} & \textbf{dropout} \\ 
\hline
transformer         & .0001 & 0 & 120,000 & false & 128 & 0 \\ \hline
nano\_gpt           & .0001 & 0 & 120,000 & false & 256 & 0 \\ \hline
nano\_gpt\_attn\_only & .0001 & 0 & 120,000 & false & 256 & 0 \\ \hline
cnn                 & .0001 & 0 & 120,000 & false & 128 & 0 \\ \hline
lstm\_attn          & .0001 & 0 & 120,000 & false & 128 & 0 \\ \hline
mamba               & .0001 & 0 & 120,000 & false & 256 & 0 \\ \hline

\end{tabular}
}
\end{table}

The following splits were tested: train, dev, ood lexical (out of distribution for constituent part vocabulary), ood cons len 7 (constituent lengths of 7), and ood cons count 7 (7 constituents).  \\

All reported metrics were averaged over 3 runs.

\subsection{Training curves \label{sec:traincurves}}
Here are the training curves for each model that we trained.  

\vspace{\baselineskip}
\noindent
Transformer \\
\includegraphics[width=\textwidth]{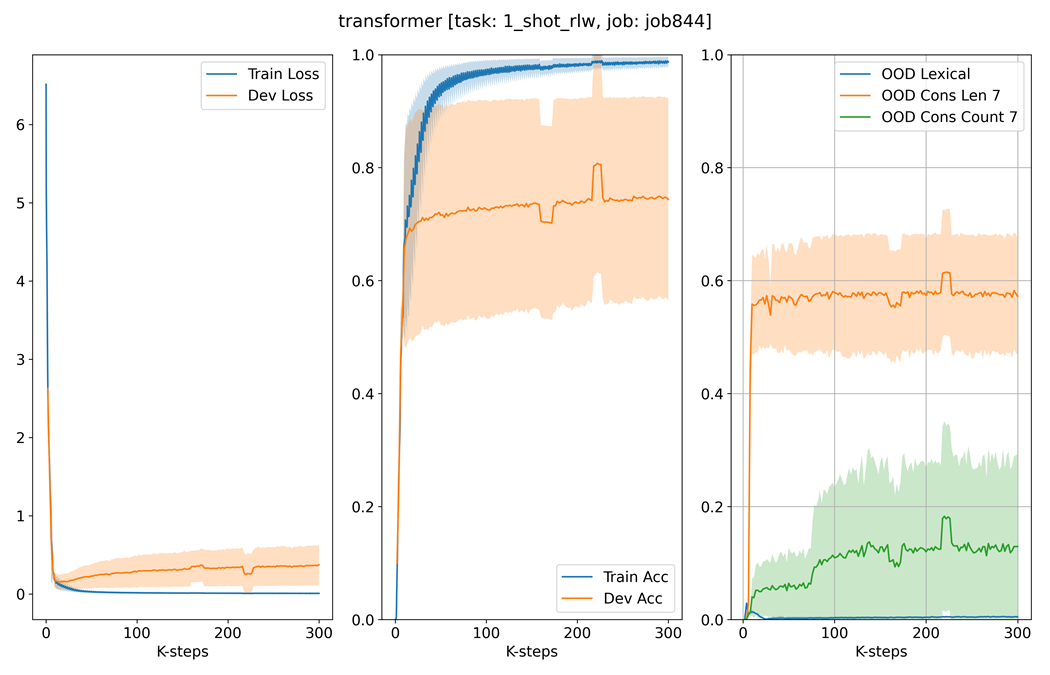}

\clearpage
\pagebreak

\vspace{\baselineskip}
\noindent
Nano GPT \\
\includegraphics[width=\textwidth]{figures/traincurves/nano\_gpt.png}

\vspace{\baselineskip}
\noindent
CNN\\
\includegraphics[width=\textwidth]{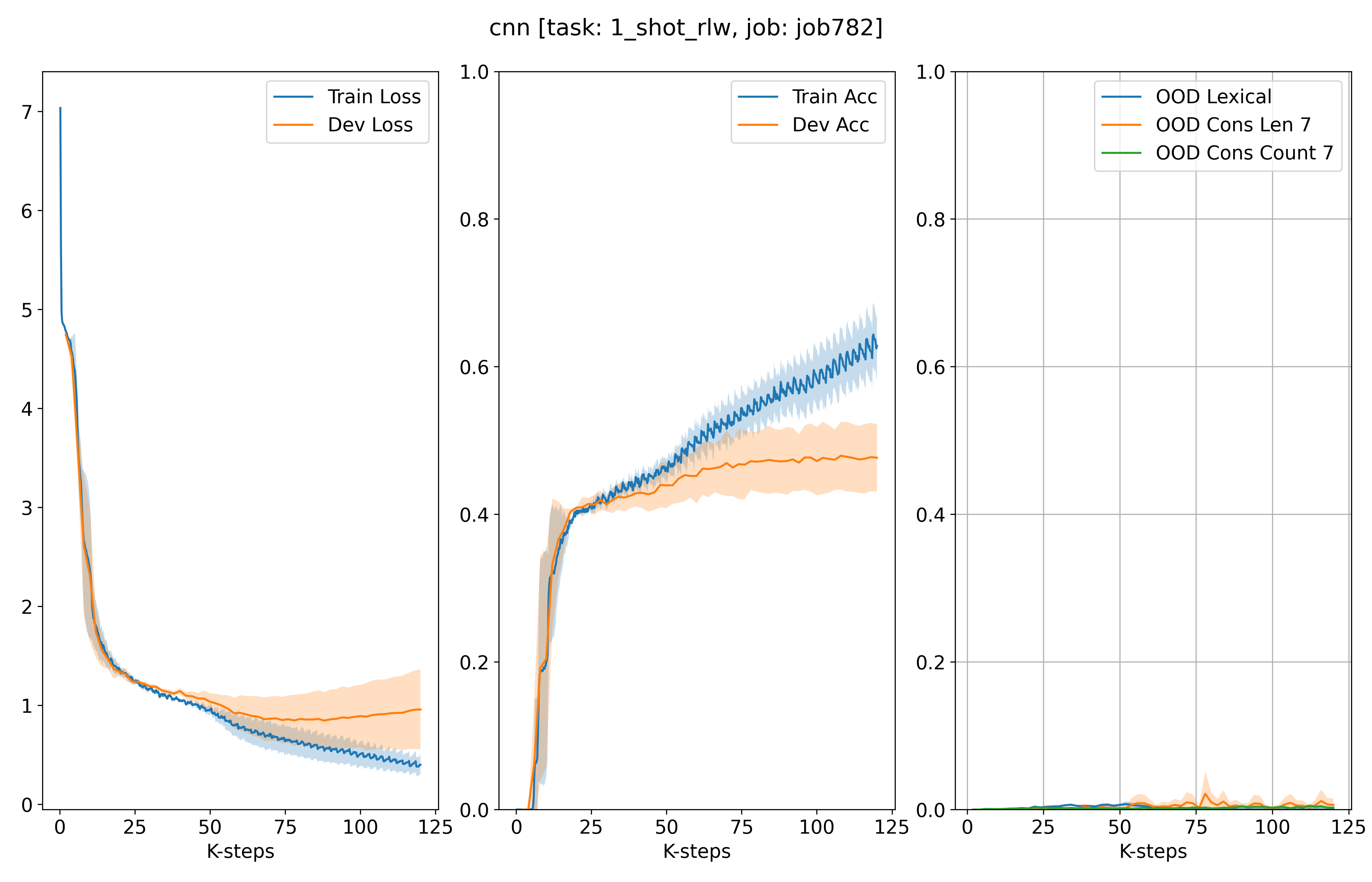}

\vspace{\baselineskip}
\noindent
LSTM with attention\\
\includegraphics[width=\textwidth]{figures/traincurves/lstm\_attn.png}

\vspace{\baselineskip}
\noindent
Mamba\\
\includegraphics[width=\textwidth]{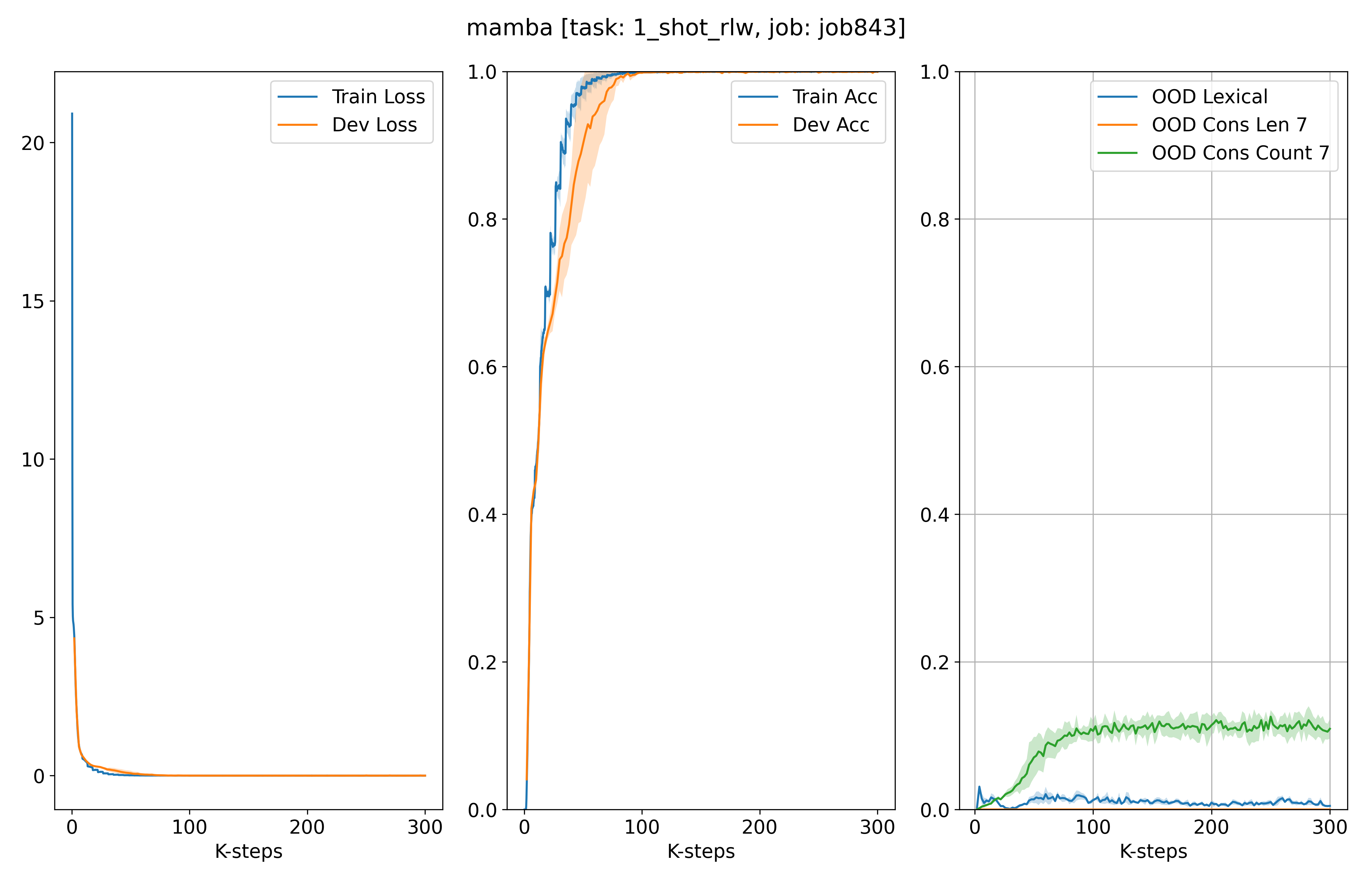}
\pagebreak
\subsection{Results by model \label{sec:tabresults}}
Here are the training results, plotted by model.

\vspace{\baselineskip}
\noindent
Transformer\\
\noindent
\includegraphics[width=\textwidth]{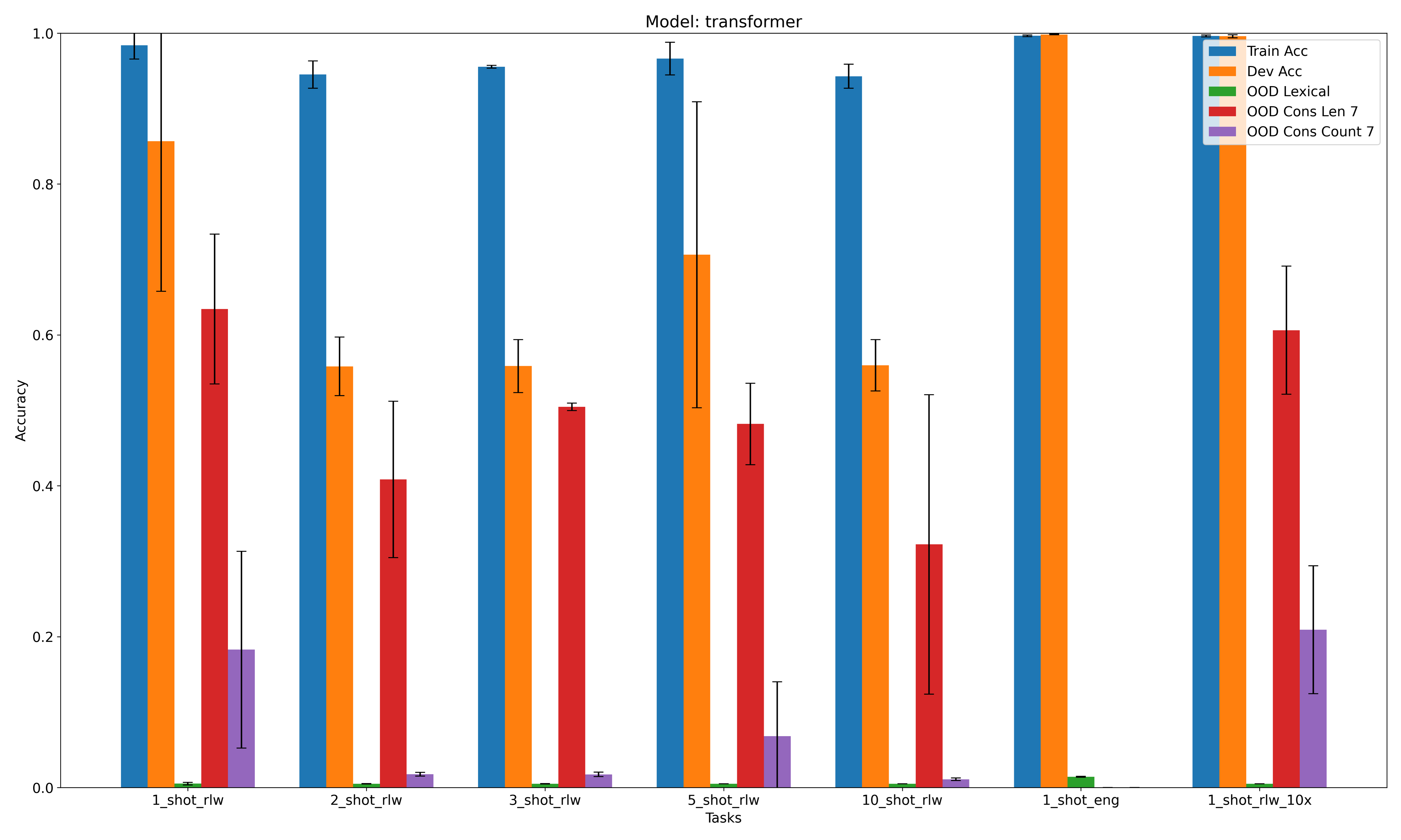}

\vspace{\baselineskip}
\noindent
Nano GPT\\
\noindent
\includegraphics[width=\textwidth]{figures/plotsbymodel/nano\_gpt.png}

\clearpage
\pagebreak

\vspace{\baselineskip}
\noindent
Nano GPT with no MLP layers\\
\noindent
\includegraphics[width=\textwidth]{figures/plotsbymodel/nano\_gpt\_attn\_only.png}

\vspace{\baselineskip}
\noindent
CNN\\
\noindent
\includegraphics[width=\textwidth]{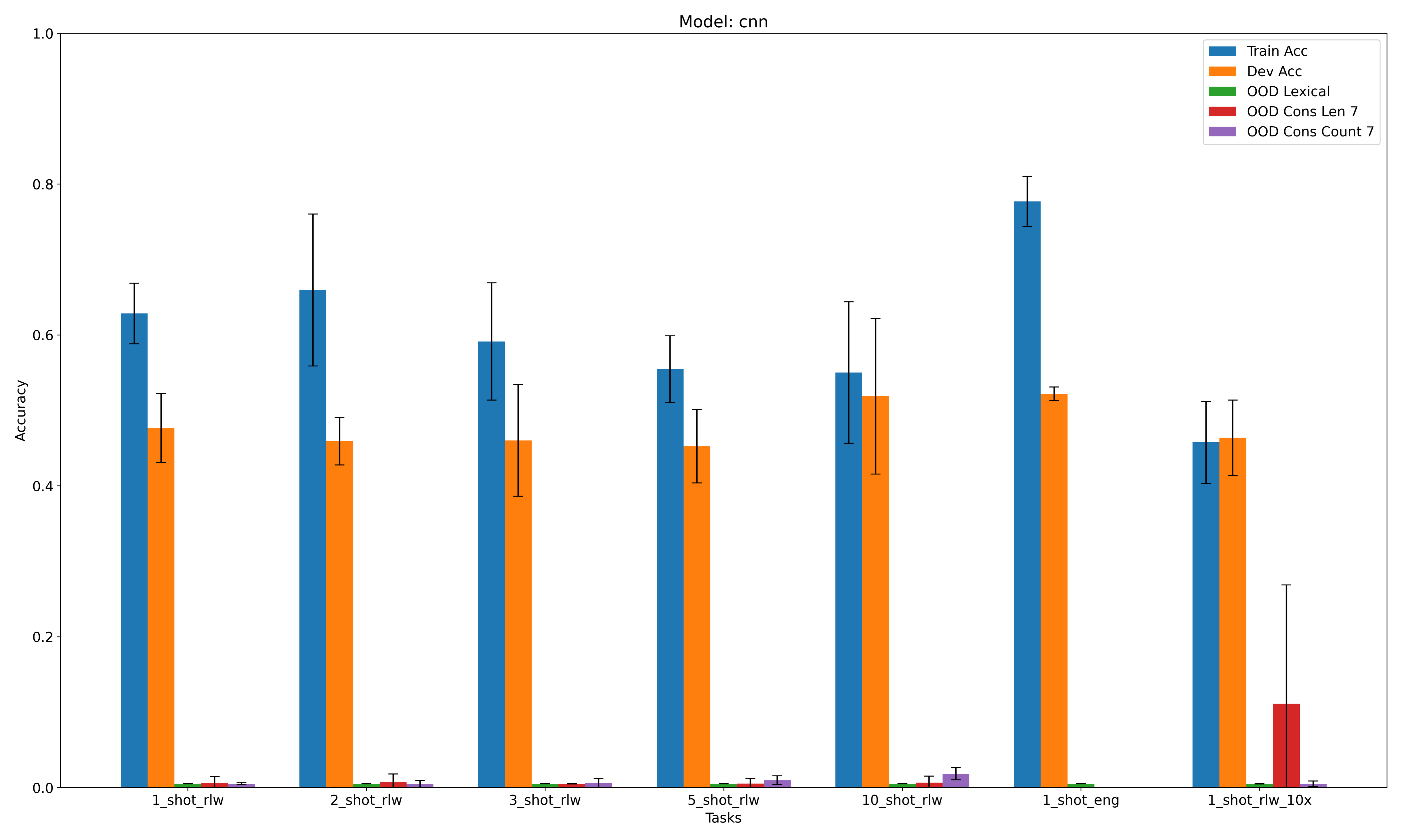}

\vspace{\baselineskip}
\noindent
LSTM with attention\\
\noindent
\includegraphics[width=\textwidth]{figures/plotsbymodel/lstm\_attn.png}

\vspace{\baselineskip}
\noindent
Mamba\\
\noindent
\includegraphics[width=\textwidth]{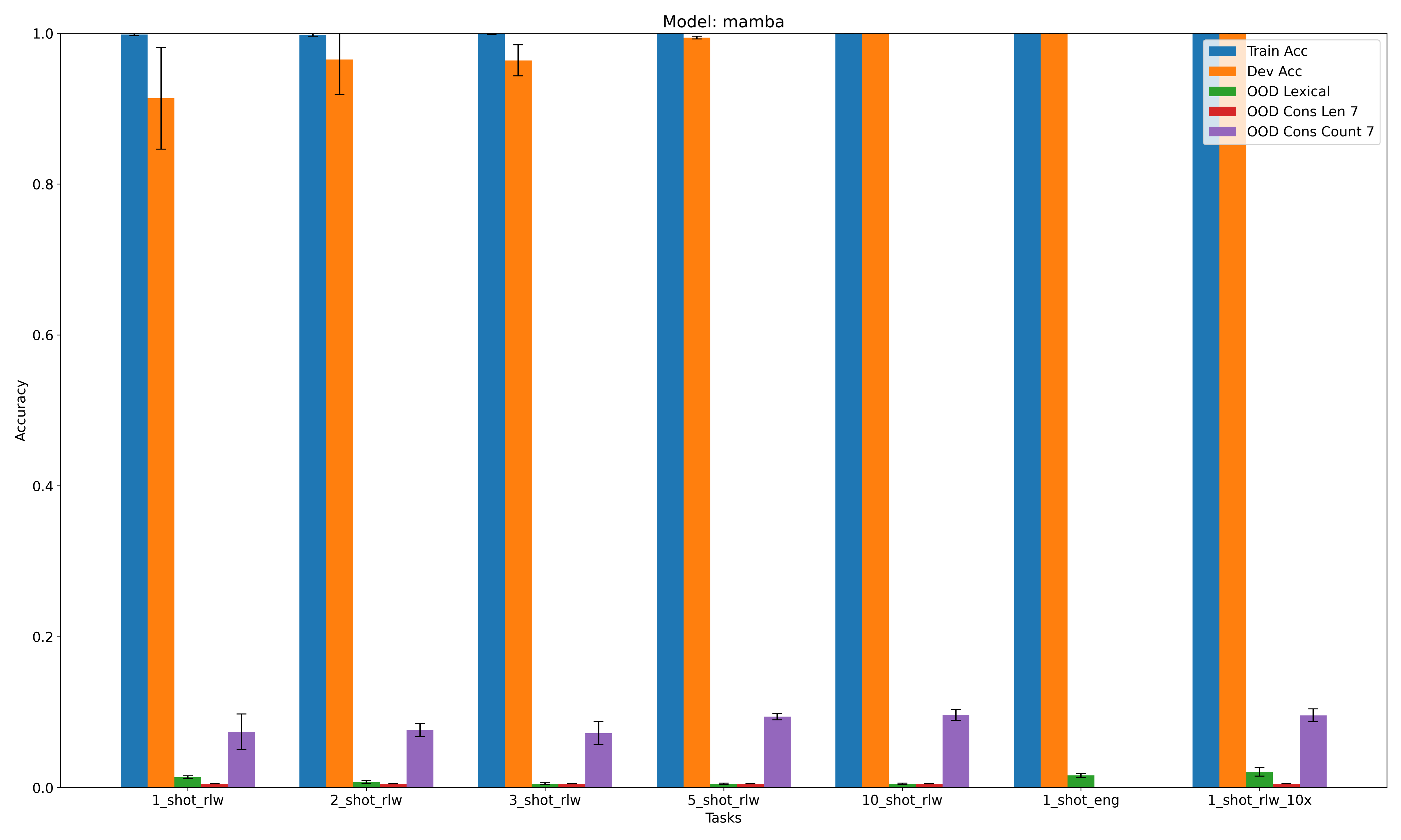}

\end{document}